\documentclass{alt2027}
\usepackage{tikz}
\usepackage{graphicx}
\usetikzlibrary{arrows.meta}

\usepackage{times}
\usepackage{natbib}
\usepackage[T1]{fontenc}
\usepackage{enumitem}
\usepackage{mathtools,bm,mathrsfs,booktabs,array,enumitem,microtype,needspace}
\usepackage{algorithm}
\usepackage{algorithmic} %
\hypersetup{pdftitle={On the Feature Evolution of Temporal Difference Learning},
 pdfsubject={Feature learning through initial and repeated temporal-difference updates},pdfauthor={}}
\allowdisplaybreaks[3]
\numberwithin{equation}{section}
\numberwithin{theorem}{section}
\theorembodyfont{\normalfont}
\newtheorem{condition}[theorem]{Condition}

\makeatletter
\let\ps@jmlrtps\ps@plain
\expandafter\def\csname jmlr@thm@definition@body@font\endcsname{\normalfont}
\expandafter\def\csname jmlr@thm@remark@body@font\endcsname{\normalfont}
\makeatother

\newcommand{\E}{\mathbb E}

\newcommand{\one}{\mathbf 1}
\newcommand{\relu}{\operatorname{ReLU}}

\newcommand{\Var}{\operatorname{Var}}

\newcommand{\Reg}{\operatorname{Reg}}
\DeclareMathOperator*{\argmax}{arg\,max}

\title[Feature Evolution in TD Learning]{On the Feature Evolution of Temporal Difference Learning}
\altauthor{%
  \Name{Shangzhe Li} \Email{shangzhe@unc.edu}\\
  \addr University of North Carolina at Chapel Hill
  \AND
  \Name{Dechen Zhang} \Email{dechenzhang@connect.hku.hk}\\
  \addr The University of Hong Kong
  \AND
  \Name{Difan Zou} \Email{dzou@hku.hk}\\
  \addr The University of Hong Kong
  \AND
  \Name{Weitong Zhang} \Email{weitongz@unc.edu}\\
  \addr University of North Carolina at Chapel Hill
}
\begin{document}
\maketitle

\begin{abstract}
Temporal-difference (TD) learning trains neural representations using
bootstrapped targets generated by evolving value estimates, yet how
predictive features emerge and propagate under incomplete neural
updates remains poorly understood.
We investigate this problem using finite-width two-layer ReLU critics
on a constructed family of smooth stochastic horizon-$H$ MDPs with
unknown quadratic structure and $d$-dimensional observations.
We first show that an informative TD update acquires the quadratic
Bellman signal and yields a representation admitting a bounded
predictive readout. The simultaneous hidden-weight and head update
achieves a strict finite-sample prediction-risk improvement over
matched head-only SGD, with a $2(d+1)$-fold larger population
quadratic response.
We then establish that an imperfect learned continuation can already
provide informative supervision to an earlier zero-reward stage.
By tracking coupled neural updates with one SGD step per target
copy, without requiring intermediate Bellman regressions to be solved,
we obtain a uniform prediction-error transient
$\exp[-\Omega(K/(Hd))]$, up to action-independent shifts and explicit
sampling and finite-width errors.
On the constructed signed-action MDP family whose total reward is bounded by $1$, we show that an exploration-then-commit algorithm can achieve
$\widetilde{\mathcal O}(d^3H^{1/3}T^{2/3})$ regret over $T$ episodes, using
$K=\widetilde{\mathcal O}(Hd)$ global updates.
Our analysis connects bootstrapped feature acquisition, concurrent
neural refinement, and finite-budget policy performance.
\end{abstract}


\section{Introduction}\label{sec:lat-introduction}

Temporal-difference (TD) learning \citep{Sutton1988} is fundamental
to reinforcement learning (RL), estimating future cumulative rewards
by regressing value functions onto Bellman targets combining immediate
rewards and estimated future values. Deep RL algorithms such as
DQN \citep{mnih2013playing,dqn_mnih2015},
DDPG \citep{lillicrap2015continuous}, and
TD3 \citep{fujimoto2018addressing} extend this principle using
neural function approximation. Unlike supervised regression,
however, TD targets depend on the learner's evolving value estimates,
coupling feature learning with bootstrapped supervision.

Despite advances in convergence analysis \citep{dqn_zhang2023},
mean-field representation dynamics \citep{zhang2020tdfeatures},
and regret guarantees \citep{dqn_liu2022}, how predictive features
emerge and evolve under actual neural TD updates remains less
understood. In particular, fitted value iteration analyses often
assume nearly solved Bellman regressions, whereas practical deep RL
performs limited SGD updates using partially updated imperfect targets. Such targets
may already contain useful predictive signals before the downstream
critic becomes accurate. This difference between theory and practice motivates our central question:
\begin{center}
\emph{How do imperfect TD targets drive neural feature learning across stages and improve policies?}
\end{center}

We investigate this question using finite-width two-layer ReLU
critics on a constructed family of smooth stochastic MDPs with
quadratic features. The Bellman targets expose
identifiable quadratic signals, allowing us to analyze their
acquisition and predictive benefit under actual SGD updates.
In a specialized construction, we further
track how imperfect learned continuations supervise earlier
zero-reward stages while all reached critics continue learning,
without requiring intermediate Bellman regressions to be fully solved.
Our contributions are threefold:

\begin{itemize}[leftmargin=*]
\item \textbf{Initial feature acquisition.}
We characterize how an informative TD update acquires the quadratic
Bellman signal in the hidden representation of a two-layered finite-width ReLU network.
We establish a positive feature--signal covariance gain under explicit
data collection, replay sampling, and finite-width conditions
(Theorem~\ref{thm:dqn-feature-gain}).

\item \textbf{Predictive benefit of learned features.}
We quantify how the acquired representation improves prediction
beyond updating the output heads alone. We construct a bounded
predictive readout and prove a strict finite-sample risk improvement
of the actual full SGD update over a matched head-only update,
with a $2(d+1)$-fold larger population quadratic response
(Theorem~\ref{thm:dqn-bounded-bridge}).

\item \textbf{Coupled feature refinement and policy performance.}
We show that an imperfect learned continuation generates informative
supervision at earlier zero-reward stages, while previously reached
critics continue learning. Tracking these coupled neural SGD updates
without requiring accurate intermediate Bellman regression yields an
$\exp[-\Omega(K/(Hd))]$ prediction-error transient and
$\widetilde{\mathcal O}(d^3H^{1/3}T^{2/3})$ regret on the
constructed signed-action MDP family
(Theorem~\ref{thm:dqn-horizon-complete} and
Corollary~\ref{cor:dqn-complete-rates}).
\end{itemize}

\paragraph{Notation.}
Write $[n]=\{1,\ldots,n\}$, and denote Euclidean, operator,
Frobenius, and supremum norms by
$\|\cdot\|_2$, $\|\cdot\|_{\mathrm{op}}$,
$\|\cdot\|_F$, and $\|\cdot\|_\infty$, respectively.
Let $Q_{h,j}$ be the stage-$h$ critic after global TD update $j$,
with $j_h=H-h+1$ its first informative update.
The notation $\mathcal O$ and $\Omega$ hides universal constants,
while $\widetilde{\mathcal O}$ additionally hides logarithmic factors.

\section{Related Work}\label{sec:latent-related-work}

\paragraph{Feature learning under TD.}
Prior work characterizes neural TD representations through mean-field
dynamics \citep{zhang2020tdfeatures}, spectral and invariant-subspace
analyses \citep{lyle2021representation,lelan2023bootstrapped},
joint representation--readout dynamics \citep{tang2023td}, and
feature coadaptation \citep{kumar2022dr3}.
In contrast, we establish finite-sample guarantees for actual ReLU
TD updates, identifying the acquired quadratic Bellman signal and
showing how an imperfect learned continuation generates informative
features at earlier zero-reward stages.

\paragraph{Predictive benefits of learned features.}
\citet{damian2022} study feature acquisition through a hidden-weight
gradient step in Gaussian polynomial regression, followed by readout
optimization. Related results address noisy XOR \citep{lat_meng2024}
and representation learning in low-rank RL
\citep{lat_flambe2020,lat_rep2022}.
We instead quantify the predictive benefit of an actual neural TD
update, constructing a bounded readout and proving a strict
finite-sample risk improvement over matched head-only SGD using
the same bootstrapped targets and replay data.

\paragraph{Neural TD and Q-learning convergence.}
\citet{xu2020neuralq} establish
$\widetilde{\mathcal O}(K^{-1/2}+m^{-1/6})$
iterate-averaged squared value error, plus projection error, under
fixed-policy covariance and mixing conditions.
\citet{dqn_zhang2023} establish geometric convergence with
$\mathcal O(n^{-1/2})$ estimation error under realizability,
local initialization, and stationary replay assumptions.
In contrast, Theorem~\ref{thm:dqn-horizon-complete} tracks coupled,
stage-dependent neural TD updates with one SGD step per target copy,
yielding an $\exp[-\Omega(K/(Hd))]$ transient, up to
action-independent shifts and explicit sampling and finite-width
errors. This connects neural optimization dynamics to the acquisition
and refinement of bootstrapped predictive features without requiring
intermediate Bellman regressions to be solved.

\paragraph{Exploration and policy guarantees.}
\citet{dqn_dann2022} study myopic $\varepsilon$-greedy exploration,
while optimistic methods establish square-root-regret guarantees
\citep{lat_jin_linear2020,dqn_agarwal2023}.
Under favorable exploration, \citet{dqn_liu2022} obtain
$\widetilde{\mathcal O}(T^{2/3})$ regret for smooth Besov classes
and $\widetilde{\mathcal O}(T^{5/6})$ for Barron classes,
using fitted empirical regression.
For tabular MDPs, an explore-then-commit reduction of
\citet{jin2020reward}, restricted to unit-bounded episode returns,
gives $\widetilde{\mathcal O}(H^{5/3}(S^2A)^{1/3}T^{2/3}
+H^{7/2}S^2\sqrt{AT})$ regret.
On our constructed signed-action MDP family,
Corollary~\ref{cor:dqn-complete-rates} establishes
$\widetilde{\mathcal O}(d^3H^{1/3}T^{2/3})$ regret with
$K=\widetilde{\mathcal O}(Hd)$ global SGD updates.
Although the settings differ, our analysis connects finite-budget
policy performance to the actual neural feature-learning trajectory,
counting all exploration episodes.

\section{Preliminaries}\label{sec:lat-preliminaries}
\subsection{Episodic Markov Decision Processes}
We consider episodic Markov Decision Processes (MDPs), which can be represented by a tuple $\mathcal M=(\mathcal X,\mathcal A,H,P,r,\nu_1)$.
At stage $h$, the learner observes $X_h\in\mathcal X_h\subset\mathbb R^d$,
selects $A_h\in\mathcal A$, receives $R_h$, and observes
$X_{h+1}\sim P_h(\cdot\mid X_h,A_h)$, with reward function
$r_h(X_h,A_h)$. The terminal space is $\mathcal X_{H+1}=\{\partial\}$.
Rewards are nonnegative and each episode's return is at most one.
For a Markov policy $\pi=(\pi_h)_{h=1}^H$, we define the value functions
\[
 V_h^\pi(x)=\mathbb E_\pi\!\left[\sum_{s=h}^H R_s\mid X_h=x\right],
 \quad Q_h^\pi=r_h+P_hV_{h+1}^\pi,
 \quad V_h^\pi(x)=\sum_a\pi_h(a\mid x)Q_h^\pi(x,a).
\]
Write $\mathcal T_hv=r_h+P_hv$. Optimal values obey
$Q_h^\star=\mathcal T_hV_{h+1}^\star$,
$V_h^\star=\max_aQ_h^\star(\cdot,a)$, and $V_{H+1}^\star=0$.

Let $\mathscr F_{t-1}$ contain the completed history before episode
$t$, including replay draws and parameter updates. For a policy $\pi_t$ conducting round-$t$ episode collection, set
$J(\pi_t)=\mathbb E[\sum_hR_{t,h}\mid\mathscr F_{t-1}]$ and
$J^\star=\mathbb E_{\nu_1}V_1^\star(X_1)$. We study the cumulative regret in our algorithm analysis
\begin{equation}\label{eq:dqn-main-regret-definition}
 \Reg_T=\sum_{t=1}^T\{J^\star-J(\pi_t)\}.
\end{equation}

\subsection{Data Model: Quadratic Signals in Stochastic MDPs}
\label{sec:data-model}

We use a stochastic MDP family whose transition and reward functions include quadratic signals. Inspired
by polynomial feature-learning models \citep{nichani2023provable},
the construction makes the unknown predictive features explicit
while the learner receives raw observations and transitions.
Fix $H,D\ge2$ and $r=D-1\le d(d+1)/2-1$.
Observations lie on a $d$-dimensional sphere $\mathbb S_d=\sqrt d\,\mathbb S^{d-1}$,
with uniform measure $\mu$. At each stage, let
$E_{h,1},\ldots,E_{h,r}$ be unknown trace-free symmetric matrices
orthonormal in Frobenius inner product, spanning the
$r$-dimensional subspace
$\mathcal S_h:=\operatorname{span}{E_{h,1},\ldots,E_{h,r}}$. Define
\[
 g_h(x)=\frac1d(x^\top E_{h,1}x,\ldots,x^\top E_{h,r}x)^\top,
 \qquad \mathbb E_\mu g_h=0,\quad \|g_h(x)\|_2\le1.
\]
The learner observes $x$ but not its unknown quadratic features $g_h(x)$.
Choose unknown contractions $T_{h,a}\in\mathbb R^{r\times r}$,
at least one invertible at each $h<H$, and amplitudes
$\tau_h,\omega_h\in(0,1)$. For the transition kernel over each stage $h\in[H]$, at stage-$H$ the transition leads to a terminal state and
\begin{equation}\label{eq:dqn-main-smooth-kernel}
 P_h(dx'\mid x,a)=
 \left[1+\frac{\tau_h\omega_h}{D}
       (T_{h,a}g_h(x))^\top g_{h+1}(x')\right]\mu(dx'),
 \qquad h<H.
\end{equation}
The density lies in
$[1-\tau_h\omega_h/D,\,1+\tau_h\omega_h/D]$. For rewards, choose $B\in\mathbb R^{D\times r}$ with
$B^\top B=I_r$, $B^\top\mathbf1=0$. Define
\[
 \phi_h^\star(x,a)=\frac1D[\mathbf1+\tau_hBT_{h,a}g_h(x)],
 \qquad
 r_h(x,a)=\phi_h^\star(x,a)^\top\vartheta_h,
 \qquad \vartheta_h\in[0,r_h^{\max}]^D.
\]
The vector $\phi_h^\star$ has nonnegative coordinates summing to one.
Rewards are deterministic, with
$0\le\sum_h r_h(X_h,A_h)\le1$.
Transitions use fresh randomness, and episodes start independently
from $\mu$. For any bounded value continuation $v$,
\[
 P_hv(x,a)=\mathbb E_\mu v+
 \frac{\tau_h\omega_h}{D}(T_{h,a}g_h(x))^\top
                         \mathbb E_\mu[g_{h+1}v].
\]
Thus the conditional mean TD target $r_h+P_hv$ has the form
$c_a+x^\top C_ax$ (Lemma~\ref{lem:qmdp-bellman-signal}): the moment $\mathbb E_\mu[g_{h+1}v]$
determines which current-state quadratics predict the value continuation.
This identity also holds when $v$ is the action maximum of a learned
ReLU critic. It identifies the supervision supplied to the neural
update as that value continuation changes.

\subsection{Neural Critic}
The learner represents Q-functions by a finite two-layer ReLU
network applied to the raw observation $x$. With
$\sigma(z)=\max\{z,0\}$, for stage-$h$, the critic is
\begin{equation}\label{eq:dqn-critic}
 Q_{\theta_h}(x,a)=b_{h,a}
 +\frac1{\sqrt m}\sum_{i=1}^m U_{h,a,i}\sigma(w_{h,i}^\top x),
 \qquad \theta_h=(W_h,U_h,b_h),
\end{equation}
with hidden features
$\varphi_{W_h}(x)=m^{-1/2}(\sigma(w_{h,i}^\top x))_{i=1}^m$.
All parameters are trainable and separate across stages.
Here $b_h=(b_{h,a})_{a\in \mathcal A}$ denotes output biases and $U_h=(U_{h,a})_{a\in\mathcal{A}}$ denotes head layer parameters chosen by an action candidate in the following write-up for convenience. We use $\sigma'(0)=0$ and fixed action
tie-breaking for action choice when multiple actions have the same Q-value. For even $m$ and $i\le m/2$, independently draw $w_{h,i}$
uniformly on $\mathbb S^{d-1}$ and $U_{h,a,i}$ as Rademacher signs,
independently of the environment. Initialize
\[
 w_{h,i+m/2}=w_{h,i},\qquad
 U_{h,a,i+m/2}=-U_{h,a,i},\qquad b_{h,a}=0.
\]
By incorporating this initialization strategy, we can have initial $Q_{\theta_h^{(0)}}=0$ with nonzero hidden-weight Jacobians.

\section{Temporal Difference Learning}\label{sec:lat-algorithm}
We build our analysis on exploration-then-commit
procedure \citep{garivier2016explore} with a temporal difference (TD) objective \citep{Sutton1988}, detailed in Algorithm~\ref{alg:dqn-online}. During exploration, every action is drawn uniformly at
random. Each batch of $n$ fresh episodes supplies one batched TD gradient update
of the entire neural critic, followed immediately by a target copy.
After $K$ such updates, the learner commits once to the final critic's
greedy policy for the remaining episodes.

\begin{algorithm}[t]
\caption{Explore-then-commit temporal difference learning}
\label{alg:dqn-online}
\begin{algorithmic}[1]
\REQUIRE Even network width $m\ge2$; horizon $H\ge2$; episode budget $T$;
collection batch size $n$; number of TD updates $K\ge H$;
replay minibatch size $b$; $Kn\le T$.
\STATE Initialize $\boldsymbol\theta^{(0)}$ by the independent paired initialization
in Section~\ref{sec:lat-preliminaries}; set $\bar{\boldsymbol\theta}^{(0)}=\boldsymbol\theta^{(0)}$.
\label{line:dqn-init}
\STATE \textbf{Phase I: uniformly random exploration and TD updates.}
\FOR{$j=1,\ldots,K$}
 \STATE Set $\mathsf B_j=\varnothing$; hold $\boldsymbol{\theta}^{(j-1)}$ and
 $\bar{\boldsymbol{\theta}}^{(j-1)}$ fixed during collection.
 \FOR{$t=(j-1)n+1,\ldots,jn$}
  \STATE Observe a fresh initial state $X_{t,1}$.
  \FOR{$h=1,\ldots,H$}
   \STATE Draw $A_{t,h}\sim\operatorname{Unif}(\mathcal A)$,
   independently of the current observation and critic.
   \label{line:dqn-action}
   \STATE Execute $A_{t,h}$ and observe $R_{t,h},X_{t,h+1}$.
   \label{line:dqn-transition}
   \STATE Append $(h,X_{t,h},A_{t,h},R_{t,h},X_{t,h+1})$
   to $\mathsf B_j$.
   \label{line:dqn-buffer}
  \ENDFOR
 \ENDFOR
 \STATE Draw $b$ independent uniform record indices from
 $\mathsf B_j$ with replacement, forming $\mathcal I_j$.
 \label{line:dqn-minibatch}
 \STATE Form all detached targets by \eqref{eq:dqn-target} and take one gradient step with \eqref{eq:sgd-td-update}.
 \label{line:dqn-gradient}
 \STATE Copy the whole updated critic: $\bar{\boldsymbol\theta}^{(j)}=\boldsymbol\theta^{(j)}$.
 \label{line:dqn-copy}
\ENDFOR
\STATE \textbf{Phase II: commit to the learned greedy policy.}
\STATE Freeze $\boldsymbol\theta^{(K)}$ and set
$\widehat\pi_h(x)=\argmax_a Q_{\theta_h^{(K)}}(x,a)$.
\label{line:dqn-commit}
\FOR{$t=Kn+1,\ldots,T$}
 \STATE Execute $\widehat\pi$ for one complete episode.
\ENDFOR
\end{algorithmic}
\end{algorithm}

\subsection{Random Exploration and Batch Collection}
At update $j\in [K]$, we collect $n$ fresh episodes, indexed by $(j-1)n+1,\ldots,jn$, under a fixed exploration policy that selects actions uniformly from $\mathcal A$. Episodes are generated independently, both within and across collection batches, whereas transitions at different stages within the same episode may be dependent. The resulting $Hn$ transitions form the collection buffer $\mathsf B_j$. The online and target network parameters remain fixed during collection. Conditional on $\mathsf B_j$, the replay minibatch $\mathcal I_j$ consists of $b$ independent uniform draws with replacement from the buffer. The online network then performs one gradient update using $\mathcal I_j$, followed by synchronization of the target network with the updated online network. Overall, the exploration phase consists of $K$ fresh collection batches, $Kn$ episodes, and exactly $K$ gradient updates.

\subsection{One Simultaneous Neural TD Update Per Collection Batch}
Let $\boldsymbol{\theta}=(\theta_h)_{h=1}^H$.
For a sampled record $i=(h_i,X_i,A_i,R_i,X_i')$, use the target
network copied after the preceding update $\bar{\boldsymbol\theta}^{(j-1)}$ to compute the TD target $Y_i$:
\begin{align}
    Y_i&=R_i+\mathbf1\{h_i<H\}\max_{a'}
            Q_{\bar\theta_{h_{i}+1}^{(j-1)}}(X_i',a').
 \label{eq:dqn-target}
\end{align}
TD objective within the minibatch $\mathcal{I}_j$ is
\begin{align}
 \widehat L_{\mathcal I_j}(\boldsymbol\theta;\bar{\boldsymbol\theta}^{{(j-1)}})
 &=\frac1{2b}\sum_{i\in\mathcal I_j}
                 (Q_{\theta_{h_i}}(X_i,A_i)-Y_i)^2.
 \label{eq:dqn-training-loss}
\end{align}
We then update $\theta$ using constant step size minibatch SGD
\begin{equation}
\label{eq:sgd-td-update}
    \boldsymbol \theta^{(j)} = \boldsymbol \theta^{(j-1)} - \eta\left.\nabla_{\boldsymbol\theta} \widehat L_{\mathcal I_j}(\boldsymbol\theta;\bar{\boldsymbol\theta}^{(j-1)})\right|_{\boldsymbol \theta=\boldsymbol \theta^{(j-1)}}, \qquad j=1,...,K,
\end{equation}
where $\eta$ is the constant step size.
For the simplicity of analysis, we set $\eta=1/2$. 
Network parameters $W,U,\mathbf b$
change simultaneously, and we treat every target $Y_i$ as a constant in differentiation, aligning with standard practice in RL.
Thus all stage critics use the same preceding target snapshot, and
the target network is copied only after the whole gradient step.
If the minibatch contains $b_{h,j}$ \footnote{Here we slightly abuse the notations: $b_{h,j}$ the stage-wise sample count, and $b_{h,a}$ the output bias.} records from stage $h$, its
effective stage learning rate is
$\eta_{h,j}=\tfrac{b_{h,j}}{2b}=\frac{p_{h,j}}2$ with $p_{h,j}:=\tfrac{b_{h,j}}b.
$
These are precisely the stage fractions and steps used in the
multi-update analysis.

\section{Theoretical Analysis}\label{sec:lat-main-results}
We first characterize the quadratic Bellman signal acquired by an
initial informative TD update (Section~\ref{sec:dqn-initial-acquisition}).
We then quantify its predictive benefit through a bounded readout
and a matched comparison with head-only SGD
(Section~\ref{sec:dqn-predictive-benefit}).
Next, we analyze how imperfect learned continuations generate
supervision at earlier stages and track the concurrent refinement
of neural critics under evolving TD targets
(Section~\ref{sec:dqn-coupled-learning}).
Finally, we translate these prediction guarantees into explicit
finite-budget regret bounds (Section~\ref{sec:dqn-finite-budget}).

\subsection{Feature Acquisition at an Initial TD Update}
\label{sec:dqn-initial-acquisition}

We study an initial stage-specific update of
Algorithm~\ref{alg:dqn-online}. Fix a stage $h$ and
an update $j\in[K]$ at which the stage-$h$ critic is
still at its paired initialization, so that
$\theta_h^{(j-1)}=\theta_h^{(0)}$.
Let $\mathsf B_j$ be the buffer collected for this
update and put
\[
I_h:=\{i\in\mathsf B_j:h_i=h\},\qquad |I_h|=n.
\]
Each of the $n$ complete episodes contributes one
record to $I_h$. The global minibatch consists of
$b$ independent uniform draws with replacement
from $\mathsf B_j$.

Let $\mathscr H_{j-1}$ denote the information
available before collection batch $j$, including
the initializations and all completed collection
batches, replay draws, and parameter updates.
The continuation actually used in the detached
TD targets is
\[
\widehat v(x')
=\mathbf1\{h<H\}
\displaystyle\max_{a'\in\mathcal A}
Q_{\bar\theta_{h+1}^{(j-1)}}(x',a').
\]
It may depend on the completed training history,
but is $\mathscr H_{j-1}$-measurable and remains
fixed throughout the current collection batch
and its replay sampling.
Conditional on $\mathscr H_{j-1}$, the stage-$h$
records in $I_h$ are independent and identically
distributed, with state--action law $\nu:=\mu\otimes\operatorname{Unif}(\mathcal A).$
For compatibility with the formulas below, write
$\bar\pi_a:=1/A$, where $A=|\mathcal A|$.

Let $\mathcal V$ be a fixed class of bounded
continuations, and consider updates for which
$\widehat v\in\mathcal V$.
For every fixed $v\in\mathcal V$, write
$Y_i(v)=R_i+v(X_i')$.
The MDP construction gives
\begin{equation}\label{eq:dqn-feature-target-model}
(\mathcal T_hv)(x,a)
=
r_h(x,a)+(P_hv)(x,a)
=
c_a(v)+x^\top C_a(v)x,
\qquad C_a(v)\in\mathcal S_h.
\end{equation}
Since $\widehat v$ is determined before the current
batch,
\[
\mathbb E\!\left[
Y_i(\widehat v)
\mid\mathscr H_{j-1},X_i=x,A_i=a
\right]
=
c_a(\widehat v)+x^\top C_a(\widehat v)x.
\]
However, the quadratic component may vanish.
We therefore require a sufficiently strong quadratic
signal, together with regularity conditions ensuring
that sampling and finite-step errors do not overwhelm it.
\begin{condition}[Boundedness and informative signal]
\label{cond:dqn-feature}
We assume that there exist positive constants
$B_Y,c_*,M_*,S_0$ such that, for every $v\in\mathcal V$,
\[
    |Y_i(v)|\le B_Y,\qquad
    |c_a(v)|\le c_*,\qquad
    \|C_a(v)\|_{\mathrm{op}}\le M_*.
\]
The selected continuation $\widehat v$ satisfies
\[
    \sum_a\bar\pi_a^2\|C_a(\widehat v)\|_F^2
    \ge S_0.
\]
\end{condition}
For $X\sim\mu$, let $X_1$ denote its first coordinate.
Define the spherical moment constants
$\beta_d=\mathbb E|X_1|/2$ and
$\alpha_d=d\beta_d/(d+1)$, whose explicit expressions
are given in \eqref{eq:dqn-feature-moment-constants}. Let $b_h$ denote the realized number of stage-$h$ records in the replay minibatch, and define the effective stage step size $\eta_{\mathrm{eff}}=\eta b_h/b=b_h/(2b)$.

\begin{theorem}[Feature acquisition at an initial TD update]
\label{thm:dqn-feature-gain}
With probability at least $1-\delta$, the following
conclusion holds simultaneously for every
stage--update pair $(h,j)$ of
Algorithm~\ref{alg:dqn-online}
satisfying the initialization condition in
Section~\ref{sec:dqn-initial-acquisition},
Condition~\ref{cond:dqn-feature}, and the quantitative
requirements \eqref{eq:dqn-feature-width},
\eqref{eq:dqn-feature-accuracy-condition}, and
\eqref{eq:dqn-feature-stepsize}. Consider the initial-head readout
\begin{equation}\label{eq:dqn-feature-covariance}
F_W(x,a)=\frac1{\sqrt m}\sum_s U_{a,s}^{(0)}
    \sigma(w_s^\top x).
\end{equation}
The actual hidden parameters after the simultaneous
TD update satisfy
\begin{equation}\label{eq:dqn-feature-main-gain}
\begin{aligned}
&\mathbb E_\nu\!\left[
(x^\top C_a(\widehat v)x)
\{F_{W^+}(x,a)-F_{W_0}(x,a)\}
\right]=\Omega\!\left(
\frac{\eta_{\mathrm{eff}}\alpha_d^2}{d}
\sum_a\bar\pi_a^2\|C_a(\widehat v)\|_F^2
\right).
\end{aligned}
\end{equation}
Here, SGD updates the hidden weights, action heads, and
output biases simultaneously, while
\eqref{eq:dqn-feature-covariance} uses the initial heads
as a fixed readout to measure feature acquisition.
\end{theorem}

\begin{remark}[Feature learning in initial update]
\label{rem:dqn-signal-noise}
Theorem~\ref{thm:dqn-feature-gain} shows that, at any stage $h$ satisfying
Condition~\ref{cond:dqn-feature}, an initial informative TD update increases the
alignment between the hidden features and the quadratic Bellman
signal. The lower bound in the theorem 
$\Omega\!\left(
({\eta_{\mathrm{eff}}\alpha_d^2}/{d})
\sum_a\bar\pi_a^2\|C_a(\widehat v)\|_F^2
\right)$, which is strictly positive under the stated
conditions. Since the output weights $U^{(0)}_{a,j}$ are fixed when evaluating both
representations, this gain certifies that the hidden-weight update
from $W_0$ to $W^+$ produces a readout change positively correlated
with the quadratic component of the Bellman target.
\end{remark}

\subsection{Predictive Benefit of the Acquired Features}
\label{sec:dqn-predictive-benefit}

The preceding alignment gain leaves two prediction questions: how
accurately can the updated features represent the Bellman target, and
how much improvement does the actual SGD readout realize?
Theorem~\ref{thm:dqn-bounded-bridge} answers these
questions through a bounded readout and a matched comparison with a
head-only update. Corollary~\ref{cor:dqn-bounded-replay-certificate}
then places the resulting approximation error in the replay-risk bound.

We use the initial-update and sampling setup of
Section~\ref{sec:dqn-initial-acquisition}, and write $v=\widehat v$
and $Y_i=Y_i(v)$. Assume that $v$ is determined by pre-collection
information $\mathscr G$ independent of the selected stage's paired
initialization. For this fixed continuation, write
\eqref{eq:dqn-feature-target-model} as
\[
 f_a(x):=(\mathcal T_hv)(x,a)=c_a+s_a(x),
 \qquad s_a(x):=x^\top C_ax,
\]
where $c_a=c_a(v)$ and $C_a=C_a(v)$. We also define $\|c\|_{L^2(\nu)}^2=\frac1A\sum_{a\in\mathcal A}c_a^2$. Since a matrix
and its symmetric part define the same quadratic form and the trace component contributes only a constant which can be absorbed into $c_a$, we consider the following structural condition without loss of generality.
\begin{condition}[Bounded targets and centered quadratic structure]
\label{cond:dqn-predictive-step}
The targets and quadratic coefficients satisfy
\[
 |Y_i|\le1,\qquad C_a=C_a^\top,\qquad
 \operatorname{tr}(C_a)=0
 \quad (i\le n,\ a\in\mathcal A).
\]
\end{condition}

To isolate the predictive contribution of hidden-weight movement,
we compare a simultaneous update with a matched head-only step.
Let $(W^+,U^+, b^+)$ denote the parameters after the simultaneous
TD update from $(W_0,U^{(0)}, b^{(0)})$. Define
\[
\begin{aligned}
 Q_{\mathrm{full}}^+(x,a):=Q_{W^+,U^+, b^+}(x,a),\qquad
 Q_{\mathrm{head}}^+(x,a):=Q_{W_0,U^+, b^+}(x,a).
\end{aligned}
\]
The matched step updates only the heads and biases, using the same
initialization, targets, minibatch, and step size. Its updated readout
is also $(U^+, b^+)$ because both steps evaluate all gradients
at the same initial parameters.
To compare these critics, we use the full prediction risk and a
centered risk. The full risk can decrease through better estimation
of the actionwise constants $c_a$; the centered risk subtracts the
mean prediction for each action to isolate prediction error on the
quadratic signal $s_a(x)$:
\[
\begin{aligned}
 \mathcal R_\nu(Q)=\|Q-f\|_{L^2(\nu)}^2,\qquad
 \mathcal R_\nu^{\mathrm{sig}}(Q)=\|Q-\mathbb E_\mu[Q(X,\cdot)]-s\|_{L^2(\nu)}^2.
\end{aligned}
\]
For the finite-sample bounds, write $\varepsilon_{\mathrm{stat}}
=\sqrt{\left[d\log(md)+\log\frac A\delta\right]
\left(\frac1n+\frac1{b_h}+\frac1m\right)}.$

\begin{theorem}[Predictive content of an initial TD update]
\label{thm:dqn-bounded-bridge}
Under the collection setup above and
Condition~\ref{cond:dqn-predictive-step}, take $d,A\ge2$,
$\delta\in(0,1)$, even $m\ge64(d+1)^2$, $b_h>0$, and
$0<\eta_{\mathrm{eff}}\le1$.
With probability at least $1-\delta$, the following hold.
\begin{itemize}[leftmargin=*, nosep]
    \item There exists a readout $U_{a,j}^\dagger:=\frac{Ad^2U_{a,j}^{(0)}}
 {\eta_{\mathrm{eff}}\alpha_d^2(2d+1)},
 \ b_a^\dagger:=-\frac{d^2}{2d+1}c_a.$ such that
$q^\dagger:=Q_{W^+,U^\dagger, b^\dagger}$ satisfies
\begin{equation}\label{eq:dqn-bounded-bridge-approx}
\|q^\dagger-f\|_\infty
=\mathcal O\!\left(\frac{Ad^3}{\alpha_d^2}
\varepsilon_{\mathrm{stat}}\right).
\end{equation}
    \item If the step size satisfies $\frac{\eta_{\mathrm{eff}}}{A}
 \max\!\left\{1+2\beta_d^2,
             \frac{2\alpha_d^2(d+1)}{d^2}\right\}\le\frac12,$
then
\begin{equation}\label{eq:dqn-bounded-bridge-actual-gap}
\begin{aligned}
&\mathcal R_\nu(Q_{\mathrm{head}}^+)
\!-\!\mathcal R_\nu(Q_{\mathrm{full}}^+)\!\ge\!
\Omega\!\left(\frac{\eta_{\mathrm{eff}}}{A}
\left[\beta_d^2\|c\|_{L^2(\nu)}^2
\!+\!\frac{\alpha_d^2}{d}\|s\|_{L^2(\nu)}^2\right]\right)\!-\!\mathcal O\!\left(
\eta_{\mathrm{eff}}d^2\varepsilon_{\mathrm{stat}}
(1+\eta_{\mathrm{eff}}d^2\varepsilon_{\mathrm{stat}})
\right).
\end{aligned}
\end{equation}
The same bound holds for $\mathcal R_\nu^{\mathrm{sig}}$ with the
$c$-term omitted. 

Moreover, under the respective lower bounds on the record
count $n$, replay batch size $b_h$, and width $m$ in
Definition~\ref{def:dqn-step-accuracy}, the gains are strictly positive
at the displayed leading orders.
\end{itemize}
\end{theorem}

\begin{remark}[Predictive content and realized improvement]
\label{rem:dqn-predictive-response}
The bounded witness certifies the approximation available in $W^+$,
with uniform error
$\widetilde{\mathcal O}(n^{-1/2}+b_h^{-1/2}+m^{-1/2})$
at fixed $d,A$. The matched-update comparison quantifies the
improvement already realized by simultaneous SGD. Both critics use
the same initialization, target, minibatch, and head-and-bias update,
so their risk difference isolates the contribution of hidden-weight
movement. Define the quadratic response coefficient
$
 \lambda_2={2\alpha_d^2(d+1)}/{d^2}.
$
Lemma~\ref{lem:dqn-quadratic-eigenvalues} gives the leading
population quadratic responses of the full and head-only updates as
$
 {\eta_{\mathrm{eff}}}\lambda_2s/{A}$ and 
 ${\eta_{\mathrm{eff}}}
       {\lambda_2}s/({2{A}(d+1)})
$
respectively. Their ratio is $2(d+1)$, and the finite-sample remainder
bound yields a strict risk improvement at the stated resolution.
The same coefficient governs the subsequent refinement of quadratic
action contrasts in Theorem~\ref{thm:dqn-horizon-complete}.
\end{remark}

Theorem~\ref{thm:dqn-bounded-bridge} bounds the approximation error
achievable with $W^+$ and compares the full update with a matched
head-only update. The following corollary combines this approximation
guarantee with a uniform replay bound to control the actual critic's
prediction error through the empirical fit gap, approximation error,
and statistical error. For the $n$ stage-$h$ records indexed by $I$,
define the empirical squared TD loss $\widehat L_I(q;v):=\frac1n\sum_{i\in I} [q(X_i,A_i)-Y_i]^2$.
\begin{corollary}[Initial-update features in the replay-risk bound]
\label{cor:dqn-bounded-replay-certificate}
In the setting of Theorem~\ref{thm:dqn-bounded-bridge}, let
$\mathcal Q_h(W^+)\subseteq\mathcal Q_h$ contain
$q^\dagger$ and $Q_{\mathrm{full}}^+$, and take
$v\in\mathcal V_{h+1}$. Assume the deterministic classes
$\mathcal Q_h,\mathcal V_{h+1}$ satisfy the boundedness and covering
conditions of Lemma~\ref{lem:dqn-uniform-replay}, with $B=B_Q\ge1$
and covering bounds $N_Q(u),N_V(u)$.
On the intersection of the events in
Theorem~\ref{thm:dqn-bounded-bridge} and
Lemma~\ref{lem:dqn-uniform-replay}, the frozen-target approximation
error obeys
\begin{equation}\label{eq:dqn-bounded-replay-approx}
\begin{aligned}
 R_{app}:=\inf_{q\in\mathcal Q_h(W^+)}
           \|q-\mathcal T_hv\|_{L^2(\nu)}^2=\mathcal{O}\!\left(
 \frac{A^2d^6}{\alpha_d^4}
 \left[d\log(md)+\log\frac A\delta\right]
 \left(\frac1n+\frac1{b_h}+\frac1m\right)\right).
\end{aligned}
\end{equation}
Moreover, the actual updated critic satisfies
\begin{equation}\label{eq:dqn-bounded-replay-risk}
\begin{aligned}
 \|Q_{\mathrm{full}}^+-\mathcal T_hv\|_{L^2(\nu)}^2
 =\mathcal{O}\!\left(R_{fit}
       +R_{app}+R_{est}\right).
\end{aligned}
\end{equation}
where $R_{fit}=\widehat L_I(Q_{\mathrm{full}}^+;v)-\inf_{q\in\mathcal Q_h(W^+)}\widehat L_I(q;v)$ and $R_{est}=\frac{B_Q^2}{n}\log\!\frac{eHT^2N_Q(u)N_V(u)}{\delta}+B_Qu$.
Substituting \eqref{eq:dqn-bounded-replay-approx} gives the explicit
width and sample contribution to the actual critic's risk.
\end{corollary}
\begin{remark}[From representation quality to critic risk]
The bound in \eqref{eq:dqn-bounded-replay-risk} separates three sources
of prediction error. The empirical fit gap measures how well the
actual SGD readout fits the replay data relative to the best readout
on $W^+$. The approximation term $R_{\mathrm{app}}$ measures the best
prediction accuracy attainable within this readout class, with bound
$\widetilde{\mathcal O}(n^{-1}+b_h^{-1}+m^{-1})$ at fixed $d,A$.
The statistical terms transfer empirical loss to population risk;
they depend on the record count $n$, prediction bound $B_Q$, and
class complexity through $N_Q(u),N_V(u)$ and the covering scale $u$.
Thus better features must be accompanied by effective fitting and
controlled complexity to yield a small prediction error.
Increasing $b_h$ reduces the replay-sampling term in the approximation
bound, but does not increase the number of collected records
available for statistical estimation.
\end{remark}

\subsection{Feature Evolution in Multiple TD Updates with Imperfect Targets}
\label{sec:dqn-coupled-learning}
The initial-update results characterize learning from an informative
value continuation target. We now show how Algorithm~\ref{alg:dqn-online}
creates such a target through its own first terminal update and follow
the actual neural parameters as the copied value continuation continues to
change. Lemma~\ref{lem:dqn-horizon-feature-gain} identifies the
initial acquisition at newly reached stages, and
Theorem~\ref{thm:dqn-horizon-complete} controls prediction along
the subsequent coupled trajectory. The analysis specializes the family in
Section~\ref{sec:data-model} to signed actions and terminal rewards.
All reached stages continue training with one SGD step per target
copy, retaining their values' current estimation errors.

\paragraph{Specific Construction.}
For Theorem~\ref{thm:dqn-horizon-complete}, we further specialize the MDP defined in Section~\ref{sec:data-model}, take $d\ge4$,
$\mathcal A=\{-1,+1\}$, $T_{h,a}=aI_r$, and
$\tau_h=\omega_h=1/2$, giving
\begin{equation}\label{eq:dqn-horizon-data}
\begin{aligned}
 r_h(x,a)&=0 \quad(h<H),\quad
 r_H(x,a)=\frac12+\frac{a\,g_{H,1}(x)}{8D},\\
 P_h(dx'\mid x,a)
 &=\left[1+\frac{a}{4D}g_h(x)^\top g_{h+1}(x')\right]\mu(dx')
 \quad(h<H).
\end{aligned}
\end{equation}
This specializes the general family with
$r_h^{\max}=0$, $\vartheta_h=0$ for $h<H$, and
$r_H^{\max}=1$, $\vartheta_H=\mathbf1/2+Be_1/4$,
where $e_1$ is the first coordinate vector. For $h\ge2$, choose unknown unit vectors $u_h$ and set
\begin{equation}\label{eq:dqn-axial-downstream-model}
 E_{h,1}=\frac{d u_hu_h^\top-I_d}{\sqrt{d(d-1)}},
 \qquad
 g_{h,1}(x)=\frac{(u_h^\top x)^2-1}{\sqrt{d(d-1)}}.
\end{equation}

At stage one, $E_{1,1}$ may be any unit trace-free symmetric matrix.
Complete each stage's basis with $D-2$ trace-free orthonormal matrices.
Lemma~\ref{lem:dqn-axial-stage-projection} gives the optimal values
and confirms minimal common dimension $D$.

\begin{figure}[t]
\centering
\resizebox{\linewidth}{!}{%
\definecolor{TDcoarse}{HTML}{BE7826}
\definecolor{TDrefined}{HTML}{247D8D}

\begin{tikzpicture}[
  x=1cm, y=1cm, font=\small,
  >={Stealth[length=2mm]},
  state/.style={
      circle,
      draw=black!60,
      fill=white,
      minimum size=11mm,
      text width=10mm,
      text height=1.5ex,
      text depth=0.5ex,
      align=center,
      inner sep=0pt,
      outer sep=0pt
    },
    critic/.style={
      rectangle,
      rounded corners=2pt,
      minimum width=18mm,
      minimum height=6mm,
      text width=18mm,
      text height=1.5ex,
      text depth=0.5ex,
      align=center,
      inner sep=0pt,
      outer sep=0pt
    },
  coarse/.style={critic, draw=TDcoarse, dashed,
                 fill=TDcoarse!10},
  refined/.style={critic, draw=TDrefined,
                  fill=TDrefined!12},
  forward/.style={->, draw=black!60, line width=.6pt},
  update/.style={->, draw=black!35, line width=.65pt},
  early/.style={->, draw=TDcoarse, dashed, line width=.95pt},
  late/.style={->, draw=TDrefined, line width=.95pt},
  note/.style={anchor=west, align=left, font=\footnotesize}
]

\node[font=\small\bfseries] at (3.6,.85)
  {Forward interaction, backward learning};
\foreach \x/\h/\name in {%
  0/1/one,%
  3.2/{H-2}/hmTwo,%
  5.2/{H-1}/hmOne,%
  7.2/H/last%
}
  \node[state] (s-\name) at (\x,0) {$X_{\h}$};
\node (s-dots) at (1.6,0) {$\cdots$};
\node (terminal) at (8.65,0) {$\partial$};
\draw[forward] (s-one) -- node[above,font=\footnotesize] {$P_1$} (s-dots);
\draw[forward] (s-dots) -- (s-hmTwo);
\draw[forward] (s-hmTwo) -- node[above,font=\footnotesize] {$P_{H-2}$} (s-hmOne);
\draw[forward] (s-hmOne) -- node[above,font=\footnotesize] {$P_{H-1}$} (s-last);
\draw[forward] (s-last) -- (terminal);
\node[font=\footnotesize,text=black!60] at (2.4,-.75)
  {$r_h=0\quad(h<H)$};

\draw[forward] (-2.20,-1.12) -- (-2.20,-5.7)
  node[midway,above,rotate=90,font=\footnotesize] {Global update $j$};
\foreach \y/\j in {-1.45/1,-2.65/2,-4.25/{K-1},-5.45/K}
  \node[anchor=east] at (-.82,\y) {$\j$};

\foreach \x in {0,3.2,5.2}
  \node[text=black!35] at (\x,-1.45) {$0$};
\foreach \x in {0,3.2}
  \node[text=black!35] at (\x,-2.65) {$0$};
\node[coarse] (qHone) at (7.2,-1.45) {$Q_{H,1}$};
\node[coarse] (qHmTwo) at (5.2,-2.65) {$Q_{H-1,2}$};
\node[coarse,fill=TDcoarse!20] (qHtwo) at (7.2,-2.65) {$Q_{H,2}$};

\draw[forward] (s-last.south) --
  node[right,font=\footnotesize] {$r_H$} (qHone.north);
\draw[early] (qHone.south west) -- (qHmTwo.north east);
\draw[update] (qHone.south) -- (qHtwo.north);

\foreach \y in {-1.45,-2.65,-4.25,-5.45}
  \node[text=black!45] at (1.6,\y) {$\cdots$};
\foreach \x in {0,3.2,5.2,7.2}
  \node[text=black!45] at (\x,-3.45) {$\vdots$};

\foreach \x/\h/\name in {%
  0/1/one,%
  3.2/{H-2}/hmTwo,%
  5.2/{H-1}/hmOne,%
  7.2/H/last%
}{
  \node[refined] (prev-\name) at (\x,-4.25) {$Q_{\h,K-1}$};
  \node[refined,fill=TDrefined!22] (next-\name) at (\x,-5.45) {$Q_{\h,K}$};
  \draw[update] (prev-\name.south) -- (next-\name.north);
}
\draw[late] (prev-last.south west) -- (next-hmOne.north east);
\draw[late] (prev-hmOne.south west) -- (next-hmTwo.north east);

\node[note,text=TDcoarse] at (8.2,-1.45)
  {Inaccurate,\\but useful};
\node[note,text=black!60] at (8.2,-2.65)
  {Already reached stages\\keep learning};
\node[note,text=TDrefined] at (8.2,-4.85)
  {Refined\\supervision};

\draw[late] (.4,-6.04) -- (0,-6.44);
\node[note] at (.6,-6.24) {Copied continuation};
\draw[update] (4.8,-6.02) -- (4.8,-6.46);
\node[note] at (5.02,-6.24) {Continued SGD};
\node[note,text=black!60] at (8.2,-6.24) {$0$: not yet reached};
\end{tikzpicture}
}
\vspace{-1em}
\caption{Backward propagation and concurrent refinement of TD supervision.
Initially inaccurate but useful estimates supervise earlier stages
while reached critics continue learning. Colors are schematic.}
\label{fig:td-backward-refinement}
\vspace{-2em}
\end{figure}
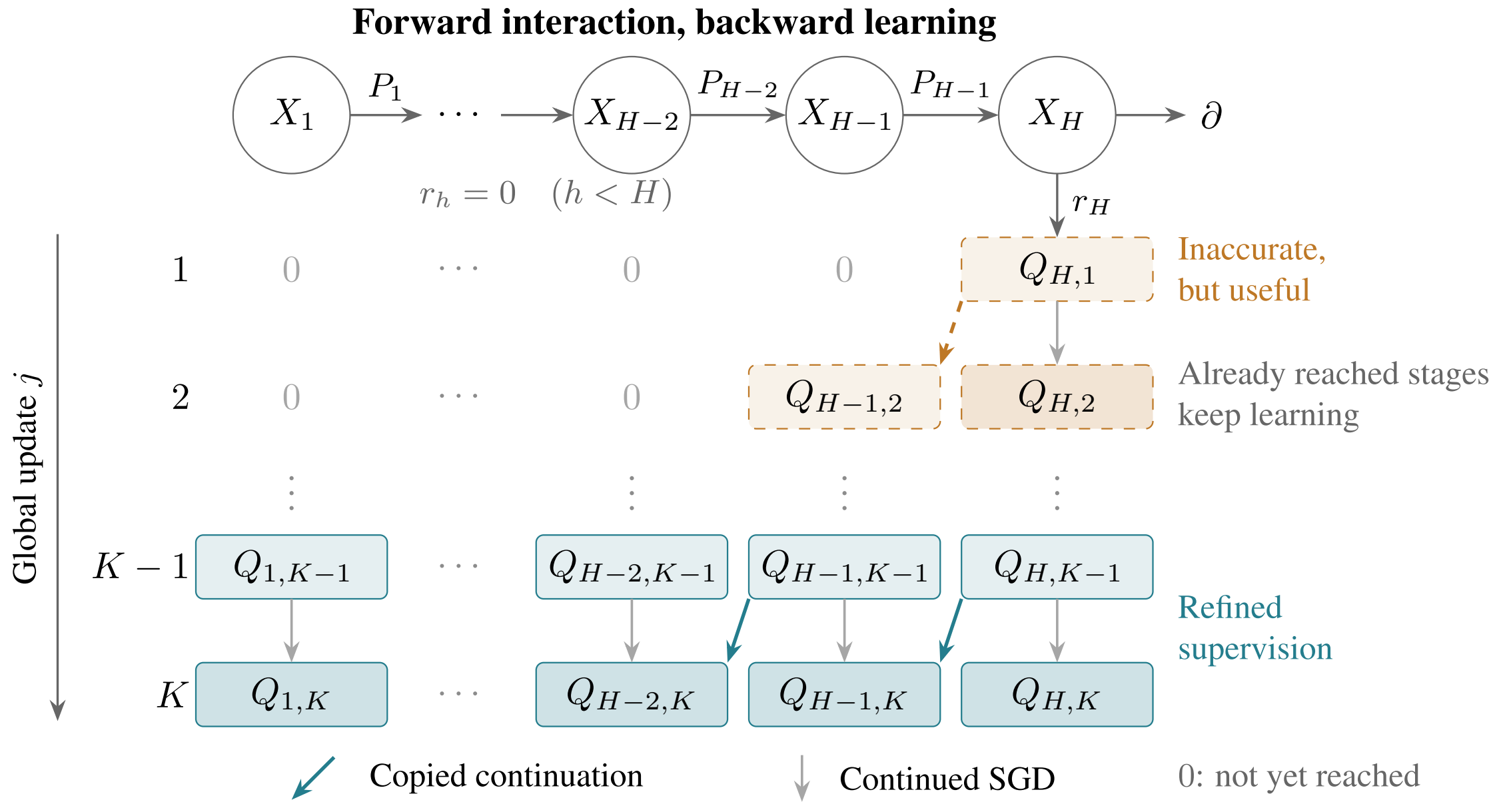

\paragraph{Backward propagation and refinement of TD supervision.}
Figure~\ref{fig:td-backward-refinement} illustrates learning with the constructed MDP in
\eqref{eq:dqn-horizon-data} with SGD.
Stage $h$ is first potentially reached at $j_h=H-h+1$, while
already reached critics continue to refine concurrently.
Although uniform collection under $\pi_h^{\mathrm{rnd}}(a\mid x)=1/2$
gives $Q_h^{\pi^{\mathrm{rnd}}}(x,a)=1/2$ for $h<H$,
greedy bootstrapping can create a quadratic signal.
For the copied continuation
$\widehat v_{h,j}(x'):=\max_{a'\in\{-1,+1\}}
Q_{\bar\theta_{h+1}^{j-1}}(x',a')$,
the conditional TD mean is
\[
 P_h\widehat v_{h,j}(x,a)
 =\mathbb E_\mu[\widehat v_{h,j}]
 +\frac{a}{4D}g_h(x)^\top
       \mathbb E_\mu[g_{h+1}\widehat v_{h,j}].
\]
Thus correlation with $g_{h+1}$ supplies current-state quadratic
supervision despite zero immediate reward.
For example, action values $c\pm\beta g_{h+1,1}(x')$ with $\beta>0$
have maximum $c+\beta|g_{h+1,1}(x')|$, yielding the quadratic term
$
 \frac{a\beta}{4D}
 \mathbb E_\mu[g_{h+1,1}|g_{h+1,1}|]\,g_{h,1}(x)$ with
 $\mathbb E_\mu[g_{h+1,1}|g_{h+1,1}|]>0$
by Lemma~\ref{lem:dqn-axial-stage-projection}.
Useful supervision can therefore precede accurate prediction.
The analysis below characterizes the signal propagation and concurrent refinement under the following condition.

\begin{condition}[The finite-horizon learning configuration]
\label{cond:dqn-horizon}
Use the horizon-$H$ model \eqref{eq:dqn-horizon-data} and the
exploration-then-commit Algorithm~\ref{alg:dqn-online}, with
$H\ge2$, positive integers $n,b,K$ satisfying $K\ge H$,
$L=1$, and even $m\ge2$. Write
$\zeta_d=\E_\mu|g_{h,1}|$ and
$\kappa_d=\E_\mu[g_{h,1}|g_{h,1}|]$ for any $h\ge2$.
Lemma~\ref{lem:dqn-axial-stage-projection} gives
$1/(8d^2)\le\kappa_d\le2/[d(d+2)]$ and the optimal values
\begin{equation}\label{eq:dqn-horizon-E}
\begin{aligned}
 \beta_h^\star&=\frac1{8D}
          \left(\frac{\kappa_d}{4D}\right)^{H-h},\quad
 c_h^\star=\frac12+\zeta_d\sum_{\ell=h+1}^H\beta_\ell^\star,\quad
 Q_h^\star(x,a)=c_h^\star+\beta_h^\star a g_{h,1}(x),
 \quad h\in[H].
\end{aligned}
\end{equation}
\end{condition}

\begin{theorem}[Prediction after multiple coupled TD updates]
\label{thm:dqn-horizon-complete}
Under Condition~\ref{cond:dqn-horizon}, assume $Kn\le T$ and
the training budget in Definition~\ref{def:dqn-reachable-budget}.
Let $Q_{h,j}$ denote the actual stage-$h$ critic after positive
SGD step $j$. With probability at least $1-\delta$, there are
action-independent offsets $|\gamma_h|=\mathcal O(1)$ such that
\begin{equation}\label{eq:dqn-horizon-calibrated-readout}
 \widetilde Q_h(x,a)=Q_{h,K}(x,a)+\gamma_h,
 \qquad
 \max_h\|\widetilde Q_h-Q_h^\star\|_\infty\le e,
\end{equation}
where
\begin{equation}\label{eq:dqn-horizon-main-error}
\begin{aligned}
 e&:=\exp\!\left(-\frac{\lambda_2K}{16H}\right)
 +\mathcal O\!\left(
    \frac{(d+1)K}{H}\sqrt{\frac{(d+1)K\ell}{H}}
    \left[\frac1{\sqrt n}+\sqrt{\frac Hb}
       +\frac{(d+1)(1+K/H)}{\sqrt m}\right]\right).
\end{aligned}
\end{equation}
Each squared prediction risk of
$\widetilde Q$ under every test distribution is at most $e^2$.
The error bound follows from the coupled recursion of all $h\in[H]$ under the changing targets in
\eqref{eq:dqn-horizon-actual-targets}. Here $\ell$ is the logarithm specified in
Definition~\ref{def:dqn-reachable-budget}, satisfying
$\ell=\mathcal O(d\log(md)+\log(HK/\delta))$.
\end{theorem}

\begin{remark}[From quadratic response to coupled refinement]
The bound in \eqref{eq:dqn-horizon-main-error} separates an
exponentially decaying optimization term from accumulated collection,
replay, and finite-width errors. The replay term $\sqrt{H/b}$ reflects
the approximately $b/H$ replay draws allocated to each stage.
When the network is sufficiently wide, increasing $b$ to order $Hn$
reduces replay error to the collection scale $n^{-1/2}$; further
replay does not improve the order of the prediction-error bound at
fixed $n,m,K$. Increasing the number of global SGD updates $K$
reduces the optimization term but enlarges the accumulated-error
bound, creating a trade-off between optimization progress and error
accumulation.

\noindent The exponential term $\exp(-\lambda_2K/(16H))$ captures
the refinement achieved through these coupled updates. Supervision
starts at stage $H$ and propagates backward through the copied
targets. Earlier stages initially learn from imperfect downstream
estimates. As all reached stages continue training, improvements
in these estimates refine the targets supplied to earlier stages.
The coupled recursion \eqref{eq:dqn-long-contrast-recursion}
quantifies this process, yielding exponential decay up to the
accumulated collection, replay, and finite-width error bound.

\end{remark}

\subsection{Regret Analysis}
\label{sec:dqn-finite-budget}
The preceding theorem separates the optimization transient from
cumulated regret. We now further provide the regret and sampling complexity analysis for Algorithm~\ref{alg:dqn-online} in the following corollary:

\begin{corollary}[Regret analysis]
\label{cor:dqn-complete-rates}
Let $\Gamma_T=\log\!\frac{H(d+1)(T+1)}{\delta}$ and choose the number of SGD steps as $K=\Theta(H\lambda_2^{-1}\Gamma_T)$, collection batch size
$n=\Theta\!\left((d+1)^{5/3}\lambda_2^{-1/3}
\Gamma_T^{2/3}(T/H)^{2/3}\right)$,
update batch size $b=Hn$, and
network width $m=\Theta\!\left((d+1)^2(1+K/H)^2n\right)$,
with sufficiently large universal constants, integer $K,n$,
and even $m$.
Run Algorithm~\ref{alg:dqn-online} with one SGD step
and target copy per collection batch, followed by greedy commitment.
If $Kn\le T$ and Definition~\ref{def:dqn-reachable-budget} holds,
then with probability at least $1-\delta$,
\begin{equation}\label{eq:dqn-complete-two-thirds-rate}
\Reg_T=\mathcal O\!\left(
\min\left\{T,\,
(d+1)^{5/3}\lambda_2^{-4/3}\Gamma_T^{5/3}
H^{1/3}T^{2/3}\right\}\right).
\end{equation}
Since $\lambda_2^{-1}=\mathcal O(d+1)$, this gives
$
\Reg_T=\widetilde{\mathcal O}(d^3H^{1/3}T^{2/3})$,
$\quad K=\widetilde{\mathcal O}(Hd)$, and
$\quad m=\widetilde{\mathcal O}(d^6(T/H)^{2/3})$.
A sufficient budget condition is
$T/H=\Omega((d+1)^{7/2}\lambda_2^{-7}\Gamma_T^8)$;
otherwise, $\Reg_T\le T$.
The same event guarantees the first-update feature conclusions
of Lemma~\ref{lem:dqn-horizon-feature-gain}, including positive
covariance gains under its signal-resolution conditions.
\end{corollary}
\begin{remark}[Policy performance under regret]
Balancing $Kn$ exploration episodes against the committed policy's
error $\mathcal O(\sqrt d\,Te)$, with
$K=\widetilde{\mathcal O}(Hd)$, yields
$\Reg_T=\widetilde{\mathcal O}(d^3H^{1/3}T^{2/3})$
and width $m=\widetilde{\mathcal O}(d^6(T/H)^{2/3})$
on our constructed signed-action MDP family.
For comparison, \citet{dqn_liu2022} obtain
$\widetilde{\mathcal O}(T^{2/3})$ for smooth Besov classes $\epsilon$-greedy exploration
and $\widetilde{\mathcal O}(T^{5/6})$ for two-layer Barron networks.
An explore-then-commit reduction of the reward-free exploration
guarantee of \citet{jin2020reward} gives
$\widetilde{\mathcal O}\!\left(
H^{5/3}(S^2A)^{1/3}T^{2/3}
+H^{7/2}S^2\sqrt{AT}\right)$
for tabular MDPs with $S$ states and $A$ actions.
These guarantees share the same rate $\mathcal{\tilde O}(T^{2/3})$ but differ in function classes and exploration
assumptions: ours specifically tracks finite ReLU critic SGD on a structured MDP, while their results are derived in general tabular MDPs without accounting gradient optimization.
\end{remark}

\section{Conclusion}
We provide a finite-sample characterization of how bootstrapped TD
supervision drives predictive feature learning in finite ReLU critics.
We show that an initial informative update acquires the quadratic
Bellman signal, with hidden-weight movement yielding a strict
prediction-risk improvement over matched head-only SGD.
On a constructed signed-action MDP family, even an imperfect learned
continuation generates informative supervision at earlier zero-reward
stages. By tracking coupled neural updates with evolving targets,
we establish geometric refinement of action contrasts, up to
sampling and finite-width errors, and derive a finite-budget
regret guarantee.
Our analysis connects initial feature acquisition to multi-stage
neural TD dynamics and policy performance without requiring
intermediate Bellman regressions to be solved.

\label{sec:alt-main-end}
\FloatBarrier

\bibliography{reference}
\clearpage
\appendix

\section{Proof Sketch of Theorem~\ref{thm:dqn-horizon-complete}}
\label{sec:lat-proof-sketch}

In this section, we provide the proof sketch of our key technical result Theorem~\ref{thm:dqn-horizon-complete}. Use Condition~\ref{cond:dqn-horizon}, $Kn\le T$, and the budget in
Definition~\ref{def:dqn-reachable-budget}.
Work on the common event of probability at least $1-\delta$ in
Lemma~\ref{lem:dqn-reachable-budget}, with $\ell$ and
$E_{\mathrm{run}}$ defined there.
Write $Q_{h,j}$ for the actual stage-$h$ critic after update $j$.

\begin{lemma}[Initial acquisition; adaptation of
Theorem~\ref{thm:dqn-feature-gain}]
\label{lem:dqn-horizon-feature-gain}
At the first reached update $j_h=H-h+1$, stage $h$ remains at its
paired initialization. Let $S_h$ be the actual target signal in
\eqref{eq:dqn-horizon-signal} and $\mathcal G_h$ the initial-head
covariance gain in \eqref{eq:dqn-feature-main-gain}.
Simultaneously for every stage,
\begin{equation}\label{eq:dqn-horizon-main-gain}
 \mathcal G_h\ge
 \Omega\!\left(\frac{\alpha_d^2S_h}{Hd}\right)
 -\mathcal O\!\left(
 \frac{d^2\ell}{H}\left(\frac1n+\frac Hb\right)\right).
\end{equation}
Under \eqref{eq:dqn-feature-signal-resolution}, the gain is
$\Omega(\alpha_d^2S_h/(Hd))>0$.
The sampling bounds \eqref{eq:dqn-terminal-feature-budget} and
\eqref{eq:dqn-endstage-feature-budget} give positive gains at
stages $H$ and $H-1$, respectively. The latter uses $Q_{H,1}$
while the terminal critic continues training.
\end{lemma}

The signed target has a common offset and opposite quadratic
components, and the initial Rademacher heads cancel the offset in
the covariance calculation. The proof is constructed by adapting the proof of
Theorem~\ref{thm:dqn-feature-gain} to the specific construction and conditions for Theorem~\ref{thm:dqn-horizon-complete}. The details are established in
Lemma~\ref{lem:dqn-reachable-budget}.

\paragraph{Coupled prediction.}
Compare $Q_{h,j}$ with the population reference
$q_{h,j}(x,a)=c_{h,j}+a\,u_{h,j}(x)$ driven by the same conditional
targets. Lemma~\ref{lem:dqn-horizon-iterates} controls sampling,
parameter movement, and ReLU activation changes, giving
$\max_{h,j}\|Q_{h,j}-q_{h,j}\|_\infty\le E_{\mathrm{run}}$.
The initial population tangent operator has quadratic eigenvalue
$\lambda_2/2$. With $\eta_{h,j}=b_{h,j}/(2b)$, for $h<H$,
\[
\begin{aligned}
 \|u_{h,j}-\beta_h^\star g_{h,1}\|_\infty
 &\le\left(1-\frac{\eta_{h,j}\lambda_2}{2}\right)
       \|u_{h,j-1}-\beta_h^\star g_{h,1}\|_\infty\\
 &\quad+\frac{\eta_{h,j}\lambda_2}{8D}
       \left(\|u_{h+1,j-1}-\beta_{h+1}^\star g_{h+1,1}\|_\infty
                    +E_{\mathrm{run}}\right).
\end{aligned}
\]
The terminal contrast contracts toward its exact target.
Common continuation offsets vanish from the signed projection,
and $1/(4D)$ attenuates the downstream error.
Since $\eta_{h,j}\ge1/(4H)$ and $D\ge2$, the maximum contrast
error contracts at rate $\Theta(\lambda_2/H)$, up to the floor
$E_{\mathrm{run}}/7$, while all reached stages continue training.

Set $\gamma_h=c_h^\star-c_{h,K}$.
The tracking lemma gives $|\gamma_h|<2$; combining tracking
with the contrast recursion yields
\[
 \max_h\|Q_{h,K}+\gamma_h-Q_h^\star\|_\infty
 \le \exp\!\left(-\frac{\lambda_2K}{16H}\right)
       +2E_{\mathrm{run}}.
\]
Expanding $E_{\mathrm{run}}$ gives
\eqref{eq:dqn-horizon-main-error}; squaring gives the prediction-risk
bound. The common shifts preserve greedy actions.
The policy and regret consequences are established in
Corollary~\ref{cor:dqn-complete-rates}.

\section{Formal Proofs}\label{sec:lat-formal-proofs}

\subsection{Technical lemmas}\label{sec:lat-technical-lemmas}
\subsubsection{Mixture Construction and Learner Information}\label{sec:dqn-mixture-details}
Fix a horizon $H\ge2$, a finite action set $\mathcal A$ with
$|\mathcal A|\ge2$, and a representation dimension $D\ge2$.
At each decision stage the agent observes
$x\in\mathbb S_d:=\{x\in\mathbb R^d:\|x\|_2=\sqrt d\}$, where
$\mu$ denotes the uniform probability measure and
\stepcounter{equation}
\begin{equation*}
 d\ge2,\qquad r:=D-1\le d(d+1)/2-1.
\end{equation*}
For each $h\in[H]$, choose unknown matrices
$E_{h,1},\ldots,E_{h,r}\in\mathbb R^{d\times d}$ and define
\begin{align}
 E_{h,\ell}&=E_{h,\ell}^{\top},\quad
 \operatorname{tr}(E_{h,\ell})=0,\quad
 \operatorname{tr}(E_{h,\ell}E_{h,k})=\mathbf1\{\ell=k\},
 \label{eq:qmdp-basis}\\
 g_h(x)&=d^{-1}(x^{\top}E_{h,1}x,\ldots,
                    x^{\top}E_{h,r}x)^{\top}.
 \label{eq:qmdp-coordinates}
\end{align}
The matrices select the unknown $r$-dimensional quadratic observation
subspace at stage $h$.

Choose $B\in\mathbb R^{D\times r}$, with rows $B_{j,:}$, satisfying
\begin{equation}\label{eq:qmdp-simplex-basis}
 B^{\top}B=I_r,\qquad B^{\top}\mathbf1=0,\qquad
 BB^{\top}=I_D-\mathbf1\mathbf1^{\top}/D.
\end{equation}
For each $(h,a)$, let $T_{h,a}\in\mathbb R^{r\times r}$ be an unknown
contraction. At every $h<H$, at least one action has invertible
$T_{h,a}$, and assume
\stepcounter{equation}
\begin{equation*}
 \|T_{h,a}\|_{\mathrm{op}}\le1.
\end{equation*}
With unknown amplitudes $\tau_h\in(0,1)$ for $h\in[H]$ and
$\omega_h\in(0,1)$ for $h<H$, set
\begin{align}
 \phi_{h,j}^{\star}(x,a)
   &=D^{-1}\bigl[1+\tau_h B_{j,:}[T_{h,a}g_h(x)]\bigr],
       &&j\in[D],\label{eq:qmdp-simplex-feature}\\
 \nu_{h,j}(dx')
   &=\bigl[1+\omega_h B_{j,:}g_{h+1}(x')\bigr]\mu(dx'),
       &&h<H,\label{eq:qmdp-emission}\\ P_h(dx'\mid x,a)
   &=\sum_{j=1}^D\phi_{h,j}^{\star}(x,a)\nu_{h,j}(dx'),
       &&h<H.\notag\stepcounter{equation}
\end{align}
The environment samples an unobserved mixture component and then
a value from its smooth emission law. Lemma~\ref{lem:qmdp-exact-rank}
establishes positivity, normalization, and exact transition rank $D$.

Each episode starts with a fresh draw $X_1\sim\mu$, independent of
the preceding history. Fix stage reward bounds $r_h^{\max}\ge0$
with $\sum_{h=1}^H r_h^{\max}\le1$ and unknown parameters
$\vartheta_h\in[0,r_h^{\max}]^D$. Rewards are deterministic given
the current state and action:
\begin{equation}\label{eq:qmdp-reward}
 r_h(x,a)=\phi_h^\star(x,a)^\top\vartheta_h,
 \qquad
 R_h=r_h(X_h,A_h).
\end{equation}
Since $\phi_h^\star(x,a)$ is a probability vector,
$0\le R_h\le r_h^{\max}$, so every episode's return lies in $[0,1]$.
Setting $\vartheta_h=0$ gives zero reward at stage $h$.
Nonterminal transitions use fresh randomness. After the stage-$H$
reward, the process transitions deterministically to the terminal
state $\partial$, and the episode ends.

\paragraph{Learner information.}
The learner's data consist of $(h,X_h,A_h,R_h,X_{h+1})$.
It is configured by the ReLU architecture, initialization, replay
capacity, learning rates, target-copy period, and exploration schedule.
The environment matrices $E_{h,\ell}$ and $T_{h,a}$, mixture components,
and features $\phi_h^\star$ are unknown. The critic receives raw
observations. Lemma~\ref{lem:qmdp-exact-rank} gives minimal linear
dimension $D$, independently of neural width. The model also admits
a fixed quadratic state--action expansion with
$1+A[d(d+1)/2-1]$ coordinates.

\subsubsection{The Stochastic Model and Bellman Approximation}
\begin{lemma}[Spherical moments and bounded coordinates]
\label{lem:qmdp-sphere-moments}
For $X\sim\mu$, the representation in
\eqref{eq:qmdp-coordinates} satisfies
\begin{equation}\label{eq:qmdp-coordinate-moments}
\begin{aligned}
 \|g_h(x)\|_2&\le\sqrt{1-1/d}<1,&
 \mathbb E_\mu g_h(X)&=0,\\
 \mathbb E_\mu[g_h(X)g_h(X)^{\top}]&=\frac{2}{d(d+2)}I_r.&&
\end{aligned}
\end{equation}
Every coordinate of $g_h$ is a smooth quadratic polynomial on the
sphere and is $2$-Lipschitz there with respect to $\|\cdot\|_\infty$.
The whole vector satisfies
$\|g_h(x)-g_h(y)\|_2\le2\|x-y\|_\infty$.
\end{lemma}

\begin{lemma}[Smooth transitions and exact linear dimension]
\label{lem:qmdp-exact-rank}
Equations~\eqref{eq:qmdp-simplex-feature}--\eqref{eq:qmdp-reward}
define a valid episodic MDP.  For $h<H$, its kernel has density
\begin{equation}\label{eq:qmdp-density}
 \frac{dP_h(\cdot\mid x,a)}{d\mu}(x')
 =1+\frac{\tau_h\omega_h}{D} [T_{h,a}g_h(x)]^{\top}g_{h+1}(x').
\end{equation}
This density is smooth and belongs to $[1-\frac{\tau_h\omega_h}{D},1+\frac{\tau_h\omega_h}{D}]$.
At each nonterminal stage, the transition operator has rank exactly
$D$.  The features $\phi_h^\star\in\mathbb R^D$ satisfy the standard
normalization
\stepcounter{equation}
\begin{equation*}
 \|\phi_h^\star(x,a)\|_2\le1,
 \qquad
 \|\mu_h(\mathcal X_{h+1})\|_2=\sqrt D,
 \qquad
 \|\vartheta_h\|_2\le\sqrt D\,r_h^{\max}\le\sqrt D,
\end{equation*}
where $\mu_h=(\nu_{h,1},\ldots,\nu_{h,D})^{\top}$ for $h<H$ and
every coordinate of $\mu_H$ is $\delta_\partial$.
The supplied-feature specialization therefore has minimal linear-MDP
dimension exactly $D$.
\end{lemma}

\begin{lemma}[The exact quadratic component of a TD target]
\label{lem:qmdp-bellman-signal}
For every bounded measurable $V:\mathbb S_d\to\mathbb R$ and $h<H$,
\begin{equation}\label{eq:qmdp-affine-quadratic-target}
 r_h(x,a)+[P_hV](x,a)=c_h(V)+x^\top C_{h,a}(V)x,
\end{equation}
where $c_h(V)=D^{-1}\mathbf1^\top\vartheta_h+\mathbb E_\mu V$ and
$C_{h,a}(V)$ is given by \eqref{eq:qmdp-signal-matrix} in the proof.
All such matrices lie in the same unknown $(D-1)$-dimensional
trace-free symmetric subspace. If $\|V\|_\infty\le B_V$, then
\begin{equation}\label{eq:qmdp-signal-bounds}
 |c_h(V)|\le r_h^{\max}+B_V,\qquad
 \|C_{h,a}(V)\|_{\mathrm{op}}\le\|C_{h,a}(V)\|_F
 \le\frac{\tau_h}{Dd}(\sqrt D\,r_h^{\max}+\omega_hB_V).
\end{equation}
When $\vartheta_h=0$, the nonconstant target component comes
entirely from the continuation.
\end{lemma}

\begin{lemma}[Two-layer ReLU approximation]
\label{lem:qmdp-two-layer-approximation}
Let $\mathcal S$ be an $r$-dimensional subspace of trace-free symmetric
$d\times d$ matrices, $M>0$, and $\varepsilon,\delta\in(0,1)$.
Draw $v_1,\\ldots,v_{m_0}$ independently and uniformly on the unit
sphere, and assume \eqref{eq:qmdp-approximation-width}.
With probability at least $1-\delta$, the network
$\mathcal Q_{m_0,C,c}$ in \eqref{eq:qmdp-relu-approximation} satisfies,
simultaneously for $C\in\mathcal S$, $\|C\|_F\le M$, and $c\in\mathbb R$,
\begin{equation}\label{eq:qmdp-approximation-uniform}
 \sup_{x\in\mathbb S_d}
 |\mathcal Q_{m_0,C,c}(x)-(c+x^\top Cx)|\le\varepsilon.
\end{equation}
It is a two-layer ReLU network without hidden biases and also admits
a width-$2m_0$ realization with the paired initialization directions
and a trainable action head. The construction and width condition
are specified in the proof, with $\alpha_d$ as in
\eqref{eq:dqn-feature-moment-constants}.
\end{lemma}

\begin{lemma}[Uniform initialization data and later state distributions]
\label{lem:qmdp-state-marginals}
Initial observations are uniform on $\mathbb S_d$ conditional on all
preceding episodes.  If, during an episode, each action distribution
is fixed before the episode and does not depend on observations, then
every stage observation also has marginal law $\mu$ conditional on
that pre-episode history.
At a nonterminal stage $h<H$, under a general policy, the
value marginal has the form
\begin{equation}\label{eq:qmdp-adaptive-marginal}
 \frac{d\mathcal L(X_{h+1})}{d\mu}(x')
 =1+\frac{\tau_h\omega_h}{D}
      \mathbb E[T_{h,A_h}g_h(X_h)]^{\top}g_{h+1}(x').
\end{equation}
It therefore lies between $1-\frac{\tau_h\omega_h}{D}$ and $1+\frac{\tau_h\omega_h}{D}$.
\end{lemma}

\subsubsection{The Neural TD Update and Its Replay Gradient}
\paragraph{Coordinate updates of the deployed critic.}
Simultaneous SGD on the sampled TD loss gives
\addtocounter{equation}{3}
\begin{align*}
 w_{h,j}^+&=w_{h,j}+\frac{\eta_k}{b\sqrt m}
    \sum_{i:h_i=h}[Y_i-Q_{\theta_{h_i}}(x_i,a_i)]U_{h,a_i j}\mathbf1\{w_{h,j}^{\top}x_i>0\}x_i,
    \\
 U_{h,aj}^+&=U_{h,aj}+\frac{\eta_k}{b\sqrt m}
    \sum_{i:h_i=h,a_i=a}[Y_i-Q_{\theta_{h_i}}(x_i,a_i)]\relu(w_{h,j}^{\top}x_i),
    \\
 b_{h,a}^+&=b_{h,a}+\frac{\eta_k}{b}
    \sum_{i:h_i=h,a_i=a}[Y_i-Q_{\theta_{h_i}}(x_i,a_i)].
\end{align*}
Every right side uses the parameters before this same step. If a
minibatch contains $b_h>0$ stage-$h$ records, its update equals a
stage-normalized TD update with step size $\eta_k b_h/b$.
When $b_h=0$, that stage has no update.

\begin{definition}[Quantitative conditions for the initial TD update]
\label{def:dqn-feature-parameters} 
Fix $\delta\in(0,1)$,
define $\ell_{\mathrm{init}}:=\log \frac{8HKm5^d}{\delta}$, $\mathcal{E}_{\mathrm{init}}(n,b_h):=\sqrt{32B_Y^2d\ell_{\mathrm{init}}}\left(\frac{1}{\sqrt{n}}+\frac{1}{\sqrt{b_h}}\right)$.
The spherical moment constants are
\begin{equation}\label{eq:dqn-feature-moment-constants}
 \alpha_d=\frac{d^{3/2}\Gamma(d/2)}
 {2(d+1)\sqrt\pi\Gamma((d+1)/2)},\qquad
 \beta_d=\frac{d+1}{d}\alpha_d.
\end{equation}
Assume the width condition
\begin{equation}\label{eq:dqn-feature-width}
 m\ge128Ad\left\{Ar\log9+\log\frac{8H}{\delta}\right\}.
\end{equation}
For the selected stage-$h$ buffer window $I_h$,
$n=|I_h|$, and fitted value continuation $\widehat v$,
suppose
$\sum_a\bar\pi_a^2\|C_a(\widehat v)\|_F^2\ge S_0>0$ and
\begin{equation}\label{eq:dqn-feature-accuracy-condition}
\begin{aligned}
 4Ac_*\sqrt{\frac{A(r+1)\log9+\log(8H/\delta)}m}
 \le\frac{\alpha_d\sqrt{S_0}}{4\beta_dd},\quad
 \mathcal{E}_{\mathrm{init}}(n,b_h)
 \le\frac{\alpha_dS_0}{32dM_*}.
\end{aligned}
\end{equation}
The first inequality holds automatically when $c_*=0$.
The effective stage step size satisfies $\eta_{\mathrm{eff}}>0$ and
\begin{equation}\label{eq:dqn-feature-stepsize}
\begin{aligned}
 \frac{2\eta_{\mathrm{eff}}}{\sqrt m}
 \left[
 2\alpha_dM_*+\beta_dc_*+\mathcal{E}_{\mathrm{init}}(n,b_h)\right]\le1,\quad
 \frac{160dM_*\eta_{\mathrm{eff}}}{\alpha_dS_0\sqrt m}
 \left[2\alpha_dM_*+\beta_dc_*+\mathcal{E}_{\mathrm{init}}(n,b_h)\right]^2\le1.
\end{aligned}
\end{equation}
For a global minibatch, $\eta_{\mathrm{eff}}=\eta b_h/b$.
The target is fixed before the fresh replay indices are drawn.
\end{definition}

\begin{lemma}[Spherical quadratic moments]
\label{lem:dqn-feature-moments}
Let $X$ be uniform on the sphere of radius $\sqrt d$, let $w$ be a
unit vector, and let $C$ be symmetric and traceless. Then
\begin{align*}
 \mathbb E[X\mathbf 1\{w^\top X>0\}]&=\beta_dw,\\
 \mathbb E[(X^\top CX)X\mathbf 1\{w^\top X>0\}]
 &=\alpha_d\{2Cw-(w^\top Cw)w\},\\
 \mathbb E[(X^\top CX)\operatorname{ReLU}(w^\top X)]
 &=\alpha_dw^\top Cw.
\end{align*}
For nonzero $w$, the last expression becomes
$\alpha_dw^\top Cw/\|w\|$. For a uniform unit vector $w$,
\[
 \mathbb E\|Cw\|^2=\frac{\|C\|_F^2}{d},\qquad
 \mathbb E(w^\top Cw)^2=\frac{2\|C\|_F^2}{d(d+2)}.
\]
\end{lemma}

\begin{lemma}[A smooth covariance functional]
\label{lem:dqn-feature-hessian}
For symmetric $M$ and $w\ne0$, put $f_M(w)=w^\top Mw/\|w\|$.
Then
\[
 \nabla f_M(w)=\frac{2Mw}{\|w\|}
                -\frac{(w^\top Mw)w}{\|w\|^3},\qquad
 \|\nabla^2f_M(w)\|_{\mathrm{op}}\le\frac{3\|M\|_{\mathrm{op}}}{\|w\|}.
\]
For unit $w$ and $\Gamma=2Mw-(w^\top Mw)w$,
\[
 \|\Gamma\|^2=4\|Mw\|^2-3(w^\top Mw)^2,
 \quad \|Mw\|^2\le\|\Gamma\|^2\le4\|M\|_{\mathrm{op}}^2,
 \quad \Gamma^\top w=w^\top Mw.
\]
\end{lemma}

\begin{lemma}[Uniform random-feature alignment and offset control]
\label{lem:dqn-feature-init}
Under \eqref{eq:dqn-feature-width}, with failure probability at most
$\delta/4$, simultaneously across the $H$ stage initializations and
all collections $B_a\in\mathcal S_h$,
\begin{equation}\label{eq:dqn-feature-gram}
 \frac1m\sum_j\left\|\sum_aU_{a,j}^{(0)}B_aw_j\right\|^2
 \ge\frac1{2d}\sum_a\|B_a\|_F^2.
\end{equation}
With additional failure probability at most $\delta/4$, simultaneously
for these stages, every $z\in\mathbb R^A$, and every such $B$,
\begin{equation}\label{eq:dqn-feature-offset}
\begin{aligned}
 &\left|\frac1m\sum_j
       \left(\sum_aU_{a,j}^{(0)}z_a\right)
       w_j^\top\left(\sum_aU_{a,j}^{(0)}B_a\right)w_j\right|\\
 &\le4A\sqrt{\frac{A(r+1)\log9+\log(8H/\delta)}{m}}\|z\|_2\left(\sum_a\|B_a\|_F^2\right)^{1/2}.
\end{aligned}
\end{equation}
The collections and vectors can therefore be fitted from the same
data and initialization after the events have been established.
\end{lemma}

\begin{lemma}[Uniform initial-gradient concentration on replay]
\label{lem:dqn-feature-replay}
Use Algorithm~\ref{alg:dqn-online}. For each $(h,j)$, fix a
$\mathscr H_{j-1}$-measurable continuation-selection rule, and consider
pairs at which stage $h$ remains at initialization and the selected
continuation satisfies the conditional target model and boundedness
in Condition~\ref{cond:dqn-feature}. Let $\widehat G_{I_h}$, $G_{I_h}$,
and $\widehat G^{\mathrm{mb}}$ be the collected, conditional-population,
and stage-normalized replay negative hidden gradients defined in the proof.
With failure probability at most $\delta/4$, simultaneously over
these pairs and their prescribed continuations,
\begin{equation}\label{eq:dqn-feature-replay-error}
 \max\!\left\{\|\widehat G_{I_h}(\widehat v)-G_{I_h}(\widehat v)\|_F,
 \sqrt m\max_s\|\widehat G_{I_h,s}(\widehat v)-G_{I_h,s}(\widehat v)\|_2\right\}
 \le\sqrt{\frac{32B_Y^2d\ell_{\mathrm{init}}}{n}}.
\end{equation}
With additional failure probability at most $\delta/4$, simultaneously
over the same pairs with positive stage replay count $b_h$,
\begin{equation}\label{eq:dqn-feature-minibatch-radius}
 \max\!\left\{\|\widehat G^{\mathrm{mb}}-\widehat G_{I_h}(\widehat v)\|_F,
 \sqrt m\max_s\|\widehat G_s^{\mathrm{mb}}-\widehat G_{I_h,s}(\widehat v)\|_2\right\}
 \le\sqrt{\frac{32B_Y^2d\ell_{\mathrm{init}}}{b_h}}.
\end{equation}
For the actual copied continuation, the hidden update uses effective
step size $\eta_{\mathrm{eff}}=\eta b_h/b$. Uniformity concerns the
prescribed pre-collection rules, not post-collection selection.
\end{lemma}

\subsubsection{Uniform Replay Loss and Finite Neural Classes}

\begin{lemma}[Uniform excess TD loss on adaptive replay windows]
\label{lem:dqn-uniform-replay}
Let $\mathcal Q_h\subseteq\{f:\mathcal X_h\times\mathcal A\to[-B,B]\}$
and $\mathcal V_{h+1}\subseteq\{v:\mathcal X_{h+1}\to[-B,B]\}$ be
deterministic classes, with $B\ge1$ and $\mathcal V_{H+1}=\{0\}$.
Each episode's behavior rule is selected from the completed history;
it may update its critic and depend on within-episode observations.
Let $\mu_{i,h}$ denote its stage-$h$ state--action law conditional on
$\mathscr F_{i-1}$, integrating over the trajectory and replay draws.
The initial sigma-field may include network initialization.
Fix $u>0$ and finite supremum-norm $u$-nets of cardinalities at most
$N_Q(u),N_V(u)$ uniformly over stages.
For every nonempty consecutive window $I$, let $n=|I|$ and
$\bar\mu_{I,h}=n^{-1}\sum_{i\in I}\mu_{i,h}$. Write
\[
 Y_i(v)=R_{i,h}+v(X_{i,h+1}),\qquad
 \widehat L_I(q;v)=\frac1n\sum_{i\in I}
                 [q(X_{i,h},A_{i,h})-Y_i(v)]^2.
\]
With probability at least $1-\delta$, simultaneously for every stage,
window, and $f\in\mathcal Q_h,v\in\mathcal V_{h+1}$,
\stepcounter{equation}
\begin{equation*}
\begin{aligned}
 &\frac34\|f-\mathcal T_hv\|_{L^2(\bar\mu_{I,h})}^2
       -\frac{36(2B+1)^2}{n}\log\frac{2HT^2N_Q(u)N_V(u)}\delta-13(2B+1)u\\
 &\qquad\le\frac54\|f-\mathcal T_hv\|_{L^2(\bar\mu_{I,h})}^2
       +\frac{36(2B+1)^2}{n}\log\frac{2HT^2N_Q(u)N_V(u)}\delta+13(2B+1)u.
\end{aligned}
\end{equation*}
The same event permits data-dependent windows, critics, targets,
and comparison readouts belonging to these collections.
\end{lemma}

\begin{lemma}[Replay risk of the actual learned representation]
\label{lem:dqn-risk-certificate}
On the event of Lemma~\ref{lem:dqn-uniform-replay}, fix any fitted
representation $W$, fitted target $v\in\mathcal V_{h+1}$, and readout class
$\mathcal Q_h(W)\subseteq\mathcal Q_h$ containing the actual critic $f$.
Then the actual critic's Bellman risk satisfies
\stepcounter{equation}
\begin{equation*}
\begin{aligned}
 \|f-\mathcal T_hv\|_{L^2(\bar\mu_{I,h})}^2
 &\le\frac43\Bigl[\widehat L_I(f;v)
            -\inf_{q\in\mathcal Q_h(W)}\widehat L_I(q;v)\Bigr]\\
 &\quad+\frac53\inf_{q\in\mathcal Q_h(W)}
                    \|q-\mathcal T_hv\|_{L^2(\bar\mu_{I,h})}^2\\
 &\quad+\frac{96(2B+1)^2}{n}\log\frac{2HT^2N_Q(u)N_V(u)}\delta
                              +\frac{104}{3}(2B+1)u.
\end{aligned}
\end{equation*}
\end{lemma}

\begin{lemma}[A parameter cover for finite ReLU critics]
\label{lem:dqn-network-cover}
Suppose $\mathcal X_h\subseteq[-B_x,B_x]^d$ and a ReLU critic has
$P_\theta$ parameters in a deterministic box $[-R_\theta,R_\theta]^{P_\theta}$.
If its realization map obeys
\[
 \sup_{h,x,a}|Q_\theta(h,x,a)-Q_{\theta'}(h,x,a)|
 \le L_\theta\|\theta-\theta'\|_\infty,
\]
then
\[
 \log N_Q(u)\le P_\theta\log\left(2+\frac{2R_\theta L_\theta}{u}\right),
 \qquad N_V(u)\le N_Q(u),
\]
where the value class consists of action maxima of this critic class.
For the two-layer architecture
$Q_\theta(x,a)=b_a+m^{-1/2}\sum_{j=1}^m U_{a,j}\relu(w_j^\top x)$,
with every scalar parameter bounded by $R_\theta$,
one may take
\[
 B_Q=R_\theta+\sqrt m\,dB_xR_\theta^2,
 \qquad L_\theta=1+2\sqrt m\,dB_xR_\theta.
\]
Here $P_\theta$ counts the parameters of the finite critic
architecture; $D$ is the dimension of the unknown MDP representation.
\end{lemma}

\begin{lemma}[A deterministic envelope for unprojected TD updates]
\label{lem:dqn-parameter-envelope}
Use the two-layer architecture in Lemma~\ref{lem:dqn-network-cover}
and initialize every scalar parameter in $[-R_0,R_0]$, with $R_0\ge1$.
There are at most $K_{\mathrm{upd}}$ plain SGD updates, with predetermined
step sizes $\eta_k\ge0$. Define recursively
\begin{align*}
 B(R)&=R+\sqrt m\,dB_xR^2,\qquad G(R)=(1+2B(R))\max\{1,dB_xR/\sqrt m\},\\
 R_{k+1}&=R_k+\eta_kG(R_k),\qquad 0\le k<K_{\mathrm{upd}}.
\end{align*}
Every online parameter after $k$ updates and every parameter in the
copied target network lies in $[-R_k,R_k]$. In particular,
$R_\theta=R_{K_{\mathrm{upd}}}$ and $B_Q=B(R_\theta)$ are valid
deterministic envelopes.
\end{lemma}

\subsubsection{Finite neural prediction and bounded readouts}
\begin{definition}[Sample requirements for a strict predictive gain]
\label{def:dqn-step-accuracy}
For the fixed-value continuation update in
Condition~\ref{cond:dqn-predictive-step}, take $d\ge2$, $A\ge2$,
and $\delta\in(0,1)$. The width and effective step size satisfy
\begin{equation}\label{eq:dqn-step-width}
 m\ge64(d+1)^2,\qquad 0<\eta_{\mathrm{eff}}\le1.
\end{equation}
For the prediction-risk comparison, also require
\begin{equation}\label{eq:dqn-step-size-contraction}
 \frac{\eta_{\mathrm{eff}}}{A}
 \max\!\left\{1+2\beta_d^2,
             \frac{2\alpha_d^2(d+1)}{d^2}\right\}\le\frac12.
\end{equation}
When the denominator is positive, the sample
requirement for the full prediction risk is
\[
 \min\{n,b_h,m\}\ge
 \left[\frac{432A(d+1)^2}
 {\beta_d^2\|c\|_{L^2(\nu)}^2+
    \alpha_d^2(2d+1)\|s\|_{L^2(\nu)}^2/d^2}\right]^2
 \log\!\frac{64Am5^d(1+4m\sqrt d)^d}{\delta}.
\]
For the centered signal risk, the corresponding requirement is
\[
 \min\{n,b_h,m\}\ge
 \left[\frac{432Ad^2(d+1)^2}
 {\alpha_d^2(2d+1)\|s\|_{L^2(\nu)}^2}\right]^2
 \log\!\frac{64Am5^d(1+4m\sqrt d)^d}{\delta},
 \qquad\|s\|_{L^2(\nu)}>0.
\]
The response identities in Lemma~\ref{lem:dqn-quadratic-eigenvalues}
give the full population operator explicitly as
\stepcounter{equation}
\begin{equation*}
 ((\mathcal K_W+\mathcal K_U+\mathcal K_b)f)(x,a)
 =\frac{1+2\beta_d^2}{A}c_a
   +\frac{2\alpha_d^2(d+1)}{Ad^2}s_a(x).
\end{equation*}
\end{definition}

\begin{lemma}[Uniform prediction after an initial replay step]
\label{lem:dqn-uniform-step}
Use the conditional collection model of
Condition~\ref{cond:dqn-predictive-step}, with $m\ge64(d+1)^2$ and
$0<\eta_{\mathrm{eff}}\le1$. Draw $b_h>0$ fresh independent uniform
replay indices from the $n$ retained records and take one simultaneous
SGD step. Let $F^+$ be its initial-head readout,
$Q_{\mathrm{full}}^+$ the updated critic, and $Q_{\mathrm{head}}^+$
the matched head-and-bias update with hidden weights fixed at $W_0$.
For the population operators in Lemma~\ref{lem:dqn-quadratic-eigenvalues},
with conditional probability at least $1-\delta$ given $\mathscr G$,
\[
\begin{aligned}
 \max\Bigl\{&\|F^+-\eta_{\mathrm{eff}}\mathcal K_Wf\|_\infty,\\
 &\|Q_{\mathrm{full}}^+-\eta_{\mathrm{eff}}
                 (\mathcal K_W+\mathcal K_U+\mathcal K_b)f\|_\infty,\\
 &\|Q_{\mathrm{head}}^+-\eta_{\mathrm{eff}}
                 (\mathcal K_U+\mathcal K_b)f\|_\infty\Bigr\}\\
 &\le\eta_{\mathrm{eff}}(d+1)^2
 \left(\frac1{\sqrt n}+\frac1{\sqrt{b_h}}+\frac1{\sqrt m}\right)
 \sqrt{\log\!\frac{64Am5^d(1+4m\sqrt d)^d}{\delta}}.
\end{aligned}
\]
The norms are over $(x,a)\in\mathbb S_d\times\mathcal A$.
The same conclusion holds conditional on a positive realized stage
count in a global minibatch, with its effective step size.
\end{lemma}
\begin{lemma}[Pointwise eigenvalues of the initial TD operators]
\label{lem:dqn-quadratic-eigenvalues}
For fixed $f_a(x)=c_a+x^\top C_ax$, with each $C_a$ symmetric and
trace-free, let $\mathcal K_W$, $\mathcal K_U$, and $\mathcal K_b$
be the population hidden-weight, head, and bias operators defined
in the proof. For every $a\in\mathcal A$ and $\|x\|_2=\sqrt d$,
\begin{align}
 A(\mathcal K_Wf)(x,a)
 &=\beta_d^2c_a+{\frac{\alpha_d^2(2d+1)}{d^2}}x^\top C_ax,
 \label{eq:dqn-eigen-hidden}\\ A(\mathcal K_Uf)(x,a)
 &=\beta_d^2c_a+{\frac{\alpha_d^2}{d^2}}x^\top C_ax,\notag\stepcounter{equation}
\\ A(\mathcal K_bf)(x,a)&=c_a.\notag\stepcounter{equation}
\end{align}
\end{lemma}
\begin{lemma}[Tracking with a target copy after every SGD update]
\label{lem:dqn-horizon-iterates}
Use the horizon-$H$ signed-quadratic MDP, $H\ge2$, and independent
paired initializations. Collect $K\ge H$ fresh batches of $n$ uniform
exploration episodes; after each batch, take one global SGD
step with learning rate $1/2$ and minibatch size $b$ from its $Hn$
records, then copy the target. Let $Q_{h,j}$ be the actual critics,
$Q_{h,0}=0$, and let $\ell,E_{\mathrm{run}}$ be defined in
\eqref{eq:dqn-long-log}--\eqref{eq:dqn-long-error} in the proof.
Assume even $m$ and
\stepcounter{equation}
\begin{equation*}
 m\ge64(d+1)^2(1+K/H)^2,\qquad
 b\ge12H\log\!\frac{32HK}{\delta},\qquad
 (K/H+1)E_{\mathrm{run}}\le\frac14.
\end{equation*}
With probability at least $1-\delta$, simultaneously over all stages
and update prefixes,
\stepcounter{equation}
\begin{equation*}
 \|Q_{h,j}\|_\infty\le\frac54<2.
\end{equation*}
Every hidden row and head entry moves by at most $6(K/H)\sqrt{d/m}$,
and each output bias has magnitude at most $3K/H$.
There are action-independent $|\gamma_h|<2$ such that
$\widetilde Q_h=Q_{h,K}+\gamma_h$ satisfies
\begin{equation}\label{eq:dqn-long-centered}
 \max_h\|\widetilde Q_h-Q_h^\star\|_\infty
 \le\beta_H^\star\exp\!\left(-\frac{7\lambda_2K}{64H}\right)
       +\frac87E_{\mathrm{run}},
\end{equation}
where $\lambda_2=2\alpha_d^2(d+1)/d^2$.
The shifts preserve the deployed greedy actions and ties.
\end{lemma}

\subsubsection{Action margins and the executable schedule}
\begin{lemma}[Axial value continuation projections throughout the horizon]
\label{lem:dqn-axial-stage-projection}
Use \eqref{eq:dqn-horizon-data}--\eqref{eq:dqn-axial-downstream-model}.
For $X\sim\mu$, the moments $\zeta_d=\mathbb E_\mu|g_{h,1}(X)|$ and
$\kappa_d=\mathbb E_\mu[g_{h,1}(X)|g_{h,1}(X)|]$ are independent of
$h\ge2$ and satisfy
\begin{equation}\label{eq:dqn-axial-projection-and-bounds}
 \mathbb E_\mu[g_h|g_{h,1}|]=\kappa_de_1\quad(h\ge2),\qquad
 \frac1{8d^2}\le\kappa_d\le\frac2{d(d+2)},\qquad\zeta_d>0.
\end{equation}
For $h\ge2$ and $z\ge0$,
\begin{equation}\label{eq:dqn-axial-stage-margin}
 \mathbb P_\mu\{|g_{h,1}|\le z\}\le\min\{1,2dz\},\qquad
 \mathbb E_\mu[|g_{h,1}|\mathbf1\{|g_{h,1}|\le z\}]\le2dz^2.
\end{equation}
With the coefficients $\beta_h^\star,c_h^\star$ in
\eqref{eq:dqn-horizon-E}, the optimal values are
\begin{equation}\label{eq:dqn-horizon-optimal-values}
 Q_h^\star(x,a)=c_h^\star+\beta_h^\star a g_{h,1}(x),\qquad
 V_h^\star(x)=c_h^\star+\beta_h^\star|g_{h,1}(x)|\quad(h\in[H]).
\end{equation}
The optimal action is the sign of $g_{h,1}$, with the prescribed
tie convention, and each action is optimal on a set of positive
probability. Every nonterminal transition has rank $D$, and the
minimal common linear factorization dimension is $D$.
Uniform actions at stage $k$ followed by optimal actions give the
constant value $c_k^\star$, so all earlier action contrasts vanish.
\end{lemma}

\begin{condition}[A balanced quadratic action contrast]
\label{cond:dqn-balanced-margin}
The ambient dimension $d\ge4$ is even. The unknown first-stage
signal matrix has the form
\[
 E_{1,1}=\frac1{\sqrt d}\,
 O\,\operatorname{diag}(I_{d/2},-I_{d/2})\,O^\top
\]
for an unknown orthogonal matrix $O$. The remaining matrices
$E_{1,2},\ldots,E_{1,D-1}$ complete it to an orthonormal family
of trace-free symmetric matrices, where
$2\le D\le d(d+1)/2$. The algorithm is given observations and
rewards, and does not receive $O$ or these matrices.
\end{condition}

\begin{lemma}[Distribution of the quadratic action contrast]
\label{lem:dqn-balanced-margin}
Under Condition~\ref{cond:dqn-balanced-margin}, let
$X\sim\mu=\operatorname{Unif}(\sqrt d\,\mathbb S^{d-1})$.
Then
\[
 g_{1,1}(X)\stackrel{\mathrm d}=\frac{2U-1}{\sqrt d},
 \qquad U\sim\operatorname{Beta}(d/4,d/4).
\]
The density of $g_{1,1}(X)$ is supported on
$[-1/\sqrt d,1/\sqrt d]$ and satisfies
\stepcounter{equation}
\begin{equation*}
 \sup_z f_{g_{1,1}(X)}(z)
 =\frac{\sqrt d\,2^{\,1-d/2}}{\mathrm B(d/4,d/4)}
 \le\frac{2d}{3},
\end{equation*}
where $\mathrm B$ is the beta function. For every $z\ge0$,
\stepcounter{equation}
\begin{equation*}
 \mathbb E[|g_{1,1}(X)|\mathbf1\{|g_{1,1}(X)|\le z\}]
 \le\frac{\sqrt d\,2^{\,1-d/2}}{\mathrm B(d/4,d/4)}z^2
 \le\frac{2d}{3}z^2.
\end{equation*}
\end{lemma}

\begin{lemma}[Policy loss from all learned action contrasts]
\label{lem:dqn-quadratic-policy-loss}
In the horizon-$H$ subclass, let $\widehat\pi$ be greedy for critics
satisfying
\stepcounter{equation}
\begin{equation*}
 \|\widehat Q_h-Q_h^\star\|_\infty\le e_h,\qquad e_h\ge0,
 \quad h\in[H].
\end{equation*}
Then $J^\star-J(\widehat\pi)\le2\sum_he_h$, and if $e_h\le e$ for all $h$,
\[
 J^\star-J(\widehat\pi)\le\min\{1,8\sqrt d\,e\}.
\]
Under Condition~\ref{cond:dqn-balanced-margin},
\begin{equation}\label{eq:dqn-geometric-margin-refinement}
 J^\star-J(\widehat\pi)
 \le\sum_{h=1}^H\min\!\left\{2\beta_h^\star,
                  \frac{9de_h^2}{2\beta_h^\star}\right\}.
\end{equation}
The general margin bound \eqref{eq:dqn-greedy-margin-transfer} and
balanced first-stage refinement \eqref{eq:dqn-balanced-contrast-transfer}
are given in the proof. All bounds allow action-independent shifts
of the deployed critics. For $\varepsilon$-greedy deployment,
$\varepsilon\in[0,1]$, the bounds $2\sum_he_h$, $8\sqrt d\,e$, and
\eqref{eq:dqn-geometric-margin-refinement} hold with an added $\varepsilon$.
For Algorithm~\ref{alg:dqn-online} and $T\ge Kn$,
\begin{equation}\label{eq:dqn-horizon-policy-regret-transfer}
 \Reg_T\le Kn+2(T-Kn)\sum_{h=1}^He_h.
\end{equation}
The geometric and balanced bounds may replace the deployment term;
every run also satisfies $\Reg_T\le T$.
\end{lemma}

\begin{lemma}[An executable horizon-$H$ replay schedule]
\label{lem:dqn-horizon-schedule}
Fix $n\ge1$, $K\ge H$, the rewards in \eqref{eq:dqn-horizon-data},
and independent paired zero-output stage initializations.
Algorithm~\ref{alg:dqn-online} has an equivalent interaction-indexed
implementation with replay capacity $Hn$, $U_{\mathrm{step}}=C_{\mathrm{copy}}=1$,
and zero-based gradient-call schedule
\begin{equation}\label{eq:dqn-horizon-executable-schedule}
 \eta_k=\frac12\mathbf1\!\left\{k\in\{Hjn-1:j\in[K]\}\right\}.
\end{equation}
Positive step $j$ uses precisely the transitions from episodes
$(j-1)n+1,\ldots,jn$, with targets \eqref{eq:dqn-horizon-actual-targets},
and is followed by a target copy. Stage $h$ remains at initialization
until $j_h=H-h+1$ and has $K-H+h$ potentially nonzero updates;
all reached stages continue training together. Online and target
networks agree at every action, with zero within-episode prediction drift.
After update $K$, the critic is frozen and its greedy policy is used.
For $p_{h,j}=b_{h,j}/b$, the stage fractions of a global minibatch,
\begin{equation}\label{eq:dqn-horizon-stage-counts}
 \mathbb P\!\left\{\exists j,h:\,
 p_{h,j}\notin[1/(2H),3/(2H)]\right\}
 \le2HK\exp\!\left(-\frac{b}{12H}\right).
\end{equation}
Conditional on the stage counts, replay indices are independent
uniform draws from the corresponding $n$ records.
Before each batch, the parameters and target are fixed; conditional
on the completed prefix, episodes are independent and every stage
has state law $\mu$ and an independent uniform action.
At $j_h$, its records and copied continuation are also independent
of the unused stage initialization. The explicit target and
multinomial count laws are given in the proof.
\end{lemma}

\subsubsection{The propagated signal and training budget}
For stage $h$, let $j_h=H-h+1$ be its first reached SGD
update. Write $\widehat v_h=\max_a Q_{h+1,j_h-1}(\cdot,a)$
for the copied value continuation, with $\widehat v_H=0$. The signal matrices at this
update are
\[
 C_{h,a}(\widehat v_h)=
 \begin{cases}
 \displaystyle\frac{a}{4Dd}\sum_{j=1}^{D-1}
       \mathbb E_\mu[g_{h+1,j}\widehat v_h]E_{h,j},&h<H,\\[3pt]
 \displaystyle\frac{a}{8Dd}E_{H,1},&h=H.
 \end{cases}
\]
Define the actual squared signal and its operator-norm bound by
\begin{equation}\label{eq:dqn-horizon-signal}
 S_h=\sum_{a\in\{-1,1\}}\frac14
                 \|C_{h,a}(\widehat v_h)\|_F^2,
 \qquad M_h=\max_a\|C_{h,a}(\widehat v_h)\|_{\mathrm{op}}.
\end{equation}
In particular, $S_H=(\beta_H^\star)^2/(2d^2)$.

\begin{definition}[An explicit training budget]
\label{def:dqn-reachable-budget}
Use the logarithm, update count $K$, and tracking bounds of
Lemma~\ref{lem:dqn-horizon-iterates}, with confidence $\delta/2$.
Assume its numerical budget and the additional inequalities
\begin{equation}\label{eq:dqn-reachable-budget}
\begin{gathered}
 m\ge64(d+1)^2,\qquad
 m\ge256d\{2(D-1)\log9+\log(32H/\delta)\},\\
 n\ge4d\ell,\qquad b\ge8Hd\ell.
\end{gathered}
\end{equation}
For the feature update, put
\stepcounter{equation}
\begin{equation*}
 \mathcal E=\frac{15}{8}\sqrt{d\ell}
                   \left(\frac1{\sqrt n}+\sqrt{\frac{2H}{b}}\right).
\end{equation*}
The inequalities specify a training budget polynomial in the horizon
and model parameters, apart from logarithms. They give
$\mathcal E\le2$ uniformly over the actual signal strengths
$S_h\ge0$.
\end{definition}

\begin{lemma}[Reaching informative TD updates]
\label{lem:dqn-reachable-budget}
Under Condition~\ref{cond:dqn-horizon}, $K\ge H$, $Kn\le T$,
and Definition~\ref{def:dqn-reachable-budget}, with probability at
least $1-\delta$, Lemma~\ref{lem:dqn-horizon-iterates} holds and,
simultaneously for every $h\in[H]$,
\begin{equation}\label{eq:dqn-reachable-gain}
 \mathcal G_h\ge\frac{\alpha_d^2S_h}{16Hd}
 -\frac{2d^2\ell}{H}\left(\frac1n+\frac Hb\right).
\end{equation}
Here $\mathcal G_h$ is the initial-head covariance gain in
\eqref{eq:dqn-feature-main-gain} at update $j_h=H-h+1$, and
$S_h$ is defined in \eqref{eq:dqn-horizon-signal}.
For any stage satisfying
\begin{equation}\label{eq:dqn-feature-signal-resolution}
 S_h>0,\qquad \mathcal E\le\frac{\alpha_d\sqrt{S_h}}{d},
\end{equation}
we have
\begin{equation}\label{eq:dqn-resolved-feature-gain}
 \mathcal G_h\ge\frac{\alpha_d^2S_h}{32Hd}>0.
\end{equation}
The sampling conditions \eqref{eq:dqn-terminal-feature-budget}
and \eqref{eq:dqn-endstage-feature-budget} imply, respectively,
\begin{align}
 \mathcal G_H&\ge
 \frac{\alpha_d^2(\beta_H^\star)^2}{64Hd^3}>0,
 \label{eq:dqn-terminal-feature-gain}\\
 \mathcal G_{H-1}&\ge
 \frac{\alpha_d^2\lambda_2^2}{2^{25}H^3D^4d^7}>0.
 \label{eq:dqn-first-nonterminal-feature-gain}
\end{align}
The latter occurs at update $2$, using the terminal critic $Q_{H,1}$.
\end{lemma}

\subsection{Proofs of the Main Results}\label{sec:lat-main-proofs}

\begin{proofof}{Theorem~\ref{thm:dqn-feature-gain}}
Fix a realization in the intersection of the two
initialization events in
Lemma~\ref{lem:dqn-feature-init}
and the collection and replay events in
Lemma~\ref{lem:dqn-feature-replay}.
Their joint failure probability is at most $\delta$.
Fix an eligible stage--update pair $(h,j)$ on this event.
Throughout the proof, use $s\in[m]$ for the hidden-neuron
index and suppress the stage index on the initial weights. Define
\[
 M_s=\sum_a\bar\pi_aU_{a,s}^{(0)}C_a(\widehat v),\quad
 \chi_s=\sum_a\bar\pi_aU_{a,s}^{(0)}c_a(\widehat v),\quad
 \Gamma_s=2M_sw_s-(w_s^\top M_sw_s)w_s.
\]
Lemma~\ref{lem:dqn-feature-moments} identifies the covariance
functional exactly:
\[
 \mathbb E_{\nu}[(X^\top C_a(\widehat v)X)F_{W}(X,a)]=\frac{\alpha_d}{\sqrt m}
                   \sum_s\frac{w_s^\top M_sw_s}{\|w_s\|}.
\]
At initialization,
\[
\begin{aligned}
 G_s^{\mathrm{quad}}&:=\left.\nabla_{w_s}\mathbb E_{\nu}[(X^\top C_a(\widehat v)X)F_W(X,a)]\right|_{W=W_0}
             =\frac{\alpha_d}{\sqrt m}\Gamma_s,\\
 G_s&=G_s^{\mathrm{quad}}+\frac{\beta_d}{\sqrt m}\chi_sw_s.
\end{aligned}
\]
The second term is the radial gradient contribution of the unknown
target mean.

\paragraph{Step 1: a lower bound on alignment.}
Taking the inner product of the two gradients and using
$\Gamma_s^\top w_s=w_s^\top M_sw_s$ gives
\begin{align*}
 \langle G^{\mathrm{quad}},G\rangle
 &=\frac{\alpha_d^2}{m}\sum_s\|\Gamma_s\|^2
   +\frac{\alpha_d\beta_d}{m}\sum_s\chi_sw_s^\top M_sw_s.
\end{align*}
Lemma~\ref{lem:dqn-feature-hessian} and
\eqref{eq:dqn-feature-gram}, applied to $B_a=\bar\pi_aC_a(\widehat v)$, imply
\begin{align*}
 \frac1m\sum_s\|\Gamma_s\|^2
 &\ge\frac1m\sum_s\|M_sw_s\|^2
 \ge\frac1{2d}\sum_a\bar\pi_a^2\|C_a(\widehat v)\|_F^2
.
\end{align*}
For the second term apply \eqref{eq:dqn-feature-offset} with
$z_a=\bar\pi_ac_a(\widehat v)$ and the same $B_a$. Since
\[
 \|z\|_2^2=\sum_a\bar\pi_a^2c_a(\widehat v)^2
 \le c_*^2\sum_a\bar\pi_a^2\le c_*^2,
 \qquad\sum_a\|B_a\|_F^2=\left(\sum_a\bar\pi_a^2\|C_a(\widehat v)\|_F^2\right),
\]
we obtain, successively,
\begin{align*}
 \langle G^{\mathrm{quad}},G\rangle
 &\ge\frac{\alpha_d^2\left(\sum_a\bar\pi_a^2\|C_a(\widehat v)\|_F^2\right)}{2d}
\\
 &\quad-4A\alpha_d\beta_d\sqrt{\frac{A(r+1)\log9+\log(8H/\delta)}{m}}c_*\sqrt{\left(\sum_a\bar\pi_a^2\|C_a(\widehat v)\|_F^2\right)}\\
 &\ge\frac{\alpha_d^2\left(\sum_a\bar\pi_a^2\|C_a(\widehat v)\|_F^2\right)}{2d}
      -\frac{\alpha_d^2\sqrt{S_0\left(\sum_a\bar\pi_a^2\|C_a(\widehat v)\|_F^2\right)}}{4d}\\
 &\ge\frac{\alpha_d^2\left(\sum_a\bar\pi_a^2\|C_a(\widehat v)\|_F^2\right)}{4d}.
\end{align*}
The second inequality substitutes the offset condition; the third uses
$\sqrt{S_0\left(\sum_a\bar\pi_a^2\|C_a(\widehat v)\|_F^2\right)}\le \left(\sum_a\bar\pi_a^2\|C_a(\widehat v)\|_F^2\right)$.

\paragraph{Step 2: the actual replay update and its error.}
Fix an eligible stage--update pair $(h,j)$ on the common event
chosen at the beginning of the proof. Use $s$ for the neuron
index and suppress the stage index on its initial weights.
Write $\widehat G:=\widehat G^{\mathrm{mb}}$ defined in Lemma \ref{lem:dqn-feature-replay} and $E:=\widehat G-G$. We decompose $E$ via a gradient induced by the stage-$h$ buffer $I_h$. Define
\[
 \widehat L_{h,I_h}(\theta_h;v)
 :=\frac1{2n}\sum_{i\in I_h}
 (Q_{\theta_h}(X_i,A_i)-Y_i(v))^2.
\]
Since $Q_{\theta_h^{(0)}}=0$ and targets are detached,
\begin{align*}
 \widehat G_{I_h,s}(v):=-\left.\nabla_{w_{h,s}}
 \widehat L_{h,I_h}(\theta_h;v)
 \right|_{\theta_h=\theta_h^{(0)}}=\frac1{n\sqrt m}\sum_{i\in I_h}
 Y_i(v)U_{A_i,s}^{(0)}
 \mathbf1\{w_s^\top X_i>0\}X_i.
\end{align*}
For the per-record contribution
\[
 Z_{i,s}(\widehat v)
 :=\frac1{\sqrt m}Y_i(\widehat v)U_{A_i,s}^{(0)}
              \mathbf1\{w_s^\top X_i>0\}X_i,
\]
conditioning on $\mathscr H_{j-1}$ fixes both $\widehat v$
and the initial weights. We then define
\[
\begin{aligned}
G_{I_h,s}(\widehat v)
&:=
\mathbb E\left[
\widehat G_{I_h,s}(\widehat v)
\mid\mathscr H_{j-1}
\right]\\
&=
\frac1n\sum_{i\in I_h}
\mathbb E\left[
Z_{i,s}(\widehat v)
\mid\mathscr H_{j-1}
\right]\\
&=
\frac1{A\sqrt m}\sum_a U_{a,s}^{(0)}
\mathbb E_\mu\left[
\big(c_a(\widehat v)+X^\top C_a(\widehat v)X\big)
\mathbf1\{w_s^\top X>0\}X
\right]\\
&=
\frac1{\sqrt m}
\left(
\alpha_d\Gamma_s+\beta_d\chi_sw_s
\right)
=G_s,
\end{aligned}
\]
where the penultimate equality follows from
Lemma~\ref{lem:dqn-feature-moments}, and the last
equality uses the definition of $G_s$ in Step 1.
Consequently, $G_{I_h}(\widehat v)=G.$

We therefore decompose
\[
 E=
 \underbrace{\widehat G^{\mathrm{mb}}
             -\widehat G_{I_h}(\widehat v)}_{
                    \text{replay sampling error}}
 +\underbrace{\widehat G_{I_h}(\widehat v)
             -G_{I_h}(\widehat v)}_{
                    \text{data collection error}}.
\]
Lemma~\ref{lem:dqn-feature-replay} controls the second term
conditional on the pre-collection history and the first term
conditional on the pre-draw history and stage count.
Its probability statements have already been averaged over
these histories and made simultaneous over the $HK$
stage--update pairs. On the common event, the triangle
inequality in each norm gives
\[
 \max\left\{\|E\|_F,\sqrt m\max_s\|E_s\|_2\right\}
 \le\sqrt{\frac{32B_Y^2d\ell_{\mathrm{init}}}{n}}
   +\sqrt{\frac{32B_Y^2d\ell_{\mathrm{init}}}{b_h}}
 =\mathcal E_{\mathrm{init}}(n,b_h).
\]
Since $\sum_a\bar\pi_a=1$ and $|U_{a,s}^{(0)}|=1$,
\[
 \|M_s\|_{\mathrm{op}}\le M_*,\quad
 |\chi_s|\le c_*,\quad
 \|\Gamma_s\|_2\le2M_*,\quad
 \|G^{\mathrm{quad}}\|_F\le2\alpha_dM_*.
\]
Cauchy--Schwarz and \eqref{eq:dqn-feature-accuracy-condition}
therefore give
\[
 |\langle G^{\mathrm{quad}},E\rangle|
 \le2\alpha_dM_*\mathcal E_{\mathrm{init}}(n,b_h)
 \le\frac{\alpha_d^2S_0}{16d}
 \le\frac{\alpha_d^2}{16d}
       \sum_a\bar\pi_a^2\|C_a(\widehat v)\|_F^2.
\]
The actual hidden update is
$W^+=W_0+\eta_{\mathrm{eff}}(G+E)$, with
$\eta_{\mathrm{eff}}=\eta b_h/b$, and
\[
 \max\left\{\|G+E\|_F,
          \sqrt m\max_s\|G_s+E_s\|_2\right\}
 \le2\alpha_dM_*+\beta_dc_*
          +\mathcal E_{\mathrm{init}}(n,b_h).
\]
The first inequality in \eqref{eq:dqn-feature-stepsize} implies
$\max_s\|w_s^+-w_s\|_2\le1/2$, so every row along the update
segment has norm at least $1/2$. Its second inequality gives
\[
 \frac{10\alpha_dM_*\eta_{\mathrm{eff}}^2}{\sqrt m}
       \|G+E\|_F^2
 \le\frac{\eta_{\mathrm{eff}}\alpha_d^2S_0}{16d}.
\]

\paragraph{Step 3: convert alignment to a finite feature change.}
Lemma~\ref{lem:dqn-feature-hessian} bounds the Hessian of
the function $W\mapsto\mathbb E_\nu[(X^\top C_a(\widehat v)X)F_W(X,a)]$ on the update segment by
$20\alpha_dM_*/\sqrt m$. Taylor's theorem with its integral
remainder, followed by the bounds already obtained, gives
\begin{equation}\label{eq:dqn-feature-gain-explicit}
\begin{aligned}
 &\mathbb E_\nu[(X^\top C_a(\widehat v)X)\{F_{W^+}(X,a)-F_{W_0}(X,a)\}]\\
 &\ge\eta_{\mathrm{eff}}\langle G^{\mathrm{quad}},G+E\rangle
      -\frac{10\alpha_dM_*\eta_{\mathrm{eff}}^2}{\sqrt m}
          \|G+E\|_F^2\\
 &\ge\eta_{\mathrm{eff}}\left(
       \frac{\alpha_d^2\left(\sum_a\bar\pi_a^2\|C_a(\widehat v)\|_F^2\right)}{4d}
      -\frac{\alpha_d^2\left(\sum_a\bar\pi_a^2\|C_a(\widehat v)\|_F^2\right)}{16d}\right)
      -\frac{10\alpha_dM_*\eta_{\mathrm{eff}}^2\|G+E\|_F^2}{\sqrt m}\\
 &\ge\eta_{\mathrm{eff}}\left(
       \frac{\alpha_d^2\left(\sum_a\bar\pi_a^2\|C_a(\widehat v)\|_F^2\right)}{4d}
      -\frac{\alpha_d^2\left(\sum_a\bar\pi_a^2\|C_a(\widehat v)\|_F^2\right)}{16d}
      -\frac{\alpha_d^2S_0}{16d}\right)\\
 &\ge\frac{\eta_{\mathrm{eff}}\alpha_d^2\left(\sum_a\bar\pi_a^2\|C_a(\widehat v)\|_F^2\right)}{8d}.
\end{aligned}
\end{equation}
The third line uses the remainder bound established in Step 2, and the final
line again uses $S_0\le \left(\sum_a\bar\pi_a^2\|C_a(\widehat v)\|_F^2\right)$. The initialization events are uniform over the signal matrices
and offsets, and the collection and replay events are simultaneous
over the actual stage--update pairs and their pre-collection
continuations. Thus the pathwise argument applies to every
eligible update, with its realized continuation held fixed
throughout the covariance calculation.
\end{proofof}

\begin{proofof}{Theorem~\ref{thm:dqn-bounded-bridge}}
Condition on the prefix in Lemma~\ref{lem:dqn-uniform-step} and fix
its uniform event. Expectations below use the fresh distribution $\nu$.
The initial-head readout of the updated features is
$F_W^+(x,a)=m^{-1/2}\sum_jU_{a,j}^{(0)}
\operatorname{ReLU}((w_j^+)^\top x)$. Define the population response
functions
\begin{align*}
 \widetilde F_W&=\frac{\eta_{\mathrm{eff}}}{A}
      \left(\beta_d^2c+\frac{\alpha_d^2(2d+1)}{d^2}s\right),\\
 \widetilde Q_{\mathrm{full}}&=\frac{\eta_{\mathrm{eff}}}{A}
      \left((1+2\beta_d^2)c+\frac{2\alpha_d^2(d+1)}{d^2}s\right),\\
 \widetilde Q_{\mathrm{head}}&=\frac{\eta_{\mathrm{eff}}}{A}
      \left((1+\beta_d^2)c+\frac{\alpha_d^2}{d^2}s\right).
\end{align*}
The uniform-step lemma gives
\stepcounter{equation}
\begin{equation*}
\begin{aligned}
 &\max\{\|F_W^+-\widetilde F_W\|_\infty,
       \|Q_{\mathrm{full}}^+-\widetilde Q_{\mathrm{full}}\|_\infty,
       \|Q_{\mathrm{head}}^+-\widetilde Q_{\mathrm{head}}\|_\infty\}\\
 &\qquad\le12\eta_{\mathrm{eff}}(d+1)^2 \left(\frac1{\sqrt n}+\frac1{\sqrt{b_h}}+\frac1{\sqrt m}\right)\sqrt{\log\!\frac{64Am5^d(1+4m\sqrt d)^d}{\delta}}.
\end{aligned}
\end{equation*}
Lemma~\ref{lem:dqn-quadratic-eigenvalues} computes these response
functions from the population gradient.

\paragraph{Step 1: construct an admissible bounded readout.}
The target bound gives $|f_a(x)|\le1$. Trace-freeness and
$\mathbb E_\mu[XX^\top]=I_d$ imply
\[
 \mathbb E_\mu s_a(X)
 =\operatorname{tr}(C_a\mathbb E_\mu[XX^\top])
 =\operatorname{tr}(C_a)=0.
\]
Hence $c_a=\mathbb E_\mu f_a(X)$ and $|c_a|\le1$.
The moment identity $\beta_d=(d+1)\alpha_d/d$ gives
\[
 1-\frac{d^2\beta_d^2}{\alpha_d^2(2d+1)}
 =1-\frac{(d+1)^2}{2d+1}=-\frac{d^2}{2d+1}.
\]
Choose
\begin{equation}\label{eq:dqn-bounded-witness}
 U_{a,j}^\dagger=\frac{Ad^2U_{a,j}^{(0)}}
                         {\eta_{\mathrm{eff}}\alpha_d^2(2d+1)},
 \qquad b_a^\dagger=
       \left(-\frac{d^2}{2d+1}\right)c_a.
\end{equation}
Their magnitudes satisfy
\[
 |U_{a,j}^\dagger|=\frac{Ad^2}{\eta_{\mathrm{eff}}\alpha_d^2(2d+1)},
 \qquad |b_a^\dagger|
 \le\frac{d^2}{2d+1},
\]
so they belong to the box in
\eqref{eq:dqn-bounded-readout-box}. With the actual updated hidden weights,
\begin{align*}
 q^\dagger(x,a)
 &=\left(-\frac{d^2}{2d+1}\right)c_a
       +\frac{Ad^2F_W^+(x,a)}{\eta_{\mathrm{eff}}\alpha_d^2(2d+1)},\\
 q^\dagger(x,a)-f_a(x)
 &=\left(-\frac{d^2}{2d+1}\right)c_a
       +\frac{Ad^2F_W^+(x,a)}{\eta_{\mathrm{eff}}\alpha_d^2(2d+1)}
       -c_a-s_a(x)\\
 &=\frac{Ad^2}{\eta_{\mathrm{eff}}\alpha_d^2(2d+1)}
                        [F_W^+(x,a)-\widetilde F_W(x,a)].
\end{align*}
Taking the supremum and substituting the uniform bound gives
\begin{equation}\label{eq:dqn-bounded-witness-explicit-error}
\begin{aligned}
 \|q^\dagger-f\|_\infty
 &\le\frac{Ad^2}{\eta_{\mathrm{eff}}\alpha_d^2(2d+1)}
                             \|F_W^+-\widetilde F_W\|_\infty\\
 &\le\frac{Ad^2\,12\eta_{\mathrm{eff}}(d+1)^2}
            {\eta_{\mathrm{eff}}\alpha_d^2(2d+1)}\left(\frac1{\sqrt n}+\frac1{\sqrt{b_h}}+\frac1{\sqrt m}\right)\sqrt{\log\!\frac{64Am5^d(1+4m\sqrt d)^d}{\delta}}\\
 &=\frac{12Ad^2(d+1)^2}{\alpha_d^2(2d+1)}\left(\frac1{\sqrt n}+\frac1{\sqrt{b_h}}+\frac1{\sqrt m}\right)\sqrt{\log\!\frac{64Am5^d(1+4m\sqrt d)^d}{\delta}}.
\end{aligned}
\end{equation}
The final equality cancels the positive effective learning rate.
To obtain the order in \eqref{eq:dqn-bounded-bridge-approx}, use
the logarithm and sample-sum inequalities in
\eqref{eq:dqn-step-order-conversion}, together with
\[
 \frac{d^2(d+1)^2}{2d+1}
 \le\frac{d^2(3d/2)^2}{2d}
 =\frac98d^3.
\]
The inequality uses $d+1\le3d/2$ for $d\ge2$ and $2d+1\ge2d$.

\paragraph{Step 2: include the actual SGD heads.}
At initialization $Q_0=0$, so simultaneous SGD gives
\begin{align*}
 U_{a,j}^+-U_{a,j}^{(0)}
 &=\frac{\eta_{\mathrm{eff}}}{b_h\sqrt m}
           \sum_{i:A_i=a}Y_i\operatorname{ReLU}(w_j^\top X_i),\\
 b_a^+&=\frac{\eta_{\mathrm{eff}}}{b_h}\sum_{i:A_i=a}Y_i.
\end{align*}
Cauchy--Schwarz gives
$\operatorname{ReLU}(w_j^\top X_i)\le\|w_j\|_2\|X_i\|_2=\sqrt d$.
There are at most $b_h$ terms, including repeated replay indices, and
$|Y_i|\le1$. The triangle inequality therefore yields
\begin{align*}
 |U_{a,j}^+-U_{a,j}^{(0)}|
 &\le\frac{\eta_{\mathrm{eff}}}{b_h\sqrt m}
                           \sum_{i:A_i=a}|Y_i|\sqrt d
 \le\eta_{\mathrm{eff}}\sqrt{d/m},\\
 |U_{a,j}^+|&\le1+\eta_{\mathrm{eff}}\sqrt{d/m},\qquad |b_a^+|\le\frac{\eta_{\mathrm{eff}}}{b_h}
                         \sum_{i:A_i=a}|Y_i|\le\eta_{\mathrm{eff}}.
\end{align*}
Taking maxima of these bounds and the witness bounds gives the
readout box
\begin{equation}\label{eq:dqn-bounded-readout-box}
 \begin{aligned}
 |U_{a,j}|&\le\max\!\left\{
 \frac{Ad^2}{\eta_{\mathrm{eff}}\alpha_d^2(2d+1)},
                  1+\eta_{\mathrm{eff}}\sqrt{d/m}\right\},\\
 |b_a|&\le\max\!\left\{\frac{d^2}{2d+1},
                                      \eta_{\mathrm{eff}}\right\}.
 \end{aligned}
\end{equation}
For the orders stated in the theorem,
$d^2/(2d+1)\le d/2$, $\eta_{\mathrm{eff}}\le1$,
and $m\ge d$ imply
\[
 \begin{aligned}
 |U_{a,j}|&\le\max\!\left\{
       \frac{Ad}{2\eta_{\mathrm{eff}}\alpha_d^2},2\right\}
       \le\frac{Ad}{\eta_{\mathrm{eff}}\alpha_d^2},\qquad |b_a|\le\max\{d/2,1\}\le d.
\end{aligned}
\]
The last head inequality uses $A,d\ge2$ and
$\alpha_d\le\beta_d\le1/2$, which follows from
$\beta_d=\mathbb E|X_1|/2\le(\mathbb EX_1^2)^{1/2}/2$.
The frozen-hidden comparator uses these
same heads and biases, because all gradients are evaluated at the
same initial parameters and targets. The hidden update similarly gives
\[
 \|w_j^+-w_j\|_2
 \le\frac{\eta_{\mathrm{eff}}}{b_h\sqrt m}
                    \sum_i|Y_i|\|X_i\|_2
 \le\eta_{\mathrm{eff}}\sqrt{d/m},
 \quad \|w_j^+\|_2\le1+\eta_{\mathrm{eff}}\sqrt{d/m}.
\]
Every readout in the displayed box has the finite envelope
\begin{align*}
 |Q_{W^+,U,\mathbf b}(x,a)|
 &\le |b_a|+m^{-1/2}\sum_j|U_{a,j}|\|w_j^+\|_2\|x\|_2\\
 &\le\max\!\left\{\frac{d^2}{2d+1},
                      \eta_{\mathrm{eff}}\right\}\\
 &\quad+\sqrt{md}(1+\eta_{\mathrm{eff}}\sqrt{d/m})
      \max\!\left\{\frac{Ad^2}{\eta_{\mathrm{eff}}\alpha_d^2(2d+1)},
                   1+\eta_{\mathrm{eff}}\sqrt{d/m}\right\}.
\end{align*}

\paragraph{Step 3: compare the population prediction errors.}
For each action $\mathbb E_\mu s_a(X)=0$, so
$\langle c,s\rangle_{L^2(\nu)}=A^{-1}\sum_ac_a\mathbb E_\mu s_a(X)=0$.
Expanding the two squared norms and cancelling their cross terms gives
\begin{align*}
 \mathcal R_\nu(\widetilde Q_{\mathrm{full}})
 &=\left(1-\frac{\eta_{\mathrm{eff}}(1+2\beta_d^2)}A\right)^2\|c\|^2
   +\left(1-\frac{2\eta_{\mathrm{eff}}\alpha_d^2(d+1)}{Ad^2}\right)^2\|s\|^2,\\
 \mathcal R_\nu(\widetilde Q_{\mathrm{head}})
 &=\left(1-\frac{\eta_{\mathrm{eff}}(1+\beta_d^2)}A\right)^2\|c\|^2
   +\left(1-\frac{\eta_{\mathrm{eff}}\alpha_d^2}{Ad^2}\right)^2\|s\|^2.
\end{align*}
Here and for the remaining risk calculations, norms are in $L^2(\nu)$.
For real $u,v$, factoring a difference of squares gives
\begin{align*}
 \left(1-\frac{\eta_{\mathrm{eff}}}Av\right)^2
       -\left(1-\frac{\eta_{\mathrm{eff}}}Au\right)^2
 &=\frac{2\eta_{\mathrm{eff}}}A(u-v)
        -\frac{\eta_{\mathrm{eff}}^2}{A^2}(u^2-v^2)\\
 &=\frac{\eta_{\mathrm{eff}}}A(u-v)
                         \left[2-\frac{\eta_{\mathrm{eff}}}A(u+v)\right].
\end{align*}
Apply this identity to $u=1+2\beta_d^2$, $v=1+\beta_d^2$, and then
to $u=2\alpha_d^2(d+1)/d^2$, $v=\alpha_d^2/d^2$. Their differences
are $\beta_d^2$ and $\alpha_d^2(2d+1)/d^2$. Thus
\begin{align*}
 \mathcal R_\nu(\widetilde Q_{\mathrm{head}})-\mathcal R_\nu(\widetilde Q_{\mathrm{full}})
 &=\frac{\eta_{\mathrm{eff}}\beta_d^2}A
      \left[2-\frac{\eta_{\mathrm{eff}}(2+3\beta_d^2)}A\right]\|c\|^2\\
 &\quad+\frac{\eta_{\mathrm{eff}}\alpha_d^2(2d+1)}{Ad^2}
      \left[2-\frac{\eta_{\mathrm{eff}}\alpha_d^2(2d+3)}{Ad^2}\right]\|s\|^2.
\end{align*}
The step restriction \eqref{eq:dqn-step-size-contraction} gives
\[
 \frac{\eta_{\mathrm{eff}}(2+3\beta_d^2)}A
 \le\frac{2\eta_{\mathrm{eff}}(1+2\beta_d^2)}A\le1,
 \qquad
 \frac{\eta_{\mathrm{eff}}\alpha_d^2(2d+3)}{Ad^2}
 \le\frac{4\eta_{\mathrm{eff}}\alpha_d^2(d+1)}{Ad^2}\le1.
\]
Each bracket is therefore at least one. Multiplication by the
nonnegative squared norms yields
\[
 \mathcal R_\nu(\widetilde Q_{\mathrm{head}})-\mathcal R_\nu(\widetilde Q_{\mathrm{full}})
 \ge\frac{\eta_{\mathrm{eff}}}A
       \left[\beta_d^2\|c\|^2+\frac{\alpha_d^2(2d+1)}{d^2}\|s\|^2\right].
\]

\paragraph{Step 4: transfer the comparison to the actual finite updates.}
The step restriction puts all four coefficients in the preceding
squared-risk formulas in $[1/2,1]$. For $J\in\{\mathrm{full},\mathrm{head}\}$, orthogonality
gives
\[
 \|\widetilde Q_J-f\|^2
 \le\|c\|^2+\|s\|^2=\|f\|^2\le1.
\]
Define the actual remainder functions $r_J=Q_J^+-\widetilde Q_J$.
The probability measure $\nu$ and the finite-step bound imply
\[
 \|r_J\|\le\|r_J\|_\infty
 \le12\eta_{\mathrm{eff}}(d+1)^2\left(\frac1{\sqrt n}+\frac1{\sqrt{b_h}}+\frac1{\sqrt m}\right)\sqrt{\log\!\frac{64Am5^d(1+4m\sqrt d)^d}{\delta}}.
\]
Expansion, the triangle inequality, and Cauchy--Schwarz give
\begin{align*}
 |\mathcal R_\nu(Q_J^+)-\mathcal R_\nu(\widetilde Q_J)|
 &=|2\langle\widetilde Q_J-f,r_J\rangle+\|r_J\|^2|\\
 &\le2\|\widetilde Q_J-f\|\|r_J\|+\|r_J\|^2 \le2\|r_J\|+\|r_J\|^2.
\end{align*}
Subtracting the upper estimate for the full update from the lower
estimate for the frozen-hidden update gives
\begin{align*}
 &\mathcal R_\nu(Q_{\mathrm{head}}^+)-\mathcal R_\nu(Q_{\mathrm{full}}^+)\\
 &\ge\mathcal R_\nu(\widetilde Q_{\mathrm{head}})-\mathcal R_\nu(\widetilde Q_{\mathrm{full}})\\
 &\quad-\sum_{J\in\{\mathrm{full},\mathrm{head}\}}(2\|r_J\|+\|r_J\|^2)\\
 &\ge\frac{\eta_{\mathrm{eff}}}A
       \left[\beta_d^2\|c\|^2+\frac{\alpha_d^2(2d+1)}{d^2}\|s\|^2\right]\\
 &\quad-48\eta_{\mathrm{eff}}(d+1)^2\left(\frac1{\sqrt n}+\frac1{\sqrt{b_h}}+\frac1{\sqrt m}\right)\sqrt{\log\!\frac{64Am5^d(1+4m\sqrt d)^d}{\delta}}\\
 &\quad-288\eta_{\mathrm{eff}}^2(d+1)^4\left(\frac1{\sqrt n}+\frac1{\sqrt{b_h}}+\frac1{\sqrt m}\right)^2\,\log\!\frac{64Am5^d(1+4m\sqrt d)^d}{\delta}.
\end{align*}

For strict improvement, symmetry and Cauchy--Schwarz imply
$\beta_d=\mathbb E|X_1|/2\le(\mathbb E X_1^2)^{1/2}/2=1/2$.
Consequently $\alpha_d=d\beta_d/(d+1)\le1/2$ and
$\alpha_d^2(2d+1)/d^2\le(2d+1)/(4d^2)\le1$ for $d\ge2$.
Orthogonality then gives
\[
 0<\beta_d^2\|c\|^2+\frac{\alpha_d^2(2d+1)}{d^2}\|s\|^2
 \le\|c\|^2+\|s\|^2=\|f\|^2\le1.
\]
The sample condition in Definition~\ref{def:dqn-step-accuracy} implies
\begin{align*}
 &\left(\frac1{\sqrt n}+\frac1{\sqrt{b_h}}+\frac1{\sqrt m}\right)\sqrt{\log\!\frac{64Am5^d(1+4m\sqrt d)^d}{\delta}}\\
 &\le3\sqrt{\frac1{\min\{n,b_h,m\}}\log\!\frac{64Am5^d(1+4m\sqrt d)^d}{\delta}}\\
 &\le\frac{\beta_d^2\|c\|^2+
                   \alpha_d^2(2d+1)\|s\|^2/d^2}{144A(d+1)^2}.
\end{align*}
The first inequality bounds each reciprocal square root by the one
with the smallest denominator. The second takes the positive square
root of the numerical sample condition. Multiplication by
$12\eta_{\mathrm{eff}}(d+1)^2$ therefore gives, for each $J$,
\[
 \|r_J\|\le\frac{\eta_{\mathrm{eff}}}{12A}
      \left[\beta_d^2\|c\|^2+\frac{\alpha_d^2(2d+1)}{d^2}\|s\|^2\right]
 \le1.
\]
Since $\|r_J\|^2\le\|r_J\|$, summing the two error contributions yields
\begin{align*}
 \sum_{J\in\{\mathrm{full},\mathrm{head}\}}(2\|r_J\|+\|r_J\|^2)
 &\le3\sum_{J\in\{\mathrm{full},\mathrm{head}\}}\|r_J\|\\
 &\le\frac{\eta_{\mathrm{eff}}}{2A}
       \left[\beta_d^2\|c\|^2+\frac{\alpha_d^2(2d+1)}{d^2}\|s\|^2\right].
\end{align*}
Subtracting this bound from the population gain leaves at least
half of its positive leading term, proving the strict-risk conclusion.

\paragraph{Step 5: isolate the quadratic prediction signal.}
For any square-integrable critic, let
$(\mathsf Pq)(x,a)=q(x,a)-\mathbb E_\mu q(X,a)$. Expanding each
square and cancelling the mixed terms gives
\begin{align*}
 \|\mathsf Pq\|^2
 &=\frac1A\sum_a\mathbb E_\mu[(q(X,a)-\mathbb E_\mu q(X,a))^2] =\|q\|^2-\frac1A\sum_a(\mathbb E_\mu q(X,a))^2\le\|q\|^2.
\end{align*}
The inequality discards nonnegative squares. Because $\mathsf Pc=0$
and $\mathsf Ps=s$, the centered responses are
\[
 \mathsf P\widetilde Q_{\mathrm{full}}=
          \frac{2\eta_{\mathrm{eff}}\alpha_d^2(d+1)}{Ad^2}s,
 \qquad
 \mathsf P\widetilde Q_{\mathrm{head}}=
          \frac{\eta_{\mathrm{eff}}\alpha_d^2}{Ad^2}s.
\]
Applying the difference-of-squares identity from Step 3 gives
\begin{align*}
 \mathcal R_\nu^{\mathrm{sig}}(\widetilde Q_{\mathrm{head}})
       -\mathcal R_\nu^{\mathrm{sig}}(\widetilde Q_{\mathrm{full}})
 &=\left[\left(1-\frac{\eta_{\mathrm{eff}}\alpha_d^2}{Ad^2}\right)^2
      -\left(1-\frac{2\eta_{\mathrm{eff}}\alpha_d^2(d+1)}{Ad^2}\right)^2\right]\|s\|^2\\
 &=\frac{\eta_{\mathrm{eff}}\alpha_d^2(2d+1)}{Ad^2}
       \left[2-\frac{\eta_{\mathrm{eff}}\alpha_d^2(2d+3)}{Ad^2}\right]\|s\|^2\\
 &\ge\frac{\eta_{\mathrm{eff}}\alpha_d^2(2d+1)}{Ad^2}\|s\|^2.
\end{align*}
The last bracket is at least one by the same step restriction.
The projection contraction gives $\|\mathsf Pr_J\|\le\|r_J\|$,
while $\|\mathsf P\widetilde Q_J-s\|\le\|s\|\le1$.
Expanding the actual centered risk thus yields
\begin{align*}
 |\mathcal R_\nu^{\mathrm{sig}}(Q_J^+)
        -\mathcal R_\nu^{\mathrm{sig}}(\widetilde Q_J)|
 &\le2\|\mathsf P\widetilde Q_J-s\|\|\mathsf Pr_J\|
                                         +\|\mathsf Pr_J\|^2\\
 &\le2\|r_J\|+\|r_J\|^2.
\end{align*}
Subtracting the two errors and then substituting their explicit bounds gives
\begin{align*}
 &\mathcal R_\nu^{\mathrm{sig}}(Q_{\mathrm{head}}^+)
          -\mathcal R_\nu^{\mathrm{sig}}(Q_{\mathrm{full}}^+)\\
 &\ge\frac{\eta_{\mathrm{eff}}\alpha_d^2(2d+1)}{Ad^2}\|s\|^2
              -\sum_{J\in\{\mathrm{full},\mathrm{head}\}}(2\|r_J\|+\|r_J\|^2)\\
 &\ge\frac{\eta_{\mathrm{eff}}\alpha_d^2(2d+1)}{Ad^2}\|s\|^2\\
 &\quad-48\eta_{\mathrm{eff}}(d+1)^2\left(\frac1{\sqrt n}+\frac1{\sqrt{b_h}}+\frac1{\sqrt m}\right)\sqrt{\log\!\frac{64Am5^d(1+4m\sqrt d)^d}{\delta}}\\
 &\quad-288\eta_{\mathrm{eff}}^2(d+1)^4\left(\frac1{\sqrt n}+\frac1{\sqrt{b_h}}+\frac1{\sqrt m}\right)^2\,\log\!\frac{64Am5^d(1+4m\sqrt d)^d}{\delta}.
\end{align*}
Under the centered sample requirement, the same square-root calculation
as in Step 4 gives
\[
 \|r_J\|\le\frac{\eta_{\mathrm{eff}}\alpha_d^2(2d+1)}{12Ad^2}\|s\|^2\le1.
\]
Hence $\sum_J(2\|r_J\|+\|r_J\|^2)\le3\sum_J\|r_J\|
\le\eta_{\mathrm{eff}}\alpha_d^2(2d+1)\|s\|^2/(2Ad^2)$.
Subtracting this error proves the positive centered-risk gain
$\eta_{\mathrm{eff}}\alpha_d^2(2d+1)\|s\|^2/(2Ad^2)$.
\end{proofof}

\begin{proofof}{Corollary~\ref{cor:dqn-bounded-replay-certificate}}
Fix a realization in the intersection of the two uniform events.
The fitted comparison class is contained in the deterministic parent
class specified in Corollary~\ref{cor:dqn-bounded-replay-certificate}.
For an explicit sufficient
box, suppose a deterministic $\eta_{\min}>0$ satisfies
$\eta_{\mathrm{eff}}\ge\eta_{\min}$ on the selected step event.
The preceding readouts then satisfy
\[
 |U_{a,j}|\le\max\!\left\{
 \frac{Ad^2}{\eta_{\min}\alpha_d^2(2d+1)},1+\sqrt{d/m}\right\},
 \qquad
 |b_a|\le\max\!\left\{
 \frac{d^2}{2d+1},1\right\},
\]
and the hidden-row norms are at most $1+\sqrt{d/m}$. Indeed,
\[
 \frac{Ad^2}{\eta_{\mathrm{eff}}\alpha_d^2(2d+1)}
 \le\frac{Ad^2}{\eta_{\min}\alpha_d^2(2d+1)},
 \qquad 1+\eta_{\mathrm{eff}}\sqrt{d/m}\le1+\sqrt{d/m}.
\]
Take the parameter classes satisfying these readout bounds and the
displayed hidden-row bound, allowing every hidden matrix within that
bound. Their union with a deterministic parameter class containing
the target iterates, as in
Lemma~\ref{lem:dqn-parameter-envelope}, is a fixed parent class.
Each component has finite covers by Lemma~\ref{lem:dqn-network-cover}.
The triangle inequality bounds predictions in the two readout boxes by
\[
 \max\!\left\{\frac{d^2}{2d+1},1\right\}
 +\sqrt{md}(1+\sqrt{d/m})
 \max\!\left\{\frac{Ad^2}{\eta_{\min}\alpha_d^2(2d+1)},
                                     1+\sqrt{d/m}\right\}.
\]
Take $B_Q$ at least the maximum of this expression and the
deterministic prediction envelope of the target class.
In the horizon-$H$ configuration, $A=2$ and $\eta_{\min}=1/(4H)$
on the stage-count event, so the head bound is
$\max\{8Hd^2/[\alpha_d^2(2d+1)],1+\sqrt{d/m}\}$.
Thus the parent cover can be fixed before the collection batch.

Each collection record has conditional state--action law
$\mu\otimes\operatorname{Unif}(\mathcal A)=\nu$. Therefore
\[
 \bar\mu_{I,h}=\frac1n\sum_{i\in I}\nu=\nu.
\]
The witness $q^\dagger$ belongs to $\mathcal Q_h(W^+)$. The definition
of an infimum, the unit mass of $\nu$, and
\eqref{eq:dqn-bounded-witness-explicit-error}, respectively, imply
\begin{align*}
 &\inf_{q\in\mathcal Q_h(W^+)}
       \|q-\mathcal T_hv\|_{L^2(\nu)}^2 \le\|q^\dagger-\mathcal T_hv\|_{L^2(\nu)}^2 \le\|q^\dagger-\mathcal T_hv\|_\infty^2\\
 &\le\frac{144A^2d^4(d+1)^4}{\alpha_d^4(2d+1)^2}
 \left(\frac1{\sqrt n}+\frac1{\sqrt{b_h}}+\frac1{\sqrt m}\right)^2
 \log\!\frac{64Am5^d(1+4m\sqrt d)^d}{\delta}.
\end{align*}
The actual updated critic also belongs to $\mathcal Q_h(W^+)$.
Apply Lemma~\ref{lem:dqn-risk-certificate} to this critic, its frozen
value continuation, and the deterministic parent envelope $B=B_Q$. It yields
\begin{align*}
 \|Q_{\mathrm{full}}^+-\mathcal T_hv\|_{L^2(\nu)}^2
 &\le\frac43\left[\widehat L_I(Q_{\mathrm{full}}^+;v)
      -\inf_{q\in\mathcal Q_h(W^+)}\widehat L_I(q;v)\right]\\
 &\quad+\frac53\inf_{q\in\mathcal Q_h(W^+)}
                         \|q-\mathcal T_hv\|_{L^2(\nu)}^2\\
 &\quad+\frac{96(2B_Q+1)^2}{n}
          \log\!\frac{2HT^2N_Q(u)N_V(u)}{\delta}
          +\frac{104}{3}(2B_Q+1)u.
\end{align*}
Substituting the preceding approximation inequality proves the
explicit replay-risk conclusion. If the source events have failure
probabilities $\delta_{\mathrm{step}}$ and $\delta_{\mathrm{replay}}$,
the union bound gives probability at least
$1-\delta_{\mathrm{step}}-\delta_{\mathrm{replay}}$ for their intersection.
\end{proofof}

\begin{proofof}{Theorem~\ref{thm:dqn-horizon-complete}}
Work on the event of Lemma~\ref{lem:dqn-reachable-budget},
applied with confidence $\delta$. It has probability at least
$1-\delta$ and contains the neural tracking event of
Lemma~\ref{lem:dqn-horizon-iterates}.

\paragraph{The collection batches and the actual neural updates.}
Lemma~\ref{lem:dqn-horizon-schedule} implements the stated schedule
inside Algorithm~\ref{alg:dqn-online}. Each collection batch
supplies exactly $n$ records per stage. One positive global SGD
step uses ordinary uniform replay from that buffer, and the
target is then copied. The targets in
\eqref{eq:dqn-horizon-actual-targets} use the preceding update's
critics. Stage $h$ remains at its paired initialization until
step $H-h+1$. Previously reached stages continue to train their
hidden weights, action heads, and biases in the same global steps.

\paragraph{Coupled propagation and the final prediction.}
The proof of Lemma~\ref{lem:dqn-horizon-iterates} follows all $K$
updates using the initial population tangent operator. Before
each fresh collection batch, both the current critic and copied
target are fixed.
The reference $q_{h,j}$ uses the conditional mean of the actual
target at every step and satisfies
\[
 q_{h,j}(x,a)=c_{h,j}+a u_{h,j}(x),\qquad
 \|Q_{h,j}-q_{h,j}\|_\infty\le E_{\mathrm{run}}
 \quad(h\in[H],\ 0\le j\le K).
\]

Let $A_{h,j}=\|u_{h,j}-\beta_h^\star g_{h,1}\|_\infty$,
$A_{H+1,j}=0$, and $\omega_{h,j}=\lambda_2p_{h,j}/4$.
Common value continuation offsets cancel under the signed transition,
so the stage errors obey
\stepcounter{equation}
\begin{equation*}
 A_{h,j}\le(1-\omega_{h,j})A_{h,j-1}
       +\frac{\omega_{h,j}}{4D}
                  (A_{h+1,j-1}+E_{\mathrm{run}}).
\end{equation*}
Since $D\ge2$ and
$\omega_{h,j}\ge\lambda_2/(8H)$, it gives
\[
 \max_h A_{h,K}
 \le\beta_H^\star
          \exp\!\left(-\frac{7\lambda_2K}{64H}\right)
             +\frac17E_{\mathrm{run}}.
\]
The tracking proof compares the constant components with a bounded
population trajectory. Their perturbations accumulate at most in
proportion to $K/H$, and the explicit budget closes the output envelope.

Set $\gamma_h=c_h^\star-c_{h,K}$. The tracking lemma gives
\stepcounter{equation}
\begin{equation*}
 |\gamma_h|<2,\qquad
 \|Q_{h,K}+\gamma_h-Q_h^\star\|_\infty
 \le\exp\!\left(-\frac{\lambda_2K}{16H}\right)
                    +2E_{\mathrm{run}}=e.
\end{equation*}
Adding a common constant to the two action values preserves their
order and ties, and therefore preserves the executed greedy policy.
The analytical shift calibrates the value level by changing both
output biases by the same amount.
Squaring the uniform error and integrating against any probability
measure proves the stated squared-risk bound. Substituting
\eqref{eq:dqn-long-error}, using $K/H\ge1$, and observing that
\[
 \ell=\mathcal O\!\left(d\log(md)+\log\frac{HK}{\delta}\right)
\]
gives \eqref{eq:dqn-horizon-main-error} with universal constants.
\end{proofof}

\begin{proofof}{Corollary~\ref{cor:dqn-complete-rates}}
Work on the joint event of Lemma~\ref{lem:dqn-reachable-budget},
applied with confidence $\delta$. It has probability at least
$1-\delta$ and contains both the tracking and feature conclusions.
The tracking lemma's
data and width errors have the form
\begin{align*}
 E_{\mathrm{run}}\le C(1+\sqrt{(d+1)K/H})\bigg[
 &\frac{(d+1)K}{H}\sqrt\ell
          \left(n^{-1/2}+\sqrt{2H/b}\right)\\
 &+(d+1)^2\left(\frac KH+\frac{K^2}{H^2}\right)
                                           \sqrt{\ell/m}\bigg].
\end{align*}
Since $b=Hn$ and
$\sqrt m\ge\sqrt{C_2}(d+1)(1+K/H)\sqrt n$, the width term is
bounded by a universal multiple of the data term. Also $K/H\ge1$.
Consequently
\begin{equation}\label{eq:dqn-rate-error-coefficient}
 E_{\mathrm{run}}\le
       C(d+1)^{3/2}(K/H)^{3/2}\sqrt{\ell/n}.
\end{equation}

The identities $\mathbb E X_1^2=1$ and $\mathbb E X_1^4\le3$
imply $\mathbb E|X_1|\ge1/\sqrt3$ by H\"older's inequality.
Therefore $\alpha_d\ge2/(5\sqrt3)$ for $d\ge4$, and
$\lambda_2^{-1}\le10d$. Conversely $\alpha_d\le1/2$ gives
\[
 \lambda_2\le\frac{d+1}{2d^2}\le\frac1{d+1},
 \qquad d\ge4.
\]
In particular $d+1\le\lambda_2^{-1}\le10d$.
For $Kn\le T$, the displayed schedules imply
\[
 K/H\le C\lambda_2^{-1}\Gamma_T,\qquad
 m\le C (d+1)^2(1+(d+1)\Gamma_T)^2T,
 \qquad\ell\le C (d+1)\Gamma_T.
\]
Indeed the tracking logarithm contains only polynomial factors
in $H,K,m$ and a sphere net of exponent proportional to $d$.
Every logarithm of these factors is at most $C\Gamma_T$.

The contrast contraction factor is at most
\[
 \exp\!\left[-\frac{7\lambda_2K}{64H}\right]
       \le e^{-\Gamma_T}
       \le\frac{\delta}{C_0H(d+1)(T+1)}.
\]
The proof of Theorem~\ref{thm:dqn-horizon-complete} gives
$e\le e^{-\Gamma_T}+2E_{\mathrm{run}}$.
Its action-independent shifts preserve the deployed greedy policy.
Lemma~\ref{lem:dqn-quadratic-policy-loss} therefore bounds that
policy's one-episode loss by $8\sqrt d\,e$.
Each collection episode contributes at most one to regret, so
\begin{align*}
 \Reg_T
 &\le Kn+8\sqrt d\,T e \le Kn+1
       +C (d+1)^2(K/H)^{3/2}\sqrt\ell\,
                               \frac{T}{\sqrt n}\\
 &\le Kn+1
       +C (d+1)^{5/2}\lambda_2^{-3/2}\Gamma_T^2\,
                               \frac{T}{\sqrt n}.
\end{align*}
The sample choice balances $Kn$ with the last term. Since
$K=\mathcal O(H\lambda_2^{-1}\Gamma_T)$, both terms are bounded by
\[
 C(d+1)^{5/3}\lambda_2^{-4/3}\Gamma_T^{5/3}H^{1/3}T^{2/3}.
\]
Together with the deterministic cap $T$, this proves
\eqref{eq:dqn-complete-two-thirds-rate}. Substituting $n$ and $K/H$ in the definition of
$m$ proves the width bound, and the update count is $K$ by the
executable schedule.

\paragraph{The numerical training budget.}
The raw-output bootstrap requires
$(K/H+1)E_{\mathrm{run}}\le c$ for a universal numerical constant
$c>0$. Equation~\eqref{eq:dqn-rate-error-coefficient} shows that
\stepcounter{equation}
\begin{equation*}
 n\ge C (d+1)^3 (K/H)^5\ell
\end{equation*}
suffices, after enlarging the universal constant because $K/H\ge1$.
The width choice, $n\ge1$, and a sufficiently large $C_2$ ensure
the tracking width condition
$m\ge C (d+1)^2(1+K/H)^2$, and the count row follows from
$n\ge C\log(HK/\delta)$. The additional feature rows are also
covered by the same floor: $D\le d(d+1)/2$, the Gram-width
condition is of order $dD+d\log(H/\delta)$, and
$n\ge C d\ell$ ensures both feature sampling rows when $b=Hn$.

Using $K/H\le C\lambda_2^{-1}\Gamma_T$ and
$\ell\le C (d+1)\Gamma_T$, an explicit sufficient collection size is
\[
 n\ge n_0:=C (d+1)^4\lambda_2^{-5}\Gamma_T^6.
\]
For the optimized $n$, this condition follows from
\[
 \frac TH\ge C (d+1)^{7/2}\lambda_2^{-7}\Gamma_T^8.
\]
The availability condition $Kn\le T$ separately follows from
$T/H\ge C (d+1)^5\lambda_2^{-4}\Gamma_T^5$.
It is also implied by the preceding sufficient threshold, because
$\lambda_2^{-1}\ge d+1$ and $\Gamma_T\ge1$. The horizon dependence
of these sufficient sample and episode thresholds enters through
the ratio $T/H$ and the logarithm $\Gamma_T$.
\end{proofof}

\subsection{Proofs of the Technical Lemmas}\label{sec:lat-technical-proofs}

\begin{proofof}{Lemma~\ref{lem:dqn-horizon-feature-gain}}
Work on the common event of Lemma~\ref{lem:dqn-reachable-budget},
applied with confidence $\delta$.
Lemma~\ref{lem:dqn-horizon-schedule} shows that stage $h$ remains
at its paired initialization until $j_h=H-h+1$, independently of
the copied continuation and the fresh collection records.
At this update, write
$C_{h,a}(\widehat v_h)=aC$ and let $c$ denote the common target
offset. In the notation of the proof of
Theorem~\ref{thm:dqn-feature-gain},
\[
 M_j=\frac{U_{+,j}^{(0)}-U_{-,j}^{(0)}}2C,
 \qquad
 \chi_j=\frac{U_{+,j}^{(0)}+U_{-,j}^{(0)}}2c,
 \qquad
 \chi_jM_j=0.
\]
The last identity follows from the Rademacher initial heads.
Thus the common-offset contribution to the feature-covariance
calculation vanishes exactly. The signed-action adaptation of
the finite-step and concentration estimates is established in
Lemma~\ref{lem:dqn-reachable-budget}. In particular,
\eqref{eq:dqn-reachable-gain} gives
\[
 \mathcal G_h\ge
 \frac{\alpha_d^2S_h}{16Hd}
 -\frac{2d^2\ell}{H}
       \left(\frac1n+\frac Hb\right),
\]
which proves \eqref{eq:dqn-horizon-main-gain}.
Under \eqref{eq:dqn-feature-signal-resolution}, the sharper
finite-step estimate gives \eqref{eq:dqn-resolved-feature-gain}.
The terminal and first nonterminal certificates follow from
\eqref{eq:dqn-terminal-feature-gain} and
\eqref{eq:dqn-first-nonterminal-feature-gain}, respectively,
under their stated sampling conditions.
\end{proofof}

\begin{proofof}{Lemma~\ref{lem:qmdp-sphere-moments}} We formulate the proof of Lemma~\ref{lem:qmdp-sphere-moments} below:
\paragraph{Step 1: second moments.}
Sign symmetry gives $\mathbb E X_iX_j=0$ for $i\ne j$.
Coordinate symmetry makes the $d$ diagonal second moments equal.
Since $\sum_iX_i^2=d$ almost surely,
\[
 \sum_{i=1}^d\mathbb E X_i^2=d,
 \qquad \mathbb E X_i^2=1,
 \qquad \mathbb E XX^{\top}=I_d.
\]
Therefore,
\[
 \mathbb E g_{h,\ell}(X)
 =\frac1d\operatorname{tr}
       (E_{h,\ell}\mathbb E XX^{\top})
 =\frac1d\operatorname{tr}(E_{h,\ell})=0.
\]

\paragraph{Step 2: fourth moments.}
Put $U=X/\sqrt d$, so that $U$ is uniform on the unit sphere.
Apply the spherical Wick formula \citep[Theorem~2]{dqn_vignat2008wick}
with ambient dimension $d$ and the four coordinate linear forms
$U_i,U_j,U_k,U_l$. Its three pairings are
$(i,j)(k,l)$, $(i,k)(j,l)$, and $(i,l)(j,k)$, and its coefficient is
\[
 \frac{\Gamma(d/2)}{2^2\Gamma(2+d/2)}
 =\frac{1}{d(d+2)}.
\]
The inner products of coordinate vectors give the corresponding
indicators. Scaling the four factors by $\sqrt d$ therefore yields
\[
\begin{aligned}
 \mathbb E X_iX_jX_kX_l
 &=d^2\mathbb E U_iU_jU_kU_l\\
 &=\frac{d}{d+2}
   (\mathbf1\{i=j\}\mathbf1\{k=l\}
    +\mathbf1\{i=k\}\mathbf1\{j=l\}
    +\mathbf1\{i=l\}\mathbf1\{j=k\}).
\end{aligned}
\]
Thus, for symmetric matrices $A,C$,
\begin{align*}
 \mathbb E[(X^{\top}AX)(X^{\top}CX)]
 &=\sum_{i,j,k,l}A_{ij}C_{kl}\mathbb E X_iX_jX_kX_l\\
 &=\frac{d}{d+2}
       \{\operatorname{tr}(A)\operatorname{tr}(C)
           +2\operatorname{tr}(AC)\}.
\end{align*}
Applying \eqref{eq:qmdp-basis} and dividing by $d^2$ yields
\[
 \mathbb E[g_{h,\ell}(X)g_{h,k}(X)]
 =\frac{2}{d(d+2)}\mathbf1\{\ell=k\},
\]
which is the covariance identity in
\eqref{eq:qmdp-coordinate-moments}.

\paragraph{Step 3: a uniform norm bound.}
Use the Frobenius inner product and put
$F_x=(xx^{\top}-I_d)/d$.  Trace-freeness gives
$g_{h,\ell}(x)=\langle E_{h,\ell},F_x\rangle_F$.
Orthonormality and Bessel's inequality imply
\begin{align*}
 \|g_h(x)\|_2^2
 &=\sum_{\ell=1}^r\langle E_{h,\ell},F_x\rangle_F^2
 \le\|F_x\|_F^2\\
 &=\frac1{d^2}
       \{\operatorname{tr}(xx^{\top}xx^{\top})
                    -2\operatorname{tr}(xx^{\top})+\operatorname{tr}(I_d)\} =\frac{d^2-2d+d}{d^2}=1-1/d.
\end{align*}
Here $\operatorname{tr}(xx^{\top})=\|x\|_2^2=d$ and
$\operatorname{tr}(xx^{\top}xx^{\top})=\|x\|_2^4=d^2$.

\paragraph{Step 4: Lipschitz continuity.}
Applying Bessel's inequality to $xx^{\top}-yy^{\top}$ gives
\begin{align*}
 \|g_h(x)-g_h(y)\|_2
 &\le\frac1d\|xx^{\top}-yy^{\top}\|_F =\frac1d\|(x-y)x^{\top}+y(x-y)^{\top}\|_F\\
 &\le\frac{\|x-y\|_2\|x\|_2+\|y\|_2\|x-y\|_2}{d} =\frac{2}{\sqrt d}\|x-y\|_2
 \le2\|x-y\|_\infty.
\end{align*}
The penultimate equality uses the sphere radius; the last inequality
uses $\|x-y\|_2\le\sqrt d\|x-y\|_\infty$.
Each coordinate difference is at most the Euclidean norm of the vector
difference, proving the coordinatewise assertion as well.
\end{proofof}

\begin{proofof}{Lemma~\ref{lem:qmdp-exact-rank}} We formulate the proof of Lemma~\ref{lem:qmdp-exact-rank} below:
\paragraph{Step 1: the weights form a probability vector.}
Equation~\eqref{eq:qmdp-simplex-basis} gives
$\|B_{j,:}^{\top}\|_2^2=(BB^{\top})_{jj}=1-1/D\le1$.
Lemma~\ref{lem:qmdp-sphere-moments} and the contraction property of
$T_{h,a}$ give $\|[T_{h,a}g_h(x)]\|_2\le1$.  Hence
\[
 |B_{j,:}[T_{h,a}g_h(x)]|
 \le\|B_{j,:}^{\top}\|_2\|[T_{h,a}g_h(x)]\|_2\le1,
 \qquad
 \frac{1-\tau_h}{D}\le\phi_{h,j}^\star(x,a)
 \le\frac{1+\tau_h}{D}.
\]
In particular every weight is positive.  Also,
\begin{align*}
 \sum_{j=1}^D\phi_{h,j}^\star(x,a)
 &=1+\frac{\tau_h}{D}
          \sum_{j=1}^D B_{j,:}[T_{h,a}g_h(x)] =1,
\end{align*}
because $\sum_jB_{j,:}^{\top}=B^{\top}\mathbf1=0$.

\paragraph{Step 2: the components are probability laws.}
The same norm argument gives
$1-\omega_h\le1+\omega_h B_{j,:}g_{h+1}(x')\le1+\omega_h$.
The density is strictly positive.  Its integral is
\[
 \int\{1+\omega_h B_{j,:}g_{h+1}(x')\}\mu(dx')
 =1+\omega_h B_{j,:}\mathbb E_\mu g_{h+1}(X')=1.
\]
Thus \eqref{eq:qmdp-emission} is a probability measure.
Mixtures with the weights from Step 1 are consequently valid kernels.
Their densities and the mean reward are polynomials restricted to the
sphere. The mean reward is a convex combination of entries of
$\vartheta_h$, so it lies in $[0,r_h^{\max}]$. Both reward laws
have this support bound, and $\sum_h r_h^{\max}\le1$ bounds every
episode's return by one.

\paragraph{Step 3: expand the mixture density.}
Expanding all four terms of the product gives
\begin{align*}
 &\sum_{j=1}^D\frac{1+\tau_h B_{j,:}T_{h,a}g_h(x)}{D}
                  [1+\omega_h B_{j,:}g_{h+1}(x')]\\
 &=1+\frac{\tau_h}{D}\left(\sum_j B_{j,:}\right)T_{h,a}g_h(x)
       +\frac{\omega_h}{D}\left(\sum_j B_{j,:}\right)g_{h+1}(x')\\
 &\qquad+\frac{\tau_h\omega_h}{D}
           [T_{h,a}g_h(x)]^{\top}\left(\sum_j B_{j,:}^{\top}B_{j,:}\right)g_{h+1}(x')\\
 &=1+\frac{\tau_h\omega_h}{D}[T_{h,a}g_h(x)]^{\top}g_{h+1}(x').
\end{align*}
The sums in the linear terms vanish, and the matrix in the quadratic
term is $B^{\top}B=I_r$. This proves \eqref{eq:qmdp-density}.
Cauchy--Schwarz and the coordinate norm bound give
\[
 |[T_{h,a}g_h(x)]^{\top}g_{h+1}(x')|
 \le\|T_{h,a}g_h(x)\|_2\|g_{h+1}(x')\|_2\le1,
\]
which proves the stated density interval.

\paragraph{Step 4: the transition rank is at least $D$.}
Fix an action $a_0$ for which $T_{h,a_0}$ is invertible.  For the
transition operator acting on bounded functions, equation
\eqref{eq:qmdp-density} gives
\begin{align*}
 [P_h1](x,a_0)&=1,\\
 [P_hg_{h+1,k}](x,a_0)
 &=\mathbb E_\mu g_{h+1,k}(X')
     +\frac{\tau_h\omega_h}{D} [T_{h,a_0}g_h(x)]^{\top}
       \mathbb E_\mu[g_{h+1}(X')g_{h+1,k}(X')]\\
 &=\frac{2\tau_h\omega_h}{Dd(d+2)} [T_{h,a_0}g_h(x)]_k,
       \qquad k\in[r].
\end{align*}
These $r+1$ image functions are linearly independent.  Indeed, if
$c_0+c^{\top}[T_{h,a_0}g_h(x)]=0$ $\mu$-almost surely, integration gives
$c_0=0$, because $\mathbb E_\mu T_{h,a_0}g_h=0$.  Squaring and integrating
the remaining equality gives
\begin{align*}
 0
 &=\mathbb E_\mu(c^{\top}T_{h,a_0}g_h(X))^2 ={\frac{2}{d(d+2)}}\|T_{h,a_0}^{\top}c\|_2^2.
\end{align*}
Since $2/[d(d+2)]>0$ and $T_{h,a_0}$ is invertible, this forces $c=0$.
Also $\frac{2\tau_h\omega_h}{Dd(d+2)}>0$, so multiplication by this scalar preserves
linear independence.  The transition image therefore has dimension at
least $r+1=D$.

\paragraph{Step 5: the transition rank is at most $D$.}
The mixture representation gives
\[
 [P_hf](x,a)=\sum_{j=1}^D\phi_{h,j}^\star(x,a)
                         \int f\,d\nu_{h,j}.
\]
Every image belongs to the span of the $D$ coordinate functions
$\phi_{h,j}^\star$.  This proves rank at most $D$, and hence exact
rank $D$.  A factorization using fewer than $D$ coordinates would put
this image in a lower-dimensional span, contradicting Step 4.

\paragraph{Step 6: verify the linear-MDP norm bounds.}
For a probability vector, $\|\phi_h^\star\|_2\le\|\phi_h^\star\|_1=1$.
Every component measure has total mass one, so the vector of total
masses has Euclidean norm $\sqrt D$.
Finally,
$\|\vartheta_h\|_2^2\le D(r_h^{\max})^2\le D$,
which gives the stated reward-parameter norm bound.
At the terminal stage, $\sum_j\phi_{H,j}^\star=1$ implies
$\sum_j\phi_{H,j}^\star\delta_\partial=\delta_\partial$,
which completes the terminal factorization.
\end{proofof}

\begin{proofof}{Lemma~\ref{lem:qmdp-bellman-signal}} We formulate the proof of Lemma~\ref{lem:qmdp-bellman-signal} below:
The coefficients are
\stepcounter{equation}
\begin{equation*}
 c_h(V)=\frac1D\mathbf1^{\top}\vartheta_h+\mathbb E_\mu V(X').
\end{equation*}
and
\begin{equation}\label{eq:qmdp-signal-matrix}
 C_{h,a}(V)
 =\frac{\tau_h}{Dd}
   \sum_{\ell=1}^r
   \left[T_{h,a}^{\top}
       \{B^{\top}\vartheta_h+\omega_h \mathbb E_\mu[g_{h+1}(X')V(X')]\}\right]_\ell E_{h,\ell}.
\end{equation}

For the reward, \eqref{eq:qmdp-simplex-feature} gives
\begin{align*}
 r_h(x,a)
 &=\sum_{j=1}^D\frac{1+\tau_h B_{j,:}[T_{h,a}g_h(x)]}{D}
                         \vartheta_{h,j}\\
 &=\frac1D\mathbf1^{\top}\vartheta_h+
          \frac{\tau_h}{D}[T_{h,a}g_h(x)]^{\top}B^{\top}\vartheta_h.
\end{align*}
Integrating the transition density gives
\begin{align*}
 [P_hV](x,a)
 &=\int V(x')
       \{1+\frac{\tau_h\omega_h}{D} [T_{h,a}g_h(x)]^{\top}g_{h+1}(x')\}\mu(dx')\\
 &=\mathbb E_\mu V(X')+\frac{\tau_h\omega_h}{D}
                           [T_{h,a}g_h(x)]^{\top}\mathbb E_\mu[g_{h+1}(X')V(X')].
\end{align*}
Adding these equalities and using the quadratic coordinate formula yields
\begin{align*}
 &r_h(x,a)+[P_hV](x,a)\\
 &=\frac1D\mathbf1^{\top}\vartheta_h+\mathbb E_\mu V(X')
   +\frac{\tau_h}{D}g_h(x)^{\top}T_{h,a}^{\top}
                 \{B^{\top}\vartheta_h+\omega_h \mathbb E_\mu[g_{h+1}(X')V(X')]\}\\
 &=c_h(V)+\frac{\tau_h}{Dd}\sum_{\ell=1}^r
       [T_{h,a}^{\top}\{B^{\top}\vartheta_h+\omega_h \mathbb E_\mu[g_{h+1}(X')V(X')]\}]_\ell
                                      x^{\top}E_{h,\ell}x\\
 &=c_h(V)+x^{\top}C_{h,a}(V)x.
\end{align*}
This proves \eqref{eq:qmdp-affine-quadratic-target} and
\eqref{eq:qmdp-signal-matrix}.  Trace-freeness and symmetry follow
because each $E_{h,\ell}$ has those properties.

For the bounds, first
\[
 |\mathbb E_\mu V(X')|\le\mathbb E_\mu|V(X')|\le B_V,
 \qquad 0\le\frac1D\mathbf1^{\top}\vartheta_h\le r_h^{\max}.
\]
Also,
\begin{align*}
 &\|\mathbb E_\mu[g_{h+1}(X')V(X')]\|_2 \quad\le\mathbb E_\mu[\|g_{h+1}(X')\|_2|V(X')|]
 \le B_V,\\
 &\|B^{\top}\vartheta_h+\omega_h \mathbb E_\mu[g_{h+1}(X')V(X')]\|_2\\
 &\quad\le\|B^{\top}\|_{\mathrm{op}}\|\vartheta_h\|_2
                       +\omega_h\|\mathbb E_\mu[g_{h+1}(X')V(X')]\|_2 \quad\le\sqrt D\,r_h^{\max}+\omega_h B_V.
\end{align*}
Here $\|B^{\top}\|_{\mathrm{op}}=1$ follows from
$B^{\top}B=I_r$.
Frobenius orthonormality of the $E_{h,\ell}$ gives the exact identity
\[
 \|C_{h,a}(V)\|_F
 =\frac{\tau_h}{Dd}
   \|T_{h,a}^{\top}(B^{\top}\vartheta_h+\omega_h \mathbb E_\mu[g_{h+1}(X')V(X')])\|_2.
\]
Using $\|T_{h,a}\|_{\mathrm{op}}\le1$ and the preceding vector bound
proves \eqref{eq:qmdp-signal-bounds}.  Finally, setting
$\vartheta_h=0$ removes the reward term from
\eqref{eq:qmdp-signal-matrix}.
\end{proofof}

\begin{proofof}{Lemma~\ref{lem:qmdp-two-layer-approximation}} We formulate the proof of Lemma~\ref{lem:qmdp-two-layer-approximation} below:
For the independent directions $v_1,\ldots,v_{m_0}$, use
\begin{equation}\label{eq:qmdp-relu-approximation}
 \mathcal Q_{m_0,C,c}(x)
 =c+\frac{d^2}{{m_0}\alpha_d}\sum_{j=1}^{m_0}
                     (v_j^{\top}Cv_j)\sigma(v_j^{\top}x),
\end{equation}
The required width is
\begin{equation}\label{eq:qmdp-approximation-width}
 {m_0}\ge\frac{32M^2d^5}{\alpha_d^2\varepsilon^2}
 \left\{r\log5+d\log\!\left(1+
             \frac{8Md^{5/2}}{\alpha_d\varepsilon}\right)
                         +\log\frac2\delta\right\}.
\end{equation}

\paragraph{Step 1: an exact spherical integral.}
Let $W$ be uniform on the unit sphere, and put
\[
 a_1=\mathbb E[W_1\mathbf1\{W_1>0\}],\qquad
 a_3=\mathbb E[W_1^3\mathbf1\{W_1>0\}].
\]
The first coordinate has density
\[
 t\longmapsto
 \frac{\Gamma(d/2)}{\sqrt\pi\Gamma((d-1)/2)}
                  (1-t^2)^{(d-3)/2},\qquad -1<t<1.
\]
Writing its normalizing constant as $c_d$, the substitution $s=t^2$
gives
\begin{align*}
 a_1
 &=c_d\int_0^1t(1-t^2)^{(d-3)/2}\,dt
   =\frac{c_d}{2}\int_0^1(1-s)^{(d-3)/2}\,ds =\frac{c_d}{d-1}
   =\frac{\Gamma(d/2)}{2\sqrt\pi\Gamma((d+1)/2)},\\
 a_3
 &=\frac{c_d}{2}\int_0^1s(1-s)^{(d-3)/2}\,ds =\frac{\Gamma(d/2)}{2\sqrt\pi\Gamma((d+3)/2)}
   =\frac{2a_1}{d+1}.
\end{align*}
The integral in the second calculation is
$\operatorname{B}(2,(d-1)/2)$, and the last equalities use
$\Gamma(t+1)=t\Gamma(t)$.
By symmetry among coordinates $2,\ldots,d$ and
$\sum_{i=2}^dW_i^2=1-W_1^2$,
\begin{align*}
 \mathbb E[W_i^2W_1\mathbf1\{W_1>0\}]
 &=\frac{a_1-a_3}{d-1}
   =\frac{a_1}{d+1},\qquad i\ge2.
\end{align*}
For $x=\sqrt d\,e_1$, all off-diagonal terms in the quadratic form
have zero expectation by coordinate sign symmetry.  Trace-freeness
then gives
\begin{align*}
 \mathbb E[(W^{\top}CW)\sigma(W^{\top}x)]
 &=\sqrt d\left(C_{11}a_3+
                 \sum_{i=2}^dC_{ii}\frac{a_1}{d+1}\right) =\sqrt d\,C_{11}
                   \left(a_3-\frac{a_1}{d+1}\right)\\
 &=\frac{\sqrt d\,a_1}{d+1}C_{11}
   =\frac{a_1}{\sqrt d(d+1)}x^{\top}Cx
   ={\frac{\alpha_d}{d^2}} x^{\top}Cx.
\end{align*}
For an arbitrary $x\in\mathbb S_d$, rotate $x/\sqrt d$ to $e_1$.
Orthogonal invariance of $W$ and invariance of the trace give the same
identity.  Therefore,
\begin{equation}\label{eq:qmdp-relu-integral}
 \mathbb E[(W^{\top}CW)\sigma(W^{\top}x)]
                         ={\frac{\alpha_d}{d^2}} x^{\top}Cx.
\end{equation}

\paragraph{Step 2: finite coverings.}
Choose a Frobenius $M/2$-net $\mathcal C$ of the radius-$M$ ball in
$\mathcal S$ with $|\mathcal C|\le5^r$.
Choose a Euclidean $\rho$-net $\mathcal X$ of $\mathbb S_d$, where
$\rho=\alpha_d\varepsilon/(4Md^2)$, with
\[
 |\mathcal X|\le(1+2\sqrt d/\rho)^d
             =\left(1+\frac{8Md^{5/2}}{\alpha_d\varepsilon}\right)^d.
\]
Both bounds follow from the same packing argument: the disjoint balls
of radius half the net separation fit inside a ball whose radius is
the domain radius plus half that separation.  Comparing volumes gives
$(1+2R/\rho)^k$ in a $k$-dimensional radius-$R$ ball.  A maximal
separated set supplies a covering.

\paragraph{Step 3: concentration on the product net.}
For $\|C\|_F\le M$, define
\[
 Z_j(C,x)=(v_j^{\top}Cv_j)\sigma(v_j^{\top}x).
\]
The inequalities $\|C\|_{\mathrm{op}}\le\|C\|_F$,
$\|v_j\|_2=1$, and $\|x\|_2=\sqrt d$ imply
\[
 |Z_j(C,x)|\le M\sqrt d.
\]
Hoeffding's inequality for independent variables in an interval of
length $2M\sqrt d$ gives
\begin{align*}
 \mathbb P\left(\left|
     \frac1{m_0}\sum_jZ_j(C,x)-\mathbb E Z_j(C,x)
             \right|>\frac{\alpha_d\varepsilon}{4d^2}\right)
 &\le2\exp\left\{-\frac{2{m_0}(\alpha_d\varepsilon/(4d^2))^2}
                                      {(2M\sqrt d)^2}\right\}\\
 &=2\exp\left\{-\frac{{m_0}\alpha_d^2\varepsilon^2}{32M^2d^5}\right\}.
\end{align*}
A union bound over $\mathcal C\times\mathcal X$ and
\eqref{eq:qmdp-approximation-width} show that every such net deviation
is at most $\alpha_d\varepsilon/(4d^2)$ on an event of probability at least
$1-\delta$.

\paragraph{Step 4: extend the event to every matrix.}
Write the unscaled deviation as
\[
 E_{m_0}(C,x)=\frac1{m_0}\sum_j Z_j(C,x)-\mathbb E Z_j(C,x).
\]
It is linear in $C$.  Fix $x\in\mathcal X$ and let
$S_x=\sup_{C\in\mathcal S,\|C\|_F\le M}|E_{m_0}(C,x)|$.
For every $C$ in that ball, select $C_0\in\mathcal C$ with
$\|C-C_0\|_F\le M/2$.  Linearity and homogeneity give
\begin{align*}
 |E_{m_0}(C,x)|
 &\le|E_{m_0}(C_0,x)|+|E_{m_0}(C-C_0,x)| \le\alpha_d\varepsilon/(4d^2)+S_x/2.
\end{align*}
Taking the supremum yields $S_x\le\alpha_d\varepsilon/(4d^2)+S_x/2$.
Subtracting $S_x/2$ and multiplying by two gives
$S_x\le\alpha_d\varepsilon/(2d^2)$.

\paragraph{Step 5: extend to every observation.}
For fixed $C$ with $\|C\|_F\le M$, ReLU's one-Lipschitz property
implies
\[
 |Z_j(C,x)-Z_j(C,y)|
 \le |v_j^{\top}Cv_j|\,|v_j^{\top}(x-y)|
 \le M\|x-y\|_2.
\]
Both the empirical average and its expectation are therefore
$M$-Lipschitz, so their difference is $2M$-Lipschitz.
For $x\in\mathbb S_d$, choose $x_0\in\mathcal X$ with
$\|x-x_0\|_2\le\rho$.  Then
\begin{align*}
 |E_{m_0}(C,x)|
 &\le |E_{m_0}(C,x_0)|+2M\|x-x_0\|_2 \le\alpha_d\varepsilon/(2d^2)+2M\rho
  =\alpha_d\varepsilon/d^2.
\end{align*}
Divide by $\alpha_d/d^2>0$ and use \eqref{eq:qmdp-relu-integral} to obtain
\eqref{eq:qmdp-approximation-uniform}.  The offset $c$ cancels exactly.

\paragraph{Step 6: identify the ordinary critic parameters.}
In the width-${m_0}$ parameterization
$Q(x)=c+{m_0}^{-1/2}\sum_j a_j\sigma(w_j^{\top}x)$, choose
\[
 w_j=v_j,\qquad
 a_j=\frac{d^2v_j^{\top}Cv_j}{\alpha_d\sqrt {m_0}}.
\]
This is exactly \eqref{eq:qmdp-relu-approximation}.
For paired width $m=2{m_0}$, set $w_{j+m_0}=w_j=v_j$ and choose both
head entries equal to
$d^2(v_j^{\top}Cv_j)/(\alpha_d\sqrt{2{m_0}})$.
The pair contributes
\[
 \frac1{\sqrt{2{m_0}}}
       \frac{2d^2v_j^{\top}Cv_j}{\alpha_d\sqrt{2{m_0}}}
                     \sigma(v_j^{\top}x)
 =\frac{d^2v_j^{\top}Cv_j}{{m_0}\alpha_d}\sigma(v_j^{\top}x).
\]
Summing over the ${m_0}$ pairs gives the same network.
\end{proofof}

\begin{proofof}{Lemma~\ref{lem:qmdp-state-marginals}} 
The first assertion is the reset rule.  Suppose that $X_h\sim\mu$
and the action probabilities are constants $p_{h,a}$ independent of
$X_h$.  Then
\begin{align*}
 \mathbb E[T_{h,A_h}g_h(X_h)]
 &=\sum_{a\in\mathcal A}p_{h,a}T_{h,a}\mathbb E_\mu g_h(X_h)
 =0.
\end{align*}
Integrating \eqref{eq:qmdp-density} over $(X_h,A_h)$ consequently
gives value density one.  Starting from $X_1\sim\mu$ and
repeating this argument proves the stage-marginal assertion.
The same argument holds conditional on the pre-episode history when
the probabilities are chosen using that history.

For a general policy, integrate \eqref{eq:qmdp-density} under the
joint distribution of the action and observation. This gives
\eqref{eq:qmdp-adaptive-marginal} directly.  Jensen's inequality and
the pointwise norm bound give
\[
 \|\mathbb E[T_{h,A_h}g_h(X_h)]\|_2
 \le\mathbb E\|T_{h,A_h}g_h(X_h)\|_2\le1.
\]
Cauchy--Schwarz then gives the density interval.
\end{proofof}

\begin{proofof}{Lemma~\ref{lem:dqn-feature-moments}}
Rotation invariance first allows us to take $w=e_1$. If $Z=X/\sqrt d$
is uniform on the unit sphere, its first coordinate has density
proportional to $(1-s^2)^{(d-3)/2}$ on $[-1,1]$. Integrating this
density and substituting $t=s^2$ gives, for $k>-1$,
\[
 \mathbb E|Z_1|^k
 =\frac{\Gamma(d/2)\Gamma((k+1)/2)}
       {\sqrt\pi\Gamma((d+k)/2)}.
\]
In particular,
\begin{align*}
 \mathbb E|X_1|
 &=\frac{\sqrt d\Gamma(d/2)}{\sqrt\pi\Gamma((d+1)/2)},\qquad \mathbb E|X_1|^3
 =\frac{d^{3/2}\Gamma(d/2)}{\sqrt\pi\Gamma((d+3)/2)}
 =\frac{2d}{d+1}\mathbb E|X_1|.
\end{align*}
The last equality uses
$\Gamma((d+3)/2)=((d+1)/2)\Gamma((d+1)/2)$.
Reflection symmetry makes all coordinates of
$\mathbb E[X\mathbf 1\{X_1>0\}]$ except the first equal to zero.
Its first coordinate is $\mathbb E|X_1|/2=\beta_d$, proving the
first claimed identity.

For the quadratic identity, put
\[
 a=\mathbb E[X_1X_2^2\mathbf 1\{X_1>0\}],\qquad
 b=\mathbb E[X_1^3\mathbf 1\{X_1>0\}].
\]
Conditional spherical symmetry implies
$\mathbb E[X_2^2\mid X_1]=(d-X_1^2)/(d-1)$. Therefore
\begin{align*}
 a&=\frac{d\mathbb E|X_1|-\mathbb E|X_1|^3}{2(d-1)} =\frac{d-2d/(d+1)}{2(d-1)}\mathbb E|X_1|
   =\frac{d}{2(d+1)}\mathbb E|X_1|=\alpha_d,\\
 b&=\tfrac12\mathbb E|X_1|^3=2\alpha_d.
\end{align*}
For the first coordinate of the desired vector, reflection in any
coordinate $2,\ldots,d$ eliminates mixed odd terms. The remaining
diagonal terms give
\[
 C_{11}b+\sum_{k>1}C_{kk}a
 =C_{11}(b-a)=\alpha_dC_{11},
\]
where $\operatorname{tr}(C)=0$ was used. For coordinate $k>1$,
the only surviving terms are $C_{1k}$ and $C_{k1}$, giving
$2\alpha_dC_{k1}$. Restoring the original coordinates proves
\[
 \mathbb E[(X^\top CX)X\mathbf 1\{w^\top X>0\}]
 =\alpha_d\{2Cw-(w^\top Cw)w\}.
\]
Taking its inner product with $w$ proves the ReLU identity. Positive
homogeneity then proves the formula for an arbitrary nonzero $w$.

For the final identities let $w$ be uniform on the unit sphere.
The equality $\mathbb E[ww^\top]=I/d$ gives
\[
 \mathbb E\|Cw\|^2=\operatorname{tr}(C^2\mathbb E[ww^\top])
 =\|C\|_F^2/d.
\]
Apply the symmetric-matrix fourth-moment identity in Step~2 of the
proof of Lemma~\ref{lem:qmdp-sphere-moments} to $X=\sqrt d\,w$
and to its two matrices both equal to $C$. Since $w^\top Cw=X^\top CX/d$,
\begin{align*}
 \mathbb E(w^\top Cw)^2
 &=\frac1{d^2}\mathbb E(X^\top CX)^2 =\frac1{d^2}\frac{d}{d+2}
       \{(\operatorname{tr}C)^2+2\operatorname{tr}(C^2)\}\\
 &=\frac{2\|C\|_F^2}{d(d+2)}.
\end{align*}
The last equality uses $\operatorname{tr}C=0$ and symmetry of $C$.
\end{proofof}

\begin{proofof}{Lemma~\ref{lem:dqn-feature-hessian}}
Differentiate the quotient $w^\top Mw/\|w\|$ using symmetry of
$M$. A second differentiation gives
\begin{align*}
 \nabla^2f_M(w)
 &=\frac{2M}{\|w\|}
 -\frac{2(Mw)w^\top+2w(Mw)^\top}{\|w\|^3}
 -\frac{(w^\top Mw)I}{\|w\|^3}
 +\frac{3(w^\top Mw)ww^\top}{\|w\|^5}.
\end{align*}
Put $u=w/\|w\|$ and $P=I-uu^\top$. The same matrix equals
$\{2PMP-(u^\top Mu)P\}/\|w\|$. Since $\|P\|_{\mathrm{op}}\le1$
and $|u^\top Mu|\le\|M\|_{\mathrm{op}}$, its operator norm is at
most $3\|M\|_{\mathrm{op}}/\|w\|$.
For unit $w$, write $c=w^\top Mw$ and expand
\begin{align*}
 \|2Mw-cw\|^2
 &=4\|Mw\|^2-4c\,w^\top Mw+c^2\|w\|^2 =4\|Mw\|^2-3c^2.
\end{align*}
Cauchy--Schwarz gives $c^2\le\|Mw\|^2$, proving the lower bound.
Dropping the negative term and using $\|Mw\|\le\|M\|_{\mathrm{op}}$
proves the upper bound. Finally,
$(2Mw-cw)^\top w=2c-c=c$, proving the last identity.
\end{proofof}

\begin{proofof}{Lemma~\ref{lem:dqn-feature-init}}
\paragraph{Step 1: a uniform Gram lower bound.}
Fix a stage and equip $\mathcal S_h^A$ with norm
$\|B\|^2=\sum_a\|B_a\|_F^2$. This space has dimension $Ar$.
For each of the $k=m/2$ independent neuron pairs define the linear map
\[
 L_jB=\sum_aU_{a,j}^{(0)}B_aw_j,
 \qquad Z_j=L_j^*L_j.
\]
For unit $B$, Cauchy--Schwarz implies
\begin{align*}
 0\le\langle B,Z_jB\rangle
 &=\left\|\sum_aU_{a,j}^{(0)}B_aw_j\right\|^2 \le A\sum_a\|B_aw_j\|^2
 \le A\sum_a\|B_a\|_F^2=A.
\end{align*}
The cross terms have zero expectation over the Rademacher heads,
and $\mathbb E[w_jw_j^\top]=I/d$. Thus
\[
 \mathbb E\langle B,Z_jB\rangle
 =\sum_a\mathbb E\|B_aw_j\|^2
 =\frac1d\sum_a\|B_a\|_F^2=\frac1d.
\]
Thus $0\preceq Z_j\preceq AI$ and $\mathbb E Z_j=I/d$ on the
$Ar$-dimensional space. The $k=m/2$ matrices are independent.
Apply the matrix Chernoff lower-tail bound
\citep[Corollary~5.2 and Remark~5.3]{dqn_tropp2012matrix}
with matrix dimension $Ar$, summand bound $A$, minimum expected
eigenvalue $k/d$, and threshold fraction $1/2$. It gives
\[
 \mathbb P\!\left\{
  \lambda_{\min}\!\left(k^{-1}\sum_j Z_j\right)\le\frac1{2d}\right\}
 \le Ar\exp\!\left(-\frac{(1-1/2)^2(k/d)}{2A}\right)
 =Ar\exp\!\left(-\frac{m}{16Ad}\right).
\]
A union bound over the $H$ stages has failure probability at most
\begin{align*}
 HAr\exp\!\left(-\frac{m}{16Ad}\right)
 &\le HAr\exp\!\left(-Ar\log9-\log\frac{8H}{\delta}\right)\\
 &=\frac{\delta}{8}(Ar)9^{-Ar}\le\frac{\delta}{4}.
\end{align*}
Here \eqref{eq:dqn-feature-width} implies
$m/(16Ad)\ge8\{Ar\log9+\log(8H/\delta)\}$, and $Ar\le9^{Ar}$.
On the complementary event, the lower eigenvalue bound gives
$k^{-1}\sum_j\|L_jB\|^2\ge\|B\|^2/(2d)$ for every $B$.
The duplicate member of each pair has the same squared
norm, making the pair average equal to the all-neuron average.
This proves \eqref{eq:dqn-feature-gram}.

\paragraph{Step 2: uniform control of the unknown-offset interaction.}
For unit $z\in\mathbb R^A$ and $B\in\mathcal S_h^A$, define
\[
 Y_j(z,B)=\left(\sum_aU_{a,j}^{(0)}z_a\right)
 w_j^\top\left(\sum_aU_{a,j}^{(0)}B_a\right)w_j.
\]
The two factors have absolute value at most $\sqrt A$:
the first follows from Cauchy--Schwarz, and for the second use
\[
 \left|\sum_aU_{a,j}^{(0)}w_j^\top B_aw_j\right|
 \le\sqrt A\left(\sum_a|w_j^\top B_aw_j|^2\right)^{1/2}
 \le\sqrt A\|B\|=\sqrt A.
\]
Hence $|Y_j|\le A$. Conditional on the heads, the second factor
has zero mean over $w_j$, since every $B_a$ is traceless. Consequently
$\mathbb EY_j=0$. Hoeffding's inequality for $k$ independent pairs
gives
\[
 \mathbb P\left(\left|k^{-1}\sum_jY_j(z,B)\right|>t\right)
 \le2\exp\left(-\frac{kt^2}{2A^2}\right).
\]
Apply the packing argument in Step~2 of the proof of
Lemma~\ref{lem:qmdp-two-layer-approximation} with $R=1$,
$\rho=1/4$, and ambient dimension $k=A$ or $k=Ar$.
The bound $(1+2R/\rho)^k=9^k$ provides $1/4$-nets of the two
unit spheres, of cardinalities at most $9^A$ and $9^{Ar}$. Setting
\[
 t=2A\sqrt{\frac{A(r+1)\log9+\log(8H/\delta)}m}
\]
makes the union failure probability over both nets and the stages
at most $\delta/4$. The average $Y(z,B)$ is bilinear in $z,B$.
Approximating its two maximizing unit vectors by their net points
changes the value by at most one quarter plus one quarter of its
bilinear operator norm. Moving these terms to the left multiplies
the net bound by two. Thus the full bilinear norm is at most
$4A\sqrt{[A(r+1)\log9+\log(8H/\delta)]/m}$. Homogeneity proves the bound for arbitrary $z,B$.
Both factors change sign at the paired duplicate, so their product
is unchanged. The pair average again equals the all-neuron average.
\end{proofof}

\begin{proofof}{Lemma~\ref{lem:dqn-feature-replay}}
Suppress the stage index on the initial weights and let $s\in[m]$.
For the uniform action probabilities $\bar\pi_a=1/A$, define
\begin{align*}
\widehat G_{I_h,s}(v)
&=\frac1{n\sqrt m}\sum_{i\in I_h}
Y_i(v)U_{A_i,s}^{(0)}
\mathbf1\{w_s^\top X_i>0\}X_i,\\
M_{I_h,s}(v)
&=\sum_a\bar\pi_aU_{a,s}^{(0)}C_a(v),\qquad
\chi_{I_h,s}(v)=\sum_a\bar\pi_aU_{a,s}^{(0)}c_a(v),\\
G_{I_h,s}(v)
&=\frac1{\sqrt m}\left[
\alpha_d\{2M_{I_h,s}(v)w_s
-(w_s^\top M_{I_h,s}(v)w_s)w_s\}
+\beta_d\chi_{I_h,s}(v)w_s
\right].
\end{align*}
For a global replay minibatch with $b_h>0$ stage-$h$ records,
let $J_1,\ldots,J_{b_h}$ be their indices in sampling order,
with multiplicities retained, and define
\[
 \widehat G_s^{\mathrm{mb}}
 :=\frac1{b_h\sqrt m}\sum_{q=1}^{b_h}
 Y_{J_q}(\widehat v)U_{A_{J_q},s}^{(0)}
 \mathbf1\{w_s^\top X_{J_q}>0\}X_{J_q}.
\]

\paragraph{Step 1: conditional concentration for collection.}
Fix a stage--update pair $(h,j)$ and condition on
$\mathscr H_{j-1}$. On a prefix satisfying the lemma's
pre-collection conditions, the initial parameters and selected
continuation $\widehat v$ are fixed, while the $n$ stage-$h$
records come from independent episodes. Their state--action
law is $\mu\otimes\operatorname{Unif}(\mathcal A)$ by
Lemma~\ref{lem:qmdp-state-marginals}.
For each hidden row $s$, put
\[
 Z_{i,s}:=\frac1{\sqrt m}Y_i(\widehat v)U_{A_i,s}^{(0)}
              \mathbf1\{w_s^\top X_i>0\}X_i.
\]
The conditional target model and
Lemma~\ref{lem:dqn-feature-moments} give
\[
 \mathbb E[Z_{i,s}\mid\mathscr H_{j-1}]
 =G_{I_h,s}(\widehat v),\qquad i\in I_h.
\]
Moreover, $\|Z_{i,s}\|_2\le B_Y\sqrt{d/m}$.
For a fixed unit vector $e$, the centered projections are
conditionally independent, have mean zero, and have absolute
value at most $2B_Y\sqrt{d/m}$. Conditional Hoeffding gives the
conservative bound
\[
 \mathbb P\!\left(
 \left|\left\langle e,
 \widehat G_{I_h,s}(\widehat v)-G_{I_h,s}(\widehat v)
 \right\rangle\right|>t
 \;\middle|\;\mathscr H_{j-1}\right)
 \le2\exp\!\left(-\frac{nmt^2}{8B_Y^2d}\right).
\]
Choose a $1/2$-net $\mathcal D$ of the unit sphere with
$|\mathcal D|\le5^d$. Since
$\|z\|_2\le2\max_{e\in\mathcal D}|e^\top z|$, a union bound
over the $m$ rows and $5^d$ directions, with
\[
 t=\sqrt{\frac{8B_Y^2d\ell_{\mathrm{init}}}{mn}},
\]
gives the row bound in \eqref{eq:dqn-feature-replay-error}
with conditional failure probability at most $\delta/(4HK)$.
Squaring the row bounds and summing gives the Frobenius bound.
The prefix-eligibility conditions are
$\mathscr H_{j-1}$-measurable. Taking expectations therefore
bounds the probability of an eligible pair violating the
estimate by $\delta/(4HK)$, even when the assumptions fail on
other prefixes. A union bound over the $HK$ pairs proves the
collection statement. No union over arbitrary windows or over
continuation functions is used.
The same conditional calculation also allows a target bound
that is measurable before the current batch; it is held fixed
during that calculation and retained in the resulting radius.
The theorem above uses the deterministic bound $B_Y$.

\paragraph{Step 2: conditional concentration for replay.}
At the same pair, condition on the complete pre-draw history,
including $\mathscr H_{j-1}$ and the realized buffer $\mathsf B_j$.
All vectors $Z_{i,s}$ are now fixed. Conditional also on the
realized stage count $b_h=k>0$, the stage-$h$ indices are
independent uniform draws from $I_h$.
Indeed, conditional on the positions with stage label $h$,
the product uniform law of the original draws restricts to a
product of uniform laws on $I_h$. This product law does not
depend on the positions, so it remains unchanged after
conditioning only on their count. Hence
\[
 \mathbb E[\widehat G_s^{\mathrm{mb}}
 \mid\text{pre-draw history},b_h=k]
 =\widehat G_{I_h,s}(\widehat v).
\]
Apply the preceding bounded-projection and direction-net
argument with $n$ replaced by $k$. It gives
\eqref{eq:dqn-feature-minibatch-radius} with conditional
failure probability at most $\delta/(4HK)$ for each $k>0$.
Averaging over the count and the histories, and then taking
a union over the $HK$ pairs, gives additional failure
probability at most $\delta/4$.

\paragraph{Step 3: the effective stage step size.}
Because stage parameters are separate and targets are detached,
\begin{align*}
 &\nabla_{\theta_h}\left[
 \frac1{2b}\sum_{i\in\mathcal I_j}
 (Q_{\theta_{h_i}}(X_i,A_i)-Y_i)^2\right]\\
 &\qquad=\frac{b_h}{b}\nabla_{\theta_h}\left[
 \frac1{2b_h}\sum_{i\in\mathcal I_j:\,h_i=h}
 (Q_{\theta_h}(X_i,A_i)-Y_i)^2\right].
\end{align*}
Multiplying by $\eta$ yields $\eta_{\mathrm{eff}}=\eta b_h/b$.
When $b_h=0$, the stage receives no update.
The stage gradient of the global minibatch loss is $b_h/b$
times its stage-normalized gradient. In particular, the actual
hidden update, when the selected continuation is the algorithm's
copied continuation, uses step size $\eta_{\mathrm{eff}}=\eta b_h/b$.

\end{proofof}

\begin{proofof}{Lemma~\ref{lem:dqn-uniform-replay}}
\paragraph{Step 1: the conditional excess-loss identity.}
First fix $h$, a deterministic window $I$, and deterministic functions
$f,v$ in the indicated classes. Let $\mathscr F_{i-1}$ contain the
completed history before episode $i$, including its selected policy
and all past replay sampling. The current transition is still random
conditional on the pre-transition history inside episode $i$.
Set
\[
 e_i=f(X_{i,h},A_{i,h})-\mathcal T_hv(X_{i,h},A_{i,h}),\qquad
 \xi_i=Y_i(v)-\mathcal T_hv(X_{i,h},A_{i,h}).
\]
The Markov conditional-mean equation for this fixed $v$ gives
\[
 \E[\xi_i\mid X_{i,h},A_{i,h},\text{history before its transition}]=0.
\]
This identity holds for each fixed deterministic value continuation. We
establish a simultaneous event over the value continuation class below and
then substitute fitted targets.
Expanding both squares and cancelling $\xi_i^2$ gives
\begin{align*}
 Z_i
 &:=(f(X_{i,h},A_{i,h})-Y_i(v))^2
       -(\mathcal T_hv(X_{i,h},A_{i,h})-Y_i(v))^2\\
 &=(e_i-\xi_i)^2-(-\xi_i)^2
 =e_i^2-2e_i\xi_i+\xi_i^2-\xi_i^2
 =e_i^2-2e_i\xi_i.
\end{align*}
Since $e_i$ is measurable before the transition, iterated conditional
expectation eliminates the cross term. Consequently
\[
 \E[Z_i\mid\mathscr F_{i-1}]
 =\E[e_i^2\mid\mathscr F_{i-1}]
 =:\rho_i
 =\int(f-\mathcal T_hv)^2\,d\mu_{i,h}.
\]

\paragraph{Step 2: boundedness and conditional variance.}
Rewards lie in $[0,1]$ and $v\in[-B,B]$, so both $Y_i(v)$ and $\mathcal T_hv$
lie in $[-B,B+1]$. Since $f\in[-B,B]$,
\[
 |e_i|\le (2B+1),\qquad |\xi_i|\le (2B+1).
\]
It follows that
\begin{align*}
 |Z_i|&\le e_i^2+2|e_i||\xi_i|\\
 &\le (2B+1)|e_i|+2(2B+1)|e_i|=3(2B+1)|e_i|\le3(2B+1)^2,\\
 Z_i^2&\le9(2B+1)^2e_i^2,
 \qquad \Var(Z_i\mid\mathscr F_{i-1})
 \le\E[Z_i^2\mid\mathscr F_{i-1}]\le9(2B+1)^2\rho_i.
\end{align*}
For $U_i=Z_i-\rho_i$, we therefore have
$\E[U_i\mid\mathscr F_{i-1}]=0$, $|U_i|\le4(2B+1)^2$, and
$\E[U_i^2\mid\mathscr F_{i-1}]\le9(2B+1)^2\rho_i$.

\paragraph{Step 3: a two-sided exponential bound.}
Fix $L>0$. For either sign $s\in\{-1,+1\}$, restart the partial sums
of $sU_i$ at zero before $I$ and hold them fixed after $I$.
They form a scalar martingale with increment bound
$R=4(2B+1)^2$ and predictable variance
\[
 V_I:=\sum_{i\in I}\E[U_i^2\mid\mathscr F_{i-1}]
 \le9(2B+1)^2\sum_{i\in I}\rho_i.
\]
The bounded-increment Bernstein line-crossing bound
\citep[Fact~1(c), Proposition~2, and Corollary~4(a)]{howard2020timeuniform}
has scale $R/3$ and gives, for $a,b>0$,
\[
 \mathbb P\left(s\sum_{i\in I}U_i\ge a+bV_I\right)
 \le\exp\left(-\frac{2ab}{1+2Rb/3}\right).
\]
Take $a=36(2B+1)^2L$ and $b=(36(2B+1)^2)^{-1}$. Then
\[
 bV_I\le\frac14\sum_{i\in I}\rho_i,\qquad
 \frac{2ab}{1+2Rb/3}=\frac{2L}{1+2/27}=\frac{54L}{29}\ge L.
\]
Thus, with failure probability at most $e^{-L}$ for each sign,
$s\sum_{i\in I}U_i\le\frac14\sum_{i\in I}\rho_i+36(2B+1)^2L$.
Using the two signs and $\sum Z_i=\sum\rho_i+\sum U_i$ yields
\[
 \frac34\sum_{i\in I}\rho_i-36(2B+1)^2L
 \le\sum_{i\in I}Z_i
 \le\frac54\sum_{i\in I}\rho_i+36(2B+1)^2L.
\]

\paragraph{Step 4: finite nets and all windows.}
There are at most $T^2$ nonempty consecutive windows, $H$ stages,
$N_Q(u)N_V(u)$ pairs of net functions per stage, and two signs.
With $L=\log\frac{2HT^2N_Q(u)N_V(u)}{\delta}$, the union of the failure probabilities is at most
\[
 2HT^2N_Q(u)N_V(u)e^{-\log\frac{2HT^2N_Q(u)N_V(u)}{\delta}}=\delta.
\]
The union bound applies jointly to the dependent, overlapping windows.

\paragraph{Step 5: extend from the nets to every pair of functions.}
Choose net points $f_0,v_0$ within $u$ of $f,v$ respectively. A Markov
kernel contracts the supremum norm, hence
\[
 \|\mathcal T_hv-\mathcal T_hv_0\|_\infty
 =\|P_h(v-v_0)\|_\infty\le u.
\]
For every observation, the two prediction errors differ by at most
$2u$. Their absolute values are at most $(2B+1)$, so
\begin{align*}
 |(f-\mathcal T_hv)^2-(f_0-\mathcal T_hv_0)^2|
 &\le|(f-\mathcal T_hv)-(f_0-\mathcal T_hv_0)|\,
          (|f-\mathcal T_hv|+|f_0-\mathcal T_hv_0|)\\
 &\le(2u)(2(2B+1))=4(2B+1)u.
\end{align*}
Integration gives $|\|f-\mathcal T_hv\|_{L^2(\bar\mu_{I,h})}^2-\|f_0-\mathcal T_hv_0\|_{L^2(\bar\mu_{I,h})}^2|\le4(2B+1)u$.
Each of the residuals $f-Y(v)$ and $\mathcal T_hv-Y(v)$ changes by at most
$2u$ on replacing $(f,v)$ by $(f_0,v_0)$. Each associated squared
term therefore changes by at most $4(2B+1)u$, and
\[
 |\bigl[\widehat L_I(f;v)-\widehat L_I(\mathcal T_hv;v)\bigr]-\bigl[\widehat L_I(f_0;v_0)-\widehat L_I(\mathcal T_hv_0;v_0)\bigr]|\le8(2B+1)u.
\]
The lower estimate becomes
\begin{align*}
 &\bigl[\widehat L_I(f;v)-\widehat L_I(\mathcal T_hv;v)\bigr]\\
 &\ge\bigl[\widehat L_I(f_0;v_0)-\widehat L_I(\mathcal T_hv_0;v_0)\bigr]-8(2B+1)u\\
 &\ge\tfrac34\|f_0-\mathcal T_hv_0\|_{L^2(\bar\mu_{I,h})}^2-\frac{36(2B+1)^2}{n}\log\frac{2HT^2N_Q(u)N_V(u)}{\delta}-8(2B+1)u\\
 &\ge\tfrac34\|f-\mathcal T_hv\|_{L^2(\bar\mu_{I,h})}^2-\frac{36(2B+1)^2}{n}\log\frac{2HT^2N_Q(u)N_V(u)}{\delta}-11(2B+1)u\\
 &\ge\tfrac34\|f-\mathcal T_hv\|_{L^2(\bar\mu_{I,h})}^2-\left(\frac{36(2B+1)^2}{n}\log\frac{2HT^2N_Q(u)N_V(u)}{\delta}+13(2B+1)u\right).
\end{align*}
Similarly,
\begin{align*}
 &\bigl[\widehat L_I(f;v)-\widehat L_I(\mathcal T_hv;v)\bigr]\\
 &\le\bigl[\widehat L_I(f_0;v_0)-\widehat L_I(\mathcal T_hv_0;v_0)\bigr]+8(2B+1)u\\
 &\le\tfrac54\|f_0-\mathcal T_hv_0\|_{L^2(\bar\mu_{I,h})}^2+\frac{36(2B+1)^2}{n}\log\frac{2HT^2N_Q(u)N_V(u)}{\delta}+8(2B+1)u\\
 &\le\tfrac54\|f-\mathcal T_hv\|_{L^2(\bar\mu_{I,h})}^2+\frac{36(2B+1)^2}{n}\log\frac{2HT^2N_Q(u)N_V(u)}{\delta}+13(2B+1)u\\
 &=\tfrac54\|f-\mathcal T_hv\|_{L^2(\bar\mu_{I,h})}^2+\left(\frac{36(2B+1)^2}{n}\log\frac{2HT^2N_Q(u)N_V(u)}{\delta}+13(2B+1)u\right).
\end{align*}
The event now holds for all pairs. A random fitted pair is a member
of this already established event, which completes the substitution
argument.
\end{proofof}

\begin{proofof}{Lemma~\ref{lem:dqn-risk-certificate}}
The inclusion gives
\addtocounter{equation}{2}
\begin{align*}
 \widehat L_I(f;v)-\inf_{q\in\mathcal Q_h(W)}\widehat L_I(q;v)&\ge0,
 \\
 \inf_{q\in\mathcal Q_h(W)}\|q-\mathcal T_hv\|_{L^2(\bar\mu_{I,h})}^2
 &\le\|f-\mathcal T_hv\|_{L^2(\bar\mu_{I,h})}^2.
\end{align*}

Fix a positive number $\zeta$. By the definition of the population
infimum, choose $q_\zeta\in\mathcal Q_h(W)$ with
$\|q_\zeta-\mathcal T_hv\|_{L^2(\bar\mu_{I,h})}^2\le \inf_{q\in\mathcal Q_h(W)}\|q-\mathcal T_hv\|_{L^2(\bar\mu_{I,h})}^2+\zeta$. This comparison function may
depend on the same replay data, since the event is uniform.
The empirical excess losses share the same subtracted term
$\widehat L_I(\mathcal T_hv;v)$. Therefore
\begin{align*}
 &\widehat L_I(f;v)-\widehat L_I(\mathcal T_hv;v)\\
 &=\Bigl[\widehat L_I(f;v)-\inf_{q\in\mathcal Q_h(W)}\widehat L_I(q;v)\Bigr]
 +\inf_{q\in\mathcal Q_h(W)}\widehat L_I(q;v)-\widehat L_I(\mathcal T_hv;v)\\
 &\le\Bigl[\widehat L_I(f;v)-\inf_{q\in\mathcal Q_h(W)}\widehat L_I(q;v)\Bigr]
 +\widehat L_I(q_\zeta;v)-\widehat L_I(\mathcal T_hv;v)\\
 &=\Bigl[\widehat L_I(f;v)-\inf_{q\in\mathcal Q_h(W)}\widehat L_I(q;v)\Bigr]
 +\Bigl[\widehat L_I(q_\zeta;v)-\widehat L_I(\mathcal T_hv;v)\Bigr].
\end{align*}
Apply the lower side of the replay lemma to $f$ and its upper side
to $q_\zeta$:
\begin{align*}
 &\tfrac34\|f-\mathcal T_hv\|_{L^2(\bar\mu_{I,h})}^2
   -\frac{36(2B+1)^2}{n}\log\frac{2HT^2N_Q(u)N_V(u)}\delta-13(2B+1)u\\
 &\le\widehat L_I(f;v)-\widehat L_I(\mathcal T_hv;v)\\
 &\le\Bigl[\widehat L_I(f;v)-\inf_{q\in\mathcal Q_h(W)}\widehat L_I(q;v)\Bigr]
 +\Bigl[\widehat L_I(q_\zeta;v)-\widehat L_I(\mathcal T_hv;v)\Bigr]\\
 &\le\Bigl[\widehat L_I(f;v)-\inf_{q\in\mathcal Q_h(W)}\widehat L_I(q;v)\Bigr]
       +\tfrac54\|q_\zeta-\mathcal T_hv\|_{L^2(\bar\mu_{I,h})}^2\\
 &\quad+\frac{36(2B+1)^2}{n}\log\frac{2HT^2N_Q(u)N_V(u)}\delta+13(2B+1)u\\
 &\le\Bigl[\widehat L_I(f;v)-\inf_{q\in\mathcal Q_h(W)}\widehat L_I(q;v)\Bigr]\\
 &\quad+\tfrac54\Bigl(\inf_{q\in\mathcal Q_h(W)}\|q-\mathcal T_hv\|_{L^2(\bar\mu_{I,h})}^2+\zeta\Bigr)\\
 &\quad+\frac{36(2B+1)^2}{n}\log\frac{2HT^2N_Q(u)N_V(u)}\delta+13(2B+1)u.
\end{align*}
Add the subtracted statistical remainder to both sides and multiply
by $4/3$ to obtain
\begin{align*}
 \|f-\mathcal T_hv\|_{L^2(\bar\mu_{I,h})}^2
 &\le\tfrac43\Bigl[\widehat L_I(f;v)-\inf_{q\in\mathcal Q_h(W)}\widehat L_I(q;v)\Bigr]\\
 &\quad+\tfrac53\inf_{q\in\mathcal Q_h(W)}\|q-\mathcal T_hv\|_{L^2(\bar\mu_{I,h})}^2+\tfrac53\zeta\\
 &\quad+\tfrac83\Bigl[\frac{36(2B+1)^2}{n}\log\frac{2HT^2N_Q(u)N_V(u)}\delta
                         +13(2B+1)u\Bigr].
\end{align*}
Letting $\zeta\downarrow0$ proves the assertion for the actual SGD
output $f$.

\end{proofof}

\begin{proofof}{Lemma~\ref{lem:dqn-network-cover}}
Place a grid of coordinate spacing $u/L_\theta$ in the parameter
box. Including the two boundary endpoints if needed uses at most
$2+2R_\theta L_\theta/u$ values in each coordinate. A nearest grid
point is within $u/L_\theta$ in the infinity norm, hence the
realized critic is within $u$ uniformly over its input domain.
There are at most
$(2+2R_\theta L_\theta/u)^{P_\theta}$ grid points. Taking logarithms gives
the first inequality. For the value functions,
\begin{align*}
 |\max_aQ_\theta(x,a)-\max_aQ_{\theta'}(x,a)|
 &\le\max_a|Q_\theta(x,a)-Q_{\theta'}(x,a)|.
\end{align*}
Thus taking action maxima of a critic net gives a value net of no
larger cardinality.

For the explicit two-layer bounds, $|w_j^\top x|\le dR_\theta B_x$
and $|U_{a,j}|\le R_\theta$. Summing the $m$ terms gives
\[
 |Q_\theta(x,a)|\le R_\theta
       +m^{-1/2}mR_\theta(dR_\theta B_x)=B_Q.
\]
Put $s=\|\theta-\theta'\|_\infty$. ReLU is one-Lipschitz, so
\begin{align*}
 |U_{a,j}\relu(w_j^\top x)
       -U'_{aj}\relu((w'_j)^\top x)|
 &\le|U_{a,j}-U'_{aj}|\,|\relu(w_j^\top x)|\\
 &\quad+|U'_{aj}|\,|\relu(w_j^\top x)-\relu((w'_j)^\top x)|\\
 &\le s(dR_\theta B_x)
      +R_\theta\sum_{k=1}^d|w_{jk}-w'_{jk}|\,|x_k|\\
 &\le dR_\theta B_xs+dR_\theta B_xs
 =2dR_\theta B_xs.
\end{align*}
Add the bias difference, sum over neurons, and divide by $\sqrt m$:
\[
 |Q_\theta-Q_{\theta'}|
 \le s+2\sqrt m\,dR_\theta B_xs=L_\theta s.
\]
All inequalities hold for every input and action, completing the proof.
\end{proofof}

\begin{proofof}{Lemma~\ref{lem:dqn-parameter-envelope}}
Between updates,
target copies inherit the current online parameter envelope.
The claim holds at initialization by the definition of $R_0$.
Suppose it holds before update $k$. Lemma~\ref{lem:dqn-network-cover}
then bounds both current and target predictions by $B(R_k)$.
A reward lies in $[0,1]$, so the TD target has absolute value at most
$1+B(R_k)$. For each sampled transition its residual satisfies
\[
 |Q_\theta-Y|\le |Q_\theta|+|Y|
                 \le 1+2B(R_k).
\]
For a selected action $a$, the absolute derivatives of the critic
with respect to its three kinds of parameters are bounded by
\begin{align*}
 |\partial_{b_a}Q_\theta|&\le1,
 \qquad |\partial_{U_{a,j}}Q_\theta|
 =m^{-1/2}|\relu(w_j^\top x)|
 \le dB_xR_k/\sqrt m,\\
 |\partial_{w_{j\ell}}Q_\theta|
 &=m^{-1/2}|U_{a,j}|\,\one\{w_j^\top x>0\}\,|x_\ell|
 \le B_xR_k/\sqrt m\le dB_xR_k/\sqrt m.
\end{align*}
The same bounds apply with the prescribed zero derivative at a ReLU
knot. Unselected action derivatives are zero. Multiply these bounds
by the residual bound. Every scalar coordinate of a single-sample
gradient of the half squared loss has magnitude at most $G(R_k)$.
A uniform minibatch average has the same bound by the triangle
inequality, irrespective of repeated indices. Therefore
\[
 |\theta_{k+1,\ell}|
 \le|\theta_{k,\ell}|+\eta_k|\widehat g_{k,\ell}|
 \le R_k+\eta_kG(R_k)=R_{k+1}.
\]
The sequence $R_k$ is nondecreasing. A target copy replaces its
parameters by previously bounded online parameters and so obeys
the same envelope. Substituting this envelope into the covering
bound expresses its statistical constants in terms of
$R_{K_{\mathrm{upd}}}$.
\end{proofof}

\begin{proofof}{Lemma~\ref{lem:dqn-quadratic-eigenvalues}}
Let $w$ be uniform on the unit sphere, let
$\xi=(\xi_a)_{a\in\mathcal A}$ have independent Rademacher coordinates,
and let $X'\sim\mu$ be independent of $(w,\xi)$. Fix functions
$f_a(x)=c_a+x^\top C_ax$, with $C_a$ symmetric and trace-free,
independently of $(w,\xi)$. Define the three population operators
\begin{align*}
 (\mathcal K_Wf)(x,a)
 &=\mathbb E_{w,\xi}\!\left[
   \xi_a\mathbf1\{w^\top x>0\}x^\top
   \frac1A\sum_{a'\in\mathcal A}\xi_{a'}
       \mathbb E_{X'}[f_{a'}(X')X'\mathbf1\{w^\top X'>0\}]
                         \right],\\
 (\mathcal K_Uf)(x,a)
 &=\frac1A\mathbb E_w\!\left[
       \operatorname{ReLU}(w^\top x)
       \mathbb E_{X'}[f_a(X')\operatorname{ReLU}(w^\top X')]
                         \right],\\
 (\mathcal K_bf)(x,a)&=\frac1A\mathbb E_{X'}f_a(X').
\end{align*}

\paragraph{Step 1: spherical identities with the roles of input and weight interchanged.}
Fix $x$ with $\|x\|_2=\sqrt d$, put $u=x/\sqrt d$, and set
$Z=\sqrt d\,w$. Then $\|u\|_2=1$, $Z\sim\mu$, and
$w^\top x=u^\top Z$. By Lemma~\ref{lem:dqn-feature-moments},
\begin{align*}
 \mathbb E_w[w\mathbf1\{w^\top x>0\}]
 &=\frac1{\sqrt d}
          \mathbb E_Z[Z\mathbf1\{u^\top Z>0\}]
 =\frac{\beta_d}{\sqrt d}u
 =\frac{\beta_d}{d}x,\\
 \mathbb E_w\operatorname{ReLU}(w^\top x)
 &=u^\top\mathbb E_Z[Z\mathbf1\{u^\top Z>0\}]
 =\beta_du^\top u=\beta_d.
\end{align*}
For a symmetric trace-free $C$, the same moment lemma gives
\begin{align*}
 \mathbb E_w[(w^\top Cw)\operatorname{ReLU}(w^\top x)]
 &=\frac1d\mathbb E_Z[(Z^\top CZ)\operatorname{ReLU}(u^\top Z)]\\
 &=\frac{\alpha_d}{d}u^\top Cu
 =\frac{\alpha_d}{d^2}x^\top Cx.
\end{align*}
The factors $1/\sqrt d$ and $1/d$ above come from replacing $w$
by $Z/\sqrt d$.

\paragraph{Step 2: the hidden-weight operator.}
For fixed unit $w$, Lemma~\ref{lem:dqn-feature-moments} gives
\[
 \mathbb E_{X'}[f_{a'}(X')X'\mathbf1\{w^\top X'>0\}]
 =\beta_dc_{a'}w
       +\alpha_d\{2C_{a'}w-(w^\top C_{a'}w)w\}.
\]
Independence and centering of the Rademacher coordinates imply
$\mathbb E[\xi_a\xi_{a'}]=\mathbf1\{a=a'\}$. Taking their
expectation in the definition of $\mathcal K_W$ therefore yields
\begin{align*}
 A(\mathcal K_Wf)(x,a)
 &=\mathbb E_w\!\left[
  \mathbf1\{w^\top x>0\}x^\top
    \{\beta_dc_aw+\alpha_d(2C_aw-(w^\top C_aw)w)\}\right]\\
 &=\beta_dc_a\mathbb E_w\operatorname{ReLU}(w^\top x)
     +2\alpha_dx^\top C_a
              \mathbb E_w[w\mathbf1\{w^\top x>0\}]\\
 &\qquad-\alpha_d
       \mathbb E_w[(w^\top C_aw)\operatorname{ReLU}(w^\top x)]\\
 &=\beta_d^2c_a+
       \left(\frac{2\alpha_d\beta_d}{d}
                         -\frac{\alpha_d^2}{d^2}\right)x^\top C_ax.
\end{align*}
The second equality uses
$(w^\top x)\mathbf1\{w^\top x>0\}=\operatorname{ReLU}(w^\top x)$;
the third substitutes the identities from Step 1. Finally,
$\beta_d=(d+1)\alpha_d/d$ gives
\[
 \frac{2\alpha_d\beta_d}{d}-\frac{\alpha_d^2}{d^2}
 =\frac{\alpha_d^2[2(d+1)-1]}{d^2}
 =\frac{\alpha_d^2(2d+1)}{d^2}.
\]
This proves \eqref{eq:dqn-eigen-hidden}.

\paragraph{Step 3: the action-head and bias operators.}
For fixed unit $w$, the scalar spherical moments give
\[
 \mathbb E_{X'}[f_a(X')\operatorname{ReLU}(w^\top X')]
 =\beta_dc_a+\alpha_dw^\top C_aw.
\]
Substitute this identity into the action-head operator and apply Step 1:
\begin{align*}
 A(\mathcal K_Uf)(x,a)
 &=\mathbb E_w[\operatorname{ReLU}(w^\top x)
                       (\beta_dc_a+\alpha_dw^\top C_aw)]\\
 &=\beta_dc_a\mathbb E_w\operatorname{ReLU}(w^\top x)
       +\alpha_d\mathbb E_w[(w^\top C_aw)
                                  \operatorname{ReLU}(w^\top x)]\\
 &=\beta_d^2c_a+\frac{\alpha_d^2}{d^2}x^\top C_ax
 .
\end{align*}
For the bias operator, $\mathbb E[X'X'^\top]=I_d$ and
$\operatorname{tr}(C_a)=0$ imply
\[
 A(\mathcal K_bf)(x,a)
 =c_a+\operatorname{tr}(C_a\mathbb E[X'X'^\top])
 =c_a+\operatorname{tr}(C_a)=c_a.
\]
Adding the three rescaled identities gives the coefficients
$1+2\beta_d^2$ and $\alpha_d^2(2d+1)/d^2+\alpha_d^2/d^2=2\alpha_d^2(d+1)/d^2$ for
the operator $A(\mathcal K_W+\mathcal K_U+\mathcal K_b)$.
The operator $A(\mathcal K_U+\mathcal K_b)$ has coefficients
$1+\beta_d^2$ and $\alpha_d^2/d^2$. Dividing each coefficient
by $A$ gives the full and frozen-hidden update operators in the lemma.

Under paired initialization, the duplicated neuron has the same $w$
and the sign vector $-\xi$. The products $\xi_a\xi_{a'}$ in the
hidden operator are unchanged by this duplication. The head and bias
operators are unchanged as well. Thus their population values are
the operators computed above, with the independent neuron pairs
providing the samples in the finite-width concentration argument.
Thus the summed full-update operator has eigenvalues
$\left(1+2\beta_d^2\right)/A$ on the actionwise constants and ${\frac{2\alpha_d^2(d+1)}{d^2}}/A$ on the
trace-free quadratics. The head-and-bias operator has the corresponding
eigenvalues $(1+\beta_d^2)/A$ and ${\frac{\alpha_d^2}{d^2}}/A$.

\end{proofof}

\begin{proofof}{Lemma~\ref{lem:dqn-uniform-step}}
We work conditionally on $\mathscr G$ throughout. The target mean
$f$ is then fixed and the selected initialization has its original
paired distribution.

\paragraph{Step 1: the three gradient batches.}
Let $\mathcal I_h=(J_1,\ldots,J_{b_h})$ be the ordered stage
replay draws from $\{1,\ldots,n\}$, sampled independently and
uniformly with replacement conditional on the buffer. Sums over
$\mathcal I_h$ count repeated draws with their multiplicities.
The zero initial prediction gives the following negative gradients
of the stage-normalized half-squared loss:
\begin{align*}
 \widehat G_{W,j}
 &=\frac1{b_h\sqrt m}\sum_{i\in\mathcal I_h}
       U_{A_i,j}^{(0)}Y_iX_i\mathbf1\{(w_j^{(0)})^\top X_i>0\},\\
 \widehat G_{U,aj}
 &=\frac1{b_h\sqrt m}\sum_{i\in\mathcal I_h}
       \mathbf1\{A_i=a\}Y_i\sigma((w_j^{(0)})^\top X_i),\\
 \widehat G_{b,a}
 &=\frac1{b_h}\sum_{i\in\mathcal I_h}\mathbf1\{A_i=a\}Y_i.
\end{align*}
The update adds $\eta_{\mathrm{eff}}\widehat G$ to each batch.
Since $|Y_i|\le1$, $|U_{a,j}^{(0)}|=1$, $\|X_i\|_2=\sqrt d$,
and $\|w_j^{(0)}\|_2=1$, the triangle inequality gives
\begin{equation}\label{eq:dqn-step-deterministic-moves}
 \|\widehat G_{W,j}\|_2\le\frac{\sqrt d}{\sqrt m},\qquad
 |\widehat G_{U,aj}|\le\frac{\sqrt d}{\sqrt m},\qquad
 |\widehat G_{b,a}|\le1.
\end{equation}

\paragraph{Step 2: collection and replay concentration.}
Let $G$ denote the conditional population directions, obtained by
replacing the empirical averages by expectation under the stated
conditional law. For a fixed hidden row $j$, put
\[
 Z_i:=U_{A_i,j}^{(0)}Y_iX_i
       \mathbf1\{(w_j^{(0)})^\top X_i>0\}.
\]
For a unit vector $v$, $v^\top Z_i\in[-\sqrt d,\sqrt d]$.
Conditional Hoeffding's inequality uses this interval's length
$2\sqrt d$ and gives
\[
 \mathbb P\!\left(
  \left|\frac1n\sum_{i=1}^n
     v^\top(Z_i-\mathbb E[Z_i\mid\mathcal F_{i-1}])\right|>r
       \right)
 \le2\exp\!\left(-\frac{nr^2}{2d}\right).
\]
A $1/4$-net of the unit sphere has at most $9^d$ points.
The identity $v^\top u=1-\|v-u\|^2/2$ for unit vectors gives
$\|z\|_2\le(32/31)\max_v|v^\top z|$ on this net. Substituting
$r=\sqrt{2d\log\!\frac{64A m5^d(1+4m\sqrt d)^d}{\delta}/n}$
and using $32\sqrt2/31<3/2$ bounds the hidden vector error by
$(3/2)\sqrt{d\log\!\frac{64A m5^d(1+4m\sqrt d)^d}{\delta}/n}$.
The head summands have absolute value at most $\sqrt d$, and the
bias summands at most one. Applying the same scalar inequality and
then conditioning on the realized buffer for the fresh replay draws
gives, by a union bound over all neurons and actions,
\begin{align*}
 \max_j\|\widehat G_{W,j}-G_{W,j}\|_2
 &\le\frac{3\sqrt d}{2\sqrt m}{\sqrt{\log\!\frac{64A m5^d(1+4m\sqrt d)^d}{\delta}}\left(\frac1{\sqrt n}+\frac1{\sqrt{b_h}}\right)},\\
 \max_{a,j}|\widehat G_{U,aj}-G_{U,aj}|
 &\le\frac{3\sqrt d}{2\sqrt m}{\sqrt{\log\!\frac{64A m5^d(1+4m\sqrt d)^d}{\delta}}\left(\frac1{\sqrt n}+\frac1{\sqrt{b_h}}\right)},\\
 \max_a|\widehat G_{b,a}-G_{b,a}|
 &\le\frac32\,{\sqrt{\log\!\frac{64A m5^d(1+4m\sqrt d)^d}{\delta}}\left(\frac1{\sqrt n}+\frac1{\sqrt{b_h}}\right)}.
\end{align*}
The collection and replay terms arise from two different conditional
averages: environment records around their population mean, and
replay draws around the empirical buffer mean. The triangle
inequality adds their errors.

For the hidden linearized prediction, Cauchy--Schwarz now gives
\begin{align*}
 &\left|\frac1{\sqrt m}\sum_jU_{a,j}^{(0)}
   \mathbf1\{(w_j^{(0)})^\top x>0\}
   x^\top(\widehat G_{W,j}-G_{W,j})\right|\\
 &\qquad\le\|x\|_2
       \left(\sum_j\|\widehat G_{W,j}-G_{W,j}\|_2^2\right)^{1/2}
\\
 &\qquad\le\frac32\,d\,{\sqrt{\log\!\frac{64A m5^d(1+4m\sqrt d)^d}{\delta}}\left(\frac1{\sqrt n}+\frac1{\sqrt{b_h}}\right)}.
\end{align*}
Also $\|\varphi_{W_0}(x)\|_2\le\sqrt d$, so the head error is at most
\[
 \frac32\,d\sqrt{\log\!\frac{64Am5^d(1+4m\sqrt d)^d}{\delta}}
           \left(\frac1{\sqrt n}+\frac1{\sqrt{b_h}}\right).
\]
Adding the bias error bounds the full linearized prediction error by
\[
 \frac32(2d+1)\sqrt{\log\!\frac{64Am5^d(1+4m\sqrt d)^d}{\delta}}
           \left(\frac1{\sqrt n}+\frac1{\sqrt{b_h}}\right).
\]
These inequalities hold for every $x$.

\paragraph{Step 3: a uniform bound on small initial preactivations.}
For fixed $x\in\mathbb S_d$, the random variable $w^\top x$ has
density at most one on $[-1/2,1/2]$. For $d=2$, its density is
$1/(\pi\sqrt{2-z^2})$ there. For $d\ge3$, the density equals
\[
 \frac{\Gamma(d/2)}{\sqrt{\pi d}\,\Gamma((d-1)/2)}
       (1-z^2/d)^{(d-3)/2}.
\]
The power is at most one, and the gamma ratio is bounded by
$\sqrt{(d-1)/2}$: this follows by applying Cauchy--Schwarz to the
integrals defining $\Gamma(x+1/2)$, which gives
$\Gamma(x+1/2)^2\le\Gamma(x)\Gamma(x+1)=x\Gamma(x)^2$.
Thus
\[
 \mathbb P(|w^\top x|\le z)\le2z\qquad(0\le z\le1/2).
\]
Set $t_0=d/\sqrt m$. Choose a $1/m$-net
$\mathcal N_x\subset\mathbb S_d$ of cardinality at most
$(1+2m\sqrt d)^d$. The width condition implies $t_0+1/m\le1/4$.
The $m/2$ independently drawn weights determine all $m$ paired
weights. Hoeffding's inequality, applied to their strip indicators
and unioned over this net and the two thresholds, gives, for every
$y\in\mathcal N_x$ and $z\in\{1/m,t_0+1/m\}$,
\[
 \frac1m\sum_j\mathbf1\{|(w_j^{(0)})^\top y|\le z\}
 \le2z+\sqrt{\frac1m\log\!\frac{64A m5^d(1+4m\sqrt d)^d}{\delta}}.
\]
For any $x$, choose $y$ with $\|x-y\|_2\le 1/m$. Then
$|(w_j^{(0)})^\top x|\le t_0$ implies
$|(w_j^{(0)})^\top y|\le t_0+1/m$. Consequently the fraction of
neurons with $|(w_j^{(0)})^\top x|\le t_0$ is at most
$2(t_0+1/m)+{\sqrt{\frac1m\log\!\frac{64A m5^d(1+4m\sqrt d)^d}{\delta}}}$, uniformly in $x$.

\paragraph{Step 4: finite-width population directions.}
Put $g_j=\sqrt m\,G_{W,j}$. For every unit $v$, central
symmetry gives $|v^\top g_j|\le
\mathbb E[|v^\top X|\mathbf1\{(w_j^{(0)})^\top X>0\}]
=\beta_d\le1/2$, so $\|g_j\|_2\le1/2$.
The population hidden linearized response is the empirical mean of
\[
 Z_{aj}(x)=U_{a,j}^{(0)}\mathbf1\{(w_j^{(0)})^\top x>0\}x^\top g_j,
 \qquad |Z_{aj}(x)|\le\sqrt d/2.
\]
The two members of an initialization pair have the same $Z_{aj}$;
the $m/2$ pair values are independent. Hoeffding's inequality bounds
the difference from their initialization expectation by $\sqrt d\,{\sqrt{\frac1m\log\!\frac{64A m5^d(1+4m\sqrt d)^d}{\delta}}}$
at each point of $\mathcal N_x$, simultaneously over actions.
For neighboring $x,y$, the triangle inequality gives
\[
 |Z_{aj}(x)-Z_{aj}(y)|
 \le\frac1{2m}+
       \frac{\sqrt d}{2}\mathbf1\{|(w_j^{(0)})^\top y|\le 1/m\}.
\]
Applying Step 3 to the empirical strip fraction and the strip
probability to its expectation therefore yields
\begin{align*}
 &\sup_x\left|\frac1m\sum_jZ_{aj}(x)-\mathbb E_{\mathrm{init}}Z_{aj}(x)\right|\\
 &\le\sqrt d\,{\sqrt{\frac1m\log\!\frac{64A m5^d(1+4m\sqrt d)^d}{\delta}}}+\frac1m+\frac{\sqrt d}{2}(4/m+{\sqrt{\frac1m\log\!\frac{64A m5^d(1+4m\sqrt d)^d}{\delta}}})\\
 &=\frac{3\sqrt d}{2}{\sqrt{\frac1m\log\!\frac{64A m5^d(1+4m\sqrt d)^d}{\delta}}}+(1+2\sqrt d)/m\\
 &\le2(d+1){\sqrt{\frac1m\log\!\frac{64A m5^d(1+4m\sqrt d)^d}{\delta}}}.
\end{align*}
The population head response is the empirical mean of
$\sigma((w_j^{(0)})^\top x)\sqrt m\,G_{U,aj}$.
Since $|\sqrt m\,G_{U,aj}|\le\beta_d/A\le1/4$, its absolute
value is at most $\sqrt d/4$ and its Lipschitz constant in $x$
is at most $1/4$. The same net argument, now without
strip indicators, bounds its uniform deviation by
\begin{align*}
 &\frac{\sqrt d}{2}\sqrt{\frac1m\log\!\frac{64Am5^d(1+4m\sqrt d)^d}{\delta}}
       +\frac1{2m}
 \qquad\le(d+1)
       \sqrt{\frac1m\log\!\frac{64Am5^d(1+4m\sqrt d)^d}{\delta}}.
\end{align*}
The population bias direction is exactly $c_a/A$.
The identities of Lemma~\ref{lem:dqn-quadratic-eigenvalues} identify
the initialization expectations as
\[
 \frac{\beta_d^2c_a+{\frac{\alpha_d^2(2d+1)}{d^2}}s_a}{A},\qquad
 \frac{\beta_d^2c_a+{\frac{\alpha_d^2}{d^2}}s_a}{A},\qquad
 \frac{c_a}{A},
\]
for hidden weights, heads, and biases, respectively.

\paragraph{Step 5: the finite ReLU step.}
Let $\Delta w_j=\eta_{\mathrm{eff}}\widehat G_{W,j}$ and
$\Delta U_{a,j}=\eta_{\mathrm{eff}}\widehat G_{U,aj}$.
For all real $z,u$,
\[
 |\sigma(z+u)-\sigma(z)-\mathbf1\{z>0\}u|
 \le |u|\mathbf1\{|z|\le|u|\}.
\]
Indeed the left side is zero when the segment from $z$ to $z+u$
stays in one linear piece, and is at most $|u|$ when it crosses zero.
By \eqref{eq:dqn-step-deterministic-moves},
$|\Delta w_j^\top x|\le\eta_{\mathrm{eff}}d/\sqrt m\le t_0$.
The nonlinear remainder in the initial-head readout is consequently
bounded, using Step 3, by
\begin{align*}
 R_W
 &\le\sqrt m\frac{\eta_{\mathrm{eff}}d}{\sqrt m}
            [2(t_0+1/m)+{\sqrt{\frac1m\log\!\frac{64A m5^d(1+4m\sqrt d)^d}{\delta}}}]\\
 &=\eta_{\mathrm{eff}}d
       \left[\frac{2d}{\sqrt m}+\frac2m+{\sqrt{\frac1m\log\!\frac{64A m5^d(1+4m\sqrt d)^d}{\delta}}}\right]\\
 &\le5\eta_{\mathrm{eff}}(d+1)^2{\sqrt{\frac1m\log\!\frac{64A m5^d(1+4m\sqrt d)^d}{\delta}}}.
\end{align*}
Here $m^{-1}\le m^{-1/2}\le {\sqrt{\frac1m\log\!\frac{64A m5^d(1+4m\sqrt d)^d}{\delta}}}$ was used in the final line.
The simultaneous head update adds one product remainder. ReLU is
one-Lipschitz, so
\begin{align*}
 R_{WU}
 &\le\frac1{\sqrt m}\sum_j|\Delta U_{a,j}|
             |\sigma((w_j^{(0)}+\Delta w_j)^\top x)
                        -\sigma((w_j^{(0)})^\top x)|\\
 &\le\frac1{\sqrt m}\sum_j
       \frac{\eta_{\mathrm{eff}}\sqrt d}{\sqrt m}
       \frac{\eta_{\mathrm{eff}}d}{\sqrt m}
 =\frac{\eta_{\mathrm{eff}}^2d^{3/2}}{\sqrt m}
 \le\eta_{\mathrm{eff}}(d+1)^2{\sqrt{\frac1m\log\!\frac{64A m5^d(1+4m\sqrt d)^d}{\delta}}}.
\end{align*}
The frozen-hidden comparator has neither remainder because its
prediction is affine in the heads and biases.

\paragraph{Step 6: combine the events and the errors.}
All collection and replay inequalities use at most
$4m(A+1)9^d+4A$ scalar tail bounds. The finite-width and strip
arguments use at most $4(A+1)(1+2m\sqrt d)^d$ tail bounds.
Since $A+1\le2A$, $m5^d\ge1$,
$9^d\le5^d(1+4m\sqrt d)^d$, and
$1+2m\sqrt d\le1+4m\sqrt d$, their total number satisfies
\begin{align*}
 &4m(A+1)9^d+4A+4(A+1)(1+2m\sqrt d)^d
 \le8Am9^d+4A+8A(1+2m\sqrt d)^d\\
 &\le(8+4+8)Am5^d(1+4m\sqrt d)^d
 =20Am5^d(1+4m\sqrt d)^d.
\end{align*}
Each tail is at most
$2\exp[-\log(64Am5^d(1+4m\sqrt d)^d/\delta)]$.
Multiplying this tail bound by the count, and cancelling the common
positive factors, gives total failure probability at most
\[
 20Am5^d(1+4m\sqrt d)^d\,
 \frac{2\delta}{64Am5^d(1+4m\sqrt d)^d}
 =\frac58\delta\le\delta.
\]
On their intersection, add the collection, replay, width, and remainder
bounds, retaining the intermediate estimates in Steps 4 and 5.
The common logarithm is at least $16$ by the width condition.
Factoring its square root gives
\begin{align*}
 &\eta_{\mathrm{eff}}
 \sqrt{\log\!\frac{64Am5^d(1+4m\sqrt d)^d}{\delta}}
 \left[\begin{aligned}
 &\frac32(2d+1)\left(\frac1{\sqrt n}+\frac1{\sqrt{b_h}}\right)\\
 &+\frac{(3/4)d^2+2d+9/8}{\sqrt m}
 \end{aligned}\right]\\
 &\quad\le\eta_{\mathrm{eff}}(d+1)^2
 \sqrt{\log\!\frac{64Am5^d(1+4m\sqrt d)^d}{\delta}}
 \left[\left(\frac1{\sqrt n}+\frac1{\sqrt{b_h}}\right)
                                      +\frac1{\sqrt m}\right]\\
 &\quad=\eta_{\mathrm{eff}}(d+1)^2
 \sqrt{\log\!\frac{64Am5^d(1+4m\sqrt d)^d}{\delta}}
 \left(\frac1{\sqrt n}+\frac1{\sqrt{b_h}}+\frac1{\sqrt m}\right).
\end{align*}
The common coefficient follows from the separate inequalities
\begin{align*}
 \frac32(2d+1)&\le(d+1)^2,\\
 2\sqrt d+d+\frac{2d^2+\eta_{\mathrm{eff}}d^{3/2}}4
       +\frac{3/2+2\sqrt d+2d}{32(d+1)}
 &\le\frac34d^2+2d+\frac98\le(d+1)^2.
\end{align*}
The second line bounds the width and remainder terms using the
logarithm's lower bound, $\sqrt m\ge8(d+1)$,
$2\sqrt d\le d+1$, and $d^{3/2}\le d^2$.
Both lines use $d\ge2$.
Using the explicit operator responses of
Lemma~\ref{lem:dqn-quadratic-eigenvalues}, the three bounds read
\stepcounter{equation}
\begin{equation*}
\begin{aligned}
 \max\Bigg\{&\max_a\left\|F_a^+-\frac{\eta_{\mathrm{eff}}}{A}
       \left(\beta_d^2c_a+\frac{\alpha_d^2(2d+1)}{d^2}s_a\right)
                         \right\|_\infty,\\
 &\max_a\left\|Q_{\mathrm{full},a}^+-\frac{\eta_{\mathrm{eff}}}{A}
       \left((1+2\beta_d^2)c_a+\frac{2\alpha_d^2(d+1)}{d^2}s_a\right)
                         \right\|_\infty,\\
 &\max_a\left\|Q_{\mathrm{head},a}^+-\frac{\eta_{\mathrm{eff}}}{A}
       \left((1+\beta_d^2)c_a+\frac{\alpha_d^2}{d^2}s_a\right)
                         \right\|_\infty\Bigg\}\\
 &\le\eta_{\mathrm{eff}}(d+1)^2
 \left(\frac1{\sqrt n}+\frac1{\sqrt{b_h}}+\frac1{\sqrt m}\right)
 \sqrt{\log\!\frac{64Am5^d(1+4m\sqrt d)^d}{\delta}}.
\end{aligned}
\end{equation*}

The order statements in the main text follow from
\begin{equation}\label{eq:dqn-step-order-conversion}
\begin{aligned}
 \log\!\frac{64Am5^d(1+4m\sqrt d)^d}{\delta}
 &\le\log\frac A\delta+6\log2+\log m
                   +d\log5+d\log(5md)\\
 &\le\log\frac A\delta+(3d+1)\log m+d\log d\\
 &\le4\left[d\log(md)+\log\frac A\delta\right],\\
 \left(\frac1{\sqrt n}+\frac1{\sqrt{b_h}}+
                    \frac1{\sqrt m}\right)^2
 &\le3\left(\frac1n+\frac1{b_h}+\frac1m\right).
\end{aligned}
\end{equation}
For the first inequality, $1+4m\sqrt d\le5md$ because $m,d\ge1$.
For the second, $m\ge64(d+1)^2$ implies both
$6\log2\le d\log m$ and $2\log5\le\log m$.
The third follows from $3d+1\le4d$, $d\log d\le4d\log d$,
and $\log(A/\delta)>0$.
The final inequality is Cauchy--Schwarz applied to the three
reciprocal square roots. Finally, $d+1\le3d/2$ yields
the uniform prediction order
\[
 \mathcal O\!\left(\eta_{\mathrm{eff}}d^2
 \sqrt{\left[d\log(md)+\log\frac A\delta\right]
       \left(\frac1n+\frac1{b_h}+\frac1m\right)}\right).
\]
The hidden-only and frozen-hidden predictions omit some of these
nonnegative error terms and satisfy the same bound. Conditioning on
the stage count in a global minibatch leaves its selected stage
indices independent and uniform over the stage records. The
effective-step identity in Lemma~\ref{lem:dqn-feature-replay}
therefore gives the last assertion.
\end{proofof}

\Needspace{8\baselineskip}
\begin{proofof}{Lemma~\ref{lem:dqn-horizon-iterates}}
Put $\lambda_0=1+2\beta_d^2$ and
$\lambda_2=2\alpha_d^2(d+1)/d^2$. Define
\begin{align}
 \ell&=\log\!\left(
  \frac{32H(K+1)m(1+4m\sqrt d)^{2d}}{\delta}\right),
 \label{eq:dqn-long-log}\\
 E_{\mathrm{run}}
  &=\frac{24(d+1)K}{H}
       \left(1+\sqrt{\frac{2(d+1)K}{H}}\right)\sqrt\ell
       \nonumber\\
  &\quad\times\left[\frac1{\sqrt n}+\sqrt{\frac{2H}{b}}
          +\frac{(d+1)(1+K/H)}{\sqrt m}\right].
 \label{eq:dqn-long-error}
\end{align}

For the proof only, abbreviate the sampling and width scales by
\stepcounter{equation}
\begin{equation*}
 \chi_{\mathrm{data}}
 :=\sqrt\ell\left(\frac1{\sqrt n}+\sqrt{\frac{2H}{b}}\right),
 \qquad \rho:=\sqrt{\frac\ell m}.
\end{equation*}
Let $\nu=\mu\otimes\operatorname{Unif}\{-1,1\}$, and let
$\mathcal T_0$ be the deterministic population tangent operator
at initialization, with integration against $\nu$. Its kernel is
the expectation of the initial gradient inner product. The two
action batches are diagonal in expectation. The one-action bias,
head, and hidden-weight operators have $L^2(\mu)$ norms at most
$1$, $1/2$, and $1/2$, respectively. The head bound follows from
$\mathbb E_\mu[\operatorname{ReLU}(w^\top X)^2]=1/2$.
For the hidden part, the map
$A_wf=\mathbb E_\mu[X\mathbf1\{w^\top X>0\}f(X)]$
satisfies $A_wA_w^\ast=I_d/2$. Averaging preserves its squared
operator-norm bound. Uniform actions introduce an additional
factor $1/2$. Thus
\stepcounter{equation}
\begin{equation*}
 0\preceq\mathcal T_0,\qquad
 \|\mathcal T_0\|_{2\to2}\le1,\qquad
 \sup_z K_\infty(z,z)\le K_0:=2(d+1).
\end{equation*}
The same diagonal bound $K_0$ applies to the finite initial kernel.
The constant and trace-free quadratic eigenvalues of
$\mathcal T_0$ are $\lambda_0/2$ and $\lambda_2/2$.

\paragraph{1. Counts and uniform initialization events.}
Let $p_{h,j}$ be the stage-$h$ fraction of global minibatch $j$,
and put $\eta_{h,j}=p_{h,j}/2$. Each retained buffer has exactly
$n$ records at each stage. A binomial tail bound and a union
bound over its $HK$ stage-minibatch pairs give
\[
 \Pr\!\left\{\exists h,j:\,
       p_{h,j}\notin[1/(2H),3/(2H)]\right\}
 \le2HK e^{-b/(12H)}\le\delta/16.
\]
On the complementary event,
\begin{equation}\label{eq:dqn-long-time}
 \frac1{4H}\le\eta_{h,j}\le\frac3{4H},\qquad
 \sum_{j=1}^K\eta_{h,j}\le\frac{3K}{4H}.
\end{equation}
Stage counts are independent of the record values and
initializations. Conditional on the counts, the stagewise replay
indices are independent uniform draws from the $n$ corresponding
records.

Write $\Phi_{h,0}(z)$ for the full initial parameter gradient and
$K_{h,0}(z,z')=\langle\Phi_{h,0}(z),\Phi_{h,0}(z')\rangle$.
Set
\[
 R=\frac{135K}{64H}\sqrt{\frac dm},\qquad
 \tau=R\sqrt d=\frac{135Kd}{64H\sqrt m}.
\]
A $1/(2m)$-net of the observation sphere has at most
$(1+4m\sqrt d)^d$ points. At each initial neuron pair, the two
hidden weights coincide. The spherical strip estimate
$\Pr\{|w^\top x|\le u\}\le u$, since the coordinate density is at
most $1/2$ on $[-1,1]$, and Hoeffding's inequality
for the $m/2$ independent pairs give, simultaneously in $h,x$,
\begin{equation}\label{eq:dqn-long-strip}
 \frac1m\sum_{s=1}^m
   \mathbf1\{|w_{h,s}^{(0)\top}x|\le\tau\}
 \le3(d+1)\left(1+\frac KH\right)\rho.
\end{equation}
Indeed, apply Hoeffding on the net at width $\tau+1/m$.
The resulting bound $\tau+1/m+\rho$ is at most the displayed
right side. For strips of width $1/m$, using the sharper net
distance $1/(2m)$ gives the uniform bound $3/m+\rho\le2\rho$, since $m\ell\ge9$.

At each pair of net points, the independent paired summands in
the non-bias kernel have range length at most $4d$. Hoeffding
therefore bounds their net deviation by $4d\rho$.
ReLU products are Lipschitz, and a changed gate between a point
and its net representative lies in one of the thin strips just
controlled. Extending both arguments from the net gives
\begin{equation}\label{eq:dqn-long-initial-kernel}
 \max_h\|K_{h,0}-K_\infty\|_\infty
       \le4(d+1)^2\rho.
\end{equation}
For completeness, the finite-kernel transfer costs at most
$4d\rho+2\sqrt d/m\le6(d+1)\rho$, and the population transfer
at most $(d+2\sqrt d)/m\le3(d+1)\rho$; these costs and the net
deviation are covered by the displayed constant. Cross-action head products have mean zero. Both
members of each neuron pair give the same kernel contribution,
so these concentration estimates use the $m/2$ independent pairs.
The logarithm in \eqref{eq:dqn-long-log} covers the net
points, action pairs, stages, and both strip events.

\paragraph{2. Direct concentration of the current SGD direction.}
If independent Hilbert-space-valued variables have norm at most
$B$ and $\ell\ge3$, then
\begin{equation}\label{eq:dqn-long-hilbert}
 \left\|\frac1N\sum_{i=1}^N(Z_i-\mathbb EZ_i)\right\|
       \le2B\sqrt{\ell/N}
\end{equation}
except with probability $e^{-\ell}$. The expected norm on the
left is at most $B/\sqrt N$ by the second-moment bound. Changing
one variable changes that norm by at most $2B/N$.
Bounded differences and $1+\sqrt{2\ell}\le2\sqrt\ell$ for
$\ell\ge3$ prove the displayed inequality.

Before collection batch $j$, both $Q_{h,j-1}$ and the copied
value continuation are fixed. No positive update occurs while its $n$
episodes are collected. Conditional on the preceding history,
the new stage-$h$ episode records are independent and their
state--action marginal is $\nu$: averaging the signed transition
over a uniform action gives the next-state law $\mu$.

Let $\Phi_{h,j-1}(z)$ be the current full parameter gradient.
Apply \eqref{eq:dqn-long-hilbert} to the vector
\[
 \Phi_{h,j-1}(X,A)\{Y-Q_{h,j-1}(X,A)\}.
\]
To define its concentration event on every history, use auxiliary
values as follows: clip the copied value continuation and current
prediction to $[-5/4,5/4]$, and project the current gradient
vector onto the Euclidean ball of radius $\sqrt{3(d+1)}$.
The terminal label is its observed reward, already in $[0,1]$.
These bounded auxiliary random variables define the concentration
event. Their norms are at most
$B=(5/2)\sqrt{3(d+1)}$. They coincide with the actual update variables
up to the stopping time below, and the bootstrap will remove that
stopping time.

The error in the collection-average direction is at most
$5\sqrt{3(d+1)}\sqrt{\ell/n}$. Conditional on the buffer,
current parameters, and the stage count $b_{h,j}$, the same
argument bounds replay resampling by
$5\sqrt{3(d+1)}\sqrt{\ell/b_{h,j}}$.
Each fresh batch supplies one positive step, so its current gradient
is fixed before collection. Conditional concentration therefore applies
directly to the realized current parameter gradient. Towering the
conditional probabilities and taking the union over $h,j$ establishes
both direction events.
Together with Step 1, their total failure probability is less
than $\delta$ by \eqref{eq:dqn-long-log}.

\paragraph{3. Parameter movement and a local population recurrence.}
Work on these events. Provisionally stop at the first positive
step for which a prediction exceeds $5/4$ in absolute value or
a hidden-row norm or head-entry absolute value exceeds $9/8$.
Before such a step all residuals have absolute value at most
$5/2$. Each row and head increment is consequently at most
$(45\eta_{h,j}/16)\sqrt{d/m}$. Equation \eqref{eq:dqn-long-time}
bounds the total motion by $R$. The width condition gives
$R\le45\sqrt2/512<1/8$. Thus every row and head, including the
segments between consecutive iterates, is bounded by $9/8$; the
provisional parameter threshold cannot be reached. Each bias moves by at most $3K/H$. In particular,
\begin{equation}\label{eq:dqn-long-current-gradient}
 \sup_z\|\Phi_{h,j-1}(z)\|^2
       \le1+2(9/8)^2d\le3(d+1).
\end{equation}
The auxiliary vectors of Step 2 therefore equal the actual
vectors for every step considered in this stopped argument.

Let $f_{h,j}$ be the conditional mean of the actual copied target:
\[
 f_{h,j}(x,a)=
 \begin{cases}
  \mathbb E[\max_{a'}Q_{h+1,j-1}(X',a')\mid x,a],&h<H,\\
  Q_H^\star(x,a),&h=H.
 \end{cases}
\]
We claim the exact step has the representation
\begin{equation}\label{eq:dqn-long-local}
\begin{aligned}
 Q_{h,j}&=Q_{h,j-1}
       +\eta_{h,j}\mathcal T_0(f_{h,j}-Q_{h,j-1})+\xi_{h,j},
 \\
 \|\xi_{h,j}\|_\infty
 &\le32\eta_{h,j}(d+1)
      \left[\chi_{\mathrm{data}}
               +(d+1)\left(1+\frac KH\right)\rho\right].
\end{aligned}
\end{equation}
First multiply the two direction concentration bounds by the
evaluation norm in \eqref{eq:dqn-long-current-gradient}.
Replacing the replay gradient by its current population gradient
costs at most
\[
 15(d+1)\eta_{h,j}\sqrt\ell
            \left(n^{-1/2}+b_{h,j}^{-1/2}\right)
 \le15(d+1)\eta_{h,j}\chi_{\mathrm{data}}.
\]

Next compare the current tangent kernel to its initial kernel.
Changes in the head-feature products cost at most $17Rd/8$, and
changes in the hidden-part head products at most $17Rd/8$.
A changed gate belongs to the initial strip in
\eqref{eq:dqn-long-strip}. Using its net bound gives
\begin{align*}
 \|K_{h,j-1}-K_{h,0}\|_\infty
 &\le\frac{17}{4}Rd+2d(\tau+1/m+\rho)
 \le7(d+1)^2\left(1+\frac KH\right)\rho.
\end{align*}
Here $d^{3/2}\le d^2/4+d$. Adding
\eqref{eq:dqn-long-initial-kernel} and multiplying
by the residual bound $5/2$ makes the two kernel-replacement
errors at most
\[
 23(d+1)^2\eta_{h,j}
                  \left(1+\frac KH\right)\rho.
\]

Finally, the sharper row and head increment bound at this step
is $(45\eta_{h,j}/16)\sqrt{d/m}$. A scalar ReLU remainder
vanishes unless its gate changes. Its contributing initial-strip
fraction satisfies the net bound used in \eqref{eq:dqn-long-strip}, because
both endpoints remain within distance $R$ of initialization.
The product of the row and head increments is the remaining
second-order term. Consequently
\begin{align*}
 \|\operatorname{Rem}_{h,j}\|_\infty
 &\le\frac{405\eta_{h,j}d}{128}(\tau+1/m+\rho)
       +\frac{2025\eta_{h,j}^2d^{3/2}}{256\sqrt m}
 \le9\eta_{h,j}(d+1)^2
                      \left(1+\frac KH\right)\rho.
\end{align*}
Here $\eta_{h,j}\le3/8$, $K/H\ge1$, $\ell\ge1$,
$d/(d+1)^2\le1/4$, and $d^{3/2}\le(d+1)^2$ suffice for the last
inequality. The sampling, kernel, and remainder terms are bounded
by the right side of \eqref{eq:dqn-long-local}.

\paragraph{4. One stable semigroup for the whole trajectory.}
For steps $0\le\eta_i\le1$, write
$\mathcal S=\prod_i(I-\eta_i\mathcal T_0)$ and
$s=\sum_i\eta_i$. Every spectral value satisfies
$0\le\lambda\le\|\mathcal T_0\|_{2\to2}\le1$, so
\[
 0\le1-\prod_i(1-\eta_i\lambda)\le\min\{1,s\lambda\},
 \qquad
 \left[1-\prod_i(1-\eta_i\lambda)\right]^2\le s\lambda.
\]
The reproducing-kernel evaluation inequality therefore gives
\[
 |[(I-\mathcal S)f](z)|
 \le\sqrt{K_\infty(z,z)}
        \|\mathcal T_0^{-1/2}(I-\mathcal S)f\|_2
 \le\sqrt{K_0s}\,\|f\|_2.
\]
The inverse is zero on the null space, and the spectral inequality
bounds the expression there. One may equivalently prove this first
for finite spectral sums and pass to the $L^2$ limit. Since $\nu$
is a probability measure,
\begin{equation}\label{eq:dqn-long-semigroup}
 \|\mathcal S f\|_\infty
       \le(1+\sqrt{K_0s})\|f\|_\infty.
\end{equation}

Define the reference, using the actual conditional targets, by
\[
 q_{h,0}=0,\qquad
 q_{h,j}=q_{h,j-1}
       +\eta_{h,j}\mathcal T_0(f_{h,j}-q_{h,j-1}).
\]
This is a pathwise definition even though a copied target depends
on preceding online updates. Subtracting from
\eqref{eq:dqn-long-local}, applying
\eqref{eq:dqn-long-semigroup} to every forcing term, and
using \eqref{eq:dqn-long-time}, gives at every prefix
\begin{equation}\label{eq:dqn-long-remainder}
 z_{h,j}:=Q_{h,j}-q_{h,j},\qquad
 \|z_{h,j}\|_\infty\le E_{\mathrm{run}}.
\end{equation}

Every $f_{h,j}$ belongs to
\[
 \mathcal F_h=\{c+a\langle v,g_h(x)\rangle:
                    c\in\mathbb R,\ v\in\mathbb R^{D-1}\}.
\]
The transition density maps every bounded copied value continuation
into this quadratic space.
The space is invariant under $\mathcal T_0$, so write
$q_{h,j}(x,a)=c_{h,j}+a u_{h,j}(x)$. Put
\[
 \omega_{h,j}^{(0)}:=\frac{\lambda_0p_{h,j}}4,
 \qquad \omega_{h,j}:=\frac{\lambda_2p_{h,j}}4.
\]
These multipliers lie in $[0,1]$.

\paragraph{5. A bounded ideal trajectory closes the bootstrap.}
Define ideal scalar coefficients $\bar c_{h,0}=\bar\beta_{h,0}=0$.
At the terminal stage let
\[
 \bar c_{H,j}=(1-\omega_{H,j}^{(0)})\bar c_{H,j-1}
                                  +\omega_{H,j}^{(0)}c_H^\star,
 \qquad
 \bar\beta_{H,j}=(1-\omega_{H,j})\bar\beta_{H,j-1}
                                  +\omega_{H,j}\beta_H^\star.
\]
For $h<H$ use
\begin{align*}
 \bar c_{h,j}
  &=(1-\omega_{h,j}^{(0)})\bar c_{h,j-1}
      +\omega_{h,j}^{(0)}(\bar c_{h+1,j-1}+\zeta_d\bar\beta_{h+1,j-1}),\\
 \bar\beta_{h,j}
  &=(1-\omega_{h,j})\bar\beta_{h,j-1}
      +\omega_{h,j}\frac{\kappa_d}{4D}\bar\beta_{h+1,j-1}.
\end{align*}
These are the exact population reference updates when each target is supplied
by this ideal trajectory itself. Axial value continuation projections
give the displayed recursions. Induction and the optimal coefficient
identities show
\[
 0\le\bar c_{h,j}\le c_h^\star,
 \qquad 0\le\bar\beta_{h,j}\le\beta_h^\star.
\]
Thus $\bar q_{h,j}(x,a)=\bar c_{h,j}
                         +a\bar\beta_{h,j}g_{h,1}(x)$ satisfies
$\|\bar q_{h,j}\|_\infty\le1$.

Let
\[
 B_{h,j}=\|u_{h,j}-\bar\beta_{h,j}g_{h,1}\|_\infty,
 \qquad \Delta c_j=\max_h|c_{h,j}-\bar c_{h,j}|,
 \qquad B_{h,0}=\Delta c_0=0.
\]
Set $B_{H+1,j}=0$.
The difference between an actual copied value continuation and its ideal
counterpart, after subtracting its common constant, is at most
$B_{h+1,j-1}+E_{\mathrm{run}}$. Since
$\mathbb E_\mu g_{h+1}=0$ and $\|g_h\|_2\le1$, the corresponding
quadratic target difference is at most
$(B_{h+1,j-1}+E_{\mathrm{run}})/(4D)$. It follows that
\[
 B_{h,j}\le(1-\omega_{h,j})B_{h,j-1}
    +\frac{\omega_{h,j}}{4D}
                      (B_{h+1,j-1}+E_{\mathrm{run}}).
\]
Because $1/(4D)\le1/8$, induction gives $B_{h,j}\le E_{\mathrm{run}}/7$.
The terminal-stage difference is zero and obeys the same bound.
For the constants, their target error is at most
$\Delta c_{j-1}+8E_{\mathrm{run}}/7$. Consequently
\[
 \Delta c_j\le\Delta c_{j-1}
           +\frac{8E_{\mathrm{run}}}{7}\max_h \omega_{h,j}^{(0)},
 \qquad
 \Delta c_j\le\frac{6j}{7H}E_{\mathrm{run}}
             \le\frac{6K}{7H}E_{\mathrm{run}},
\]
where $\lambda_0\le2$ and
\eqref{eq:dqn-long-time} give
$\max_h \omega_{h,j}^{(0)}\le3/(4H)$.
Since $K/H\ge1$, the ideal envelope and
\eqref{eq:dqn-long-remainder} now yield
\[
 \|Q_{h,j}\|_\infty
 \le1+\left(\frac{6K}{7H}+\frac17+1\right)E_{\mathrm{run}}
 \le1+(K/H+1)E_{\mathrm{run}}
 \le\frac54<2.
\]
This also holds on the step that would first violate the provisional
output threshold. It closes the bootstrap and validates all preceding
events and estimates for the full trajectory.

\paragraph{6. Concurrent contraction of all action contrasts.}
Put $A_{h,j}=\|u_{h,j}-\beta_h^\star g_{h,1}\|_\infty$ and
$A_{H+1,j}=0$. For a nonterminal stage, the difference between
its actual copied value continuation and $V_{h+1}^\star$, after
subtracting $c_{h+1,j-1}-c_{h+1}^\star$, has magnitude at most
$A_{h+1,j-1}+E_{\mathrm{run}}$. The same cancellation of its
common constant in the transition projection gives
\begin{equation}\label{eq:dqn-long-contrast-recursion}
 A_{h,j}\le(1-\omega_{h,j})A_{h,j-1}
       +\frac{\omega_{h,j}}{4D}
                       (A_{h+1,j-1}+E_{\mathrm{run}}).
\end{equation}
The terminal target is exact, so this upper bound holds there
too. Initially $\max_h A_{h,0}\le\beta_H^\star$.
Let $\vartheta=1-7\lambda_2/(64H)$. The inequalities
$\omega_{h,j}\ge\lambda_2/(8H)$ and $1/(4D)\le1/8$ imply
\[
 (1-\omega_{h,j})\left(\beta_H^\star\vartheta^{j-1}
                                  +\frac{E_{\mathrm{run}}}7\right)
 +\frac{\omega_{h,j}}{4D}
       \left(\beta_H^\star\vartheta^{j-1}
                                  +\frac{8E_{\mathrm{run}}}7\right)
 \le\beta_H^\star\vartheta^j+\frac{E_{\mathrm{run}}}7.
\]
Induction in \eqref{eq:dqn-long-contrast-recursion} therefore
proves
\[
 \max_h A_{h,K}
 \le\beta_H^\star\exp\!\left(-\frac{7\lambda_2K}{64H}\right)
                      +\frac{E_{\mathrm{run}}}7.
\]
Set $\gamma_h=c_h^\star-c_{h,K}$. Step 5 bounds
\[
 |\gamma_h|
 \le c_h^\star-\bar c_{h,K}+\Delta c_K
 \le1+\frac{6K}{7H}E_{\mathrm{run}}<2.
\]
The common constants cancel after this analytical shift.
Adding the final remainder in
\eqref{eq:dqn-long-remainder} proves
\eqref{eq:dqn-long-centered}.
In particular, $K$ of order
$(H/\lambda_2)\log n$ suffices for the displayed optimization
term to be of order $1/n$.

\end{proofof}

\begin{proofof}{Lemma~\ref{lem:dqn-axial-stage-projection}}
\paragraph{The axial coordinate and its projection.}
Fix any $h\ge2$ and write $u=u_h$. The matrix $duu^\top-I_d$
has trace zero and squared Frobenius norm
$(d-1)^2+(d-1)=d(d-1)$. Thus $E_{h,1}$ has unit Frobenius norm
and admits a trace-free orthonormal completion for every allowed
$D$. By rotational invariance, $Z=(u^\top X)^2$ has law $dU$,
where $U\sim\operatorname{Beta}(1/2,(d-1)/2)$. Its density is
\begin{equation}\label{eq:dqn-axial-squared-coordinate-density}
 f_d(s)=\frac{\Gamma(d/2)}{\sqrt{\pi d}\,\Gamma((d-1)/2)}
             s^{-1/2}(1-s/d)^{(d-3)/2}\mathbf1\{0<s<d\}.
\end{equation}
This law is independent of $h$ and $u_h$, so the moments
$\zeta_d$ and $\kappa_d$ are common to all stages $h\ge2$.
The moments $\mathbb E Z=1$ and $\mathbb E Z^2=3d/(d+2)$ give
$\mathbb E(Z-1)^2=2(d-1)/(d+2)>0$, and hence $\zeta_d>0$.

For $z>0$, Cauchy--Schwarz in the gamma integral gives
$\Gamma(z+1/2)^2\le\Gamma(z)\Gamma(z+1)=z\Gamma(z)^2$.
At $z=(d-1)/2$, this bounds the normalizing coefficient in
\eqref{eq:dqn-axial-squared-coordinate-density} by $\sqrt{(d-1)/(2\pi d)}$.
Since $d\ge4$,
\[
 \mathbb E[(Z-1)^2\mathbf1\{Z\le1\}]
 \le\sqrt{\frac{d-1}{2\pi d}}\int_0^1(1-s)^2s^{-1/2}\,ds
 =\frac{16}{15}\sqrt{\frac{d-1}{2\pi d}}.
\]
For $d=4$, substituting $s=4\sin^2\theta$ gives
$\mathbb E[(Z-1)^2\mathbf1\{Z\le1\}]
=(2/\pi)\int_0^{\pi/6}(1+\cos6\theta)\,d\theta=1/3$,
and hence $\kappa_4=1/36$. For $d\ge5$, consequently,
\begin{align*}
 d^2\kappa_d
 &=\frac d{d-1}\mathbb E[(Z-1)|Z-1|]
 =\frac{2d}{d+2}
      -\frac{2d}{d-1}\mathbb E[(Z-1)^2\mathbf1\{Z\le1\}]\\
 &\ge\frac{10}{7}-\frac{16}{3\sqrt{10\pi}}>\frac49.
\end{align*}
The last numerical inequality follows from $\pi>3$.
Thus $\kappa_d\ge4/(9d^2)\ge1/(8d^2)$. The opposite bound follows from
$g_{h,1}|g_{h,1}|\le g_{h,1}^2$ and
$\mathbb E_\mu g_{h,1}^2=2/[d(d+2)]$.

To identify the vector projection, let
$M=\mathbb E[XX^\top|g_{h,1}(X)|]$.
The scalar weight and the sphere law are invariant under every
orthogonal map fixing $u$, so $M=cI_d+q uu^\top$ for some $c,q$.
For $\ell\ge2$, trace-freeness and orthogonality give
\[
 0=\operatorname{tr}(E_{h,\ell}E_{h,1})
   =\frac{d u^\top E_{h,\ell}u}{\sqrt{d(d-1)}},\qquad
 \operatorname{tr}(E_{h,\ell})=0.
\]
Thus
$\mathbb E[g_{h,\ell}(X)|g_{h,1}(X)|]
 =d^{-1}\operatorname{tr}(E_{h,\ell}M)=0$.
The first coordinate equals $\kappa_d$, proving
\eqref{eq:dqn-axial-projection-and-bounds} for any completion.

\paragraph{Margins near the optimal decision boundary.}
Put $L_d=\sqrt{d(d-1)}$. If $0\le z\le1/(2L_d)$, then
$|g_{h,1}(X)|\le z$ is equivalent to $|Z-1|\le L_dz$, an
interval inside $[1/2,3/2]$. On this interval the density in
\eqref{eq:dqn-axial-squared-coordinate-density} is at most
$1/\sqrt\pi$. The event therefore has probability at most
$2L_dz/\sqrt\pi\le2dz$. If $z\ge1/(2L_d)$, then
$2dz\ge d/L_d>1$, and the same bound follows from the trivial
probability bound. Multiplying by $|g_{h,1}|\le z$ on the event
gives its truncated first moment in
\eqref{eq:dqn-axial-stage-margin}.

\paragraph{Admissibility and exact dimension.}
The terminal reward has the required factorization because
\[
 D^{-1}(\mathbf1+\tfrac12 a B g_H(x))^\top
      \left(\frac{\mathbf1}{2}+\frac{Be_1}{4}\right)
 =\frac12+\frac{a g_{H,1}(x)}{8D}.
\]
Every coordinate of $Be_1$ has magnitude at most one. Hence
$\vartheta_H\in[1/4,3/4]^D\subset[0,1]^D$.
Taking $r_H^{\max}=1$ and $r_h^{\max}=0$, $\vartheta_h=0$ for
$h<H$ supplies the earlier zero rewards and a unit episode-return
bound.
The transition factorization of Lemma~\ref{lem:qmdp-exact-rank}
applies with $\tau_h=\omega_h=1/2$ and $T_{h,a}=aI_r$.
Its density belongs to $[1-1/(4D),1+1/(4D)]$.
For every nonterminal stage and either fixed action,
\[
 P_h1=1,\qquad
 (P_hg_{h+1,\ell})(x,a)
       =\frac{a}{2Dd(d+2)}g_{h,\ell}(x).
\]
The $D$ functions $1,g_{h,1},\ldots,g_{h,D-1}$ are linearly
independent: the last $D-1$ have zero mean and positive definite
covariance $2I_r/[d(d+2)]$. Thus the transition image has
dimension at least $D$, whereas the displayed factorization has
dimension $D$. Every nonterminal transition has rank exactly $D$,
and no smaller common reward-transition factorization is possible.
The terminal transition to $\partial$ is represented by the same
$D$ features because their coordinates sum to one.

\paragraph{Backward propagation of the optimal contrast.}

Define the positive optimal action-contrast coefficients and the
optimal constant coefficients by
\stepcounter{equation}
\begin{equation*}
\begin{aligned}
 \beta_h^\star
 &=\frac1{8D}\left(\frac{\kappa_d}{4D}\right)^{H-h},\\
 c_H^\star&=\frac12,\qquad
 c_h^\star=c_{h+1}^\star+\zeta_d\beta_{h+1}^\star\quad(h<H).
\end{aligned}
\end{equation*}
At stage $H$, \eqref{eq:dqn-horizon-optimal-values} follows
directly from the terminal reward and
$\beta_H^\star=1/(8D)>0$.
Suppose it holds at stage $h+1$, where $h<H$. This value
stage is axial. Maximization gives
$V_{h+1}^\star=c_{h+1}^\star+
                    \beta_{h+1}^\star|g_{h+1,1}|$, and the
transition and projection identities yield
\begin{align*}
 P_{h,a}V_{h+1}^\star(x)
 &=c_{h+1}^\star+\zeta_d\beta_{h+1}^\star
     +\frac{a\beta_{h+1}^\star}{4D}
       g_h(x)^\top\mathbb E_\mu[g_{h+1}|g_{h+1,1}|]\\
 &=c_{h+1}^\star+\zeta_d\beta_{h+1}^\star
     +\frac{a\kappa_d\beta_{h+1}^\star}{4D}g_{h,1}(x).
\end{align*}
Since $r_h=0$, this proves the coefficient recursion and hence
\eqref{eq:dqn-horizon-optimal-values} by induction down to $h=1$.
Each $\beta_h^\star$ is positive. A nonzero trace-free symmetric
matrix has eigenvalues of both signs; its quadratic form therefore
has both signs on open spherical sets of positive measure.
Strictly positive transition densities preserve positive
probability for these regions at every stage. This proves the
claim about the two optimal action regions, including arbitrary
allowed $E_{1,1}$.

Finally, averaging $Q_k^\star(x,+1)$ and $Q_k^\star(x,-1)$
gives the constant $c_k^\star$. With uniform actions at stage $k$
and optimal actions afterward, its value continuation value is thus
constant in its observation. Zero earlier rewards and
$P_h1=1$ propagate this constant through every earlier stage.
Their action contrasts then vanish, as asserted.
\end{proofof}

\begin{proofof}{Lemma~\ref{lem:dqn-balanced-margin}}
Rotational invariance gives $O^\top X\stackrel{\mathrm d}=X$. Hence
\begin{align*}
 g_{1,1}(X)
 &=\frac1{d\sqrt d}
 \left\{\sum_{j=1}^{d/2}(O^\top X)_j^2
              -\sum_{j=d/2+1}^{d}(O^\top X)_j^2\right\}
 =\frac1{\sqrt d}
 \left\{2\sum_{j=1}^{d/2}\frac{(O^\top X)_j^2}{d}-1\right\}.
\end{align*}
The last equality uses $\sum_j(O^\top X)_j^2=d$.
Let $q=O^\top X/\sqrt d$, which is uniform on the unit sphere in
$\mathbb R^d$. The spherical projection law
\citep[Section~4.1.1]{dqn_lotz2020projection} gives
$\sum_{j=1}^kq_j^2\sim\operatorname{Beta}(k/2,(d-k)/2)$ for
$0<k<d$. Here $d\ge4$ is even, so $k=d/2$ is admissible and
\begin{align*}
 U:=\sum_{j=1}^{d/2}\frac{(O^\top X)_j^2}{d}
 &=\sum_{j=1}^{d/2}q_j^2
       \sim\operatorname{Beta}(d/4,d/4),\\
 f_U(u)&=\frac{u^{d/4-1}(1-u)^{d/4-1}}
                   {\mathrm B(d/4,d/4)},\qquad 0<u<1.
\end{align*}
The preceding identity for $g_{1,1}(X)$ therefore gives
$g_{1,1}(X)=(2U-1)/\sqrt d$.

For $|z|<1/\sqrt d$, change variables to
$u=(1+\sqrt d\,z)/2$. Its derivative with respect to $z$ is
$\sqrt d/2$, so
\begin{align*}
 f_{g_{1,1}(X)}(z)
 &=\frac{\sqrt d}{2\mathrm B(d/4,d/4)}
       \left(\frac{1+\sqrt d\,z}{2}\right)^{d/4-1}
       \left(\frac{1-\sqrt d\,z}{2}\right)^{d/4-1}\\
 &=\frac{\sqrt d}{2\mathrm B(d/4,d/4)}
       \left(\frac{1-dz^2}{4}\right)^{d/4-1}
 \le\frac{\sqrt d\,2^{1-d/2}}{\mathrm B(d/4,d/4)}.
\end{align*}
Here $d/4-1\ge0$ and $1-dz^2\le1$.
Equality holds at zero; for $d=4$ the density is constant.

For an elementary bound on the last expression, set $a=d/4\ge1$.
The beta--gamma identity and duplication formula
\citep[Eqs.~(5.12.1), (5.5.5)]{dqn_nist_dlmf} apply because $a>0$.
Step~3 of the proof of Lemma~\ref{lem:dqn-uniform-step} gives
$\Gamma(a+1/2)^2\le a\Gamma(a)^2$; positivity of the gamma factors
therefore gives $\Gamma(a+1/2)/\Gamma(a)\le\sqrt a$. Consequently,
\begin{align*}
 \frac{\sqrt d\,2^{1-d/2}}{\mathrm B(d/4,d/4)}
 &=\sqrt d\,2^{1-2a}\frac{\Gamma(2a)}{\Gamma(a)^2}\\
 &=\frac{\sqrt d}{\sqrt\pi}
                     \frac{\Gamma(a+1/2)}{\Gamma(a)}\\
 &\le\frac{\sqrt{da}}{\sqrt\pi}
   =\frac{d}{2\sqrt\pi}\le\frac{2d}{3},
\end{align*}
where the final inequality follows from $\pi>1$.
Finally extend the density by zero outside its support. For every
$z\ge0$,
\begin{align*}
 \mathbb E[|g_{1,1}(X)|\mathbf1\{|g_{1,1}(X)|\le z\}]
 &=\int_{-z}^{z}|u|f_{g_{1,1}(X)}(u)\,du
 \le\frac{\sqrt d\,2^{1-d/2}}{\mathrm B(d/4,d/4)}
                        \int_{-z}^{z}|u|\,du\\
 &=\frac{\sqrt d\,2^{1-d/2}}{\mathrm B(d/4,d/4)}
                         \,2\int_0^z u\,du
 =\frac{\sqrt d\,2^{1-d/2}}{\mathrm B(d/4,d/4)}z^2
 \le\frac{2d}{3}z^2.
\end{align*}
The specified $E_{1,1}$ has trace zero and squared Frobenius norm
$d/d=1$. The trace-free symmetric matrix space has dimension
$d(d+1)/2-1$, so it has an orthonormal completion for every
$D-1\le d(d+1)/2-1$.
\end{proofof}

\begin{proofof}{Lemma~\ref{lem:dqn-quadratic-policy-loss}}
Let $\widehat\pi$ be their greedy policy and let $\nu_h$ be its
stage-$h$ state law. Then $\nu_1=\mu$ and, for every $h\ge2$,
\[
 1-\frac1{4D}\le\frac{d\nu_h}{d\mu}\le1+\frac1{4D}.
\]
The same density bounds hold for the state laws of any policy.

Every conditional transition density belongs to
$[1-1/(4D),1+1/(4D)]$. These bounds are preserved under mixtures,
uniformly over the previous state distribution and selected action,
so the same constants apply at every $h\ge2$.

If a greedy action of $\widehat Q_h$ is suboptimal at state $x$,
its maximizing property gives
\[
 V_h^\star(x)-Q_h^\star(x,\widehat\pi_h(x))\le2e_h.
\]
The true action gap is $2\beta_h^\star|g_{h,1}(x)|$, so such a
mistake requires $|g_{h,1}(x)|\le e_h/\beta_h^\star$. Optimal
ties have zero gap. For the stagewise expected advantage, define
\[
 \Delta_h:=\mathbb E_{\widehat\pi}\!\left[
 V_h^\star(X_h)-Q_h^\star(X_h,\widehat\pi_h(X_h))\right].
\]
Bellman telescoping gives $J^\star-J(\widehat\pi)=\sum_h\Delta_h$
and hence \eqref{eq:dqn-greedy-margin-transfer}.

\begin{equation}\label{eq:dqn-greedy-margin-transfer}
\begin{aligned}
 J^\star-J(\widehat\pi)
 &\le\sum_{h=1}^H2\beta_h^\star\,
 \mathbb E_{\nu_h}\!\left[|g_{h,1}|
          \mathbf1\{|g_{h,1}|\le e_h/\beta_h^\star\}\right] \le2\sum_{h=1}^H e_h.
\end{aligned}
\end{equation}

Under the balanced condition, Lemma~\ref{lem:dqn-balanced-margin}
bounds the first-stage contribution by
$4de_1^2/(3\beta_1^\star)$. For $h\ge2$, the density bound and
the axial margin bound give
\[
 \Delta_h\le
 4d\left(1+\frac1{4D}\right)\frac{e_h^2}{\beta_h^\star}.
\]
Since $D\ge2$, these bounds imply
\eqref{eq:dqn-balanced-contrast-transfer}. Also $|g_{h,1}|\le1$,
so every stage contribution is at most $2\beta_h^\star$.
This proves \eqref{eq:dqn-geometric-margin-refinement}.

\begin{equation}\label{eq:dqn-balanced-contrast-transfer}
\begin{aligned}
 J^\star-J(\widehat\pi)
 &\le\frac{4d}{3\beta_1^\star}e_1^2
      +4d\left(1+\frac1{4D}\right)
                \sum_{h=2}^H\frac{e_h^2}{\beta_h^\star} \le\frac{9d}{2}\sum_{h=1}^H\frac{e_h^2}{\beta_h^\star}.
\end{aligned}
\end{equation}

For the general first-stage matrix, its contribution is at most
$2e$. At stages $h\ge2$ use
\[
 \Delta_h\le\min\{2\beta_h^\star,9de^2/(2\beta_h^\star)\}.
\]
The assertion is immediate if $e=0$.
Otherwise put $\rho=\kappa_d/(4D)$ and $z=3\sqrt d\,e/2$ for
this calculation. Splitting the geometric sequence at $z$ gives
\[
 \sum_{h=2}^H\min\{\beta_h^\star,z^2/\beta_h^\star\}
 \le\sum_{\beta_h^\star<z}\beta_h^\star
    +z^2\sum_{\beta_h^\star\ge z}\frac1{\beta_h^\star}
 \le\frac{2z}{1-\rho}.
\]
Empty sums contribute zero. Since
$0<\rho\le1/[2Dd(d+2)]\le1/96$ and $d\ge4$, the resulting
coefficient is at most $8\sqrt d$. The deterministic return
bound gives the cap $1$.

Consequently, for $e_h\le e$ at every stage,
\stepcounter{equation}
\begin{equation*}
 J^\star-J(\widehat\pi)
 \le\min\left\{1,
     2e+\frac{6\sqrt d}{1-\kappa_d/(4D)}e\right\}
 \le\min\left\{1,8\sqrt d\,e\right\}.
\end{equation*}

For the exploration term, apply Bellman telescoping directly
under the state laws of the $\varepsilon$-greedy policy. At a
fixed state, its expected optimal advantage is
\[
 (1-\varepsilon)
   \{V_h^\star(x)-Q_h^\star(x,\widehat\pi_h(x))\}
       +\varepsilon\beta_h^\star|g_{h,1}(x)|.
\]
The first term has the same mistake threshold as before. Its
state distribution satisfies the same density bounds, so all
the preceding uniform-error and margin estimates still apply.
The second term sums to at most
\[
 \varepsilon\sum_{h=1}^H\beta_h^\star
 \le\frac{\varepsilon}
          {8D\{1-\kappa_d/(4D)\}}
 \le\varepsilon.
\]
The pointwise bound $|g_{h,1}|\le1$ controls this term under the
executed state laws and proves the exploration claim with the
prescribed treatment of ties.

Each collection episode costs at most one, so $Kn$ episodes
cost at most $Kn$. Summing the deployment bounds proves
\eqref{eq:dqn-horizon-policy-regret-transfer}. The same return
bound gives $\Reg_T\le T$ for every budget.
\end{proofof}

\begin{proofof}{Lemma~\ref{lem:dqn-horizon-schedule}}
During exploration, the equivalent interaction-indexed description
starts with $k=0$ and inserts one gradient call after each
interaction. The call after stage $h$
of episode $t$ therefore has index
\[
 k=H(t-1)+h-1.
\]
At the terminal stage of episode $jn$, this is $Hjn-1$.
These are precisely the positive calls in
\eqref{eq:dqn-horizon-executable-schedule}; all other calls have
step size zero. The copy period is one, so every call ends with
a target copy. Its targets are detached during differentiation.

After the terminal record of episode $jn$ is appended, the last
$Hn$ records consist exactly of the $H$ transitions in each of
episodes $(j-1)n+1,\ldots,jn$. FIFO eviction removes all older
records. For $j=1$ the buffer has just reached capacity.
Zero step sizes between positive updates leave every parameter
fixed, and intervening copies preserve these values. The online
and target networks throughout collection batch $j$ are therefore
$Q_{\cdot,j-1}$, proving \eqref{eq:dqn-horizon-actual-targets}.

Explicitly, the targets at positive update $j$ are
\begin{equation}\label{eq:dqn-horizon-actual-targets}
 Y_{i,h}^{[j]}=R_{i,h}
       +\mathbf1\{h<H\}\max_a Q_{h+1,j-1}(X_{i,h+1},a).
\end{equation}

\paragraph{Each update reaches one earlier stage.}
Before step one all online and copied outputs are zero.
Nonterminal rewards and copied value continuations are zero, so every
nonterminal TD gradient is zero. Only the terminal batch has
potentially nonzero reward residuals. Inductively, before step
$j\le H$, every batch $h\le H-j+1$ is still initialized.
If $h<H-j+1$, its value also has zero copied output, and its
own target and gradient remain zero. The batch $h=H-j+1$ starts
this step at its initial parameters and may now use a nonzero
learned value continuation. All later batches train in the same step.
The stage parameter batches are separate and targets are detached,
so a preceding-stage loss does not update its value's
parameters. This proves the induction and the stated counts.

After step $K$, the algorithm makes no further gradient calls
and executes the fixed greedy policy. During exploration, positive
updates occur only after terminal interactions and end with a target copy.
Every action therefore uses its episode's initial online
predictions, with online and target parameters equal. This proves
the stated lag and drift properties.

\paragraph{Replay counts and collection laws.}

For a global minibatch of size $b$, let $b_{h,j}$ be its number of
stage-$h$ records and put $p_{h,j}=b_{h,j}/b$. Conditional on any
realized buffer, the counts have distribution
$\operatorname{Multinomial}(b;1/H,\ldots,1/H)$. Conditional on the
counts, each stage uses fresh independent uniform replay indices
from that batch's $n$ stage records.
Each positive-step buffer contains exactly $n$ records at each
stage. A uniform replay draw chooses a stage uniformly and,
conditional on that stage, one of its $n$ records uniformly.
The counts and the conditional stagewise replay laws follow,
independently of the record values. For each $(h,j)$, a
multiplicative Chernoff bound gives
\[
 \mathbb P\!\left\{
       b_{h,j}\notin[b/(2H),3b/(2H)]\right\}
 \le2\exp\!\left(-\frac{b}{12H}\right).
\]
There are $HK$ stage-update pairs; a union bound proves
\eqref{eq:dqn-horizon-stage-counts}.

During collection, actions are uniform and independent of the
current observation. Averaging the signed transition over the
two actions gives exactly $\mu$ for every current state. Each
stage therefore has marginal law
$\mu\otimes\operatorname{Unif}(\{-1,1\})$.
Episodes have independent resets. Conditioning
on all initializations and completed collection batches fixes
both the current online network and the next copied target.
The new episodes remain independent with the displayed marginal
law. The argument uses independence across episodes at each fixed
stage, with within-episode dependence specified by the transition kernel.

At step $j=H-h+1$, every earlier update of batch $h$ was zero.
Its initialization is independent of other stage initializations,
and uniform collection actions do not use it. Because targets are
detached, value batches depend only on their own and later
parameter batches, records, and replay draws. Conditioning on
the earlier batches and those value initializations fixes the
value continuation while preserving the paired law of this unused
initialization.
\end{proofof}

\begin{proofof}{Lemma~\ref{lem:dqn-reachable-budget}}
At the first reached update $j_h=H-h+1$, write
\[
 \mathcal G_h
 :={\mathbb E}_{\mu\otimes\operatorname{Unif}(\{-1,1\})}
 \!\left[(X^\top C_{h,a}(\widehat v_h)X)
       \{F_{W_{h,j_h}}(X,a)-F_{W_{h,0}}(X,a)\}\right].
\]
Apply Lemma~\ref{lem:dqn-horizon-iterates} with confidence
$\delta/2$. Its event bounds the copied value continuations by $5/4$
and the stage fractions by $[1/(2H),3/(2H)]$. At the beginning
of collection batch $j_h$, all parameters of stage $h$ are still at their
paired initialization. Condition on the records and replay
randomization in the completed batches, and on the downstream
initializations. The target copied immediately after update $j_h-1$ is then
fixed, while batch $j_h$ supplies $n$ fresh stage records with independent
$\mu\otimes\operatorname{Unif}(\{-1,1\})$ state--action marginals.
Uniform collection actions and the separate stage parameters make
the unused stage-$h$ initialization independent of these records
and this value continuation. The conditional probability bounds are
established under the trajectory law and then intersected with the
tracking event.

\paragraph{The replay error and the actual signal.}
At the selected stage write
$C_{h,a}(\widehat v_h)=aC$, $S_h=\|C\|_F^2/2$, and
$M=M_h=\|C\|_{\mathrm{op}}$. The conditional target mean has the
form $c+a x^\top Cx$, where the offset $c$ is common to the two
actions. On the tracking event the target magnitude is at most
$B_Y=5/4$ ($B_Y=1$ at $h=H$) and
\begin{equation}\label{eq:dqn-signed-feature-coefficient-bound}
 M\le\|C\|_F\le M_*:=\frac5{16Dd},\qquad M^2\le2S_h.
\end{equation}
For $h<H$, the first inequality follows from
$\|\mathbb E_\mu[g_{h+1}\widehat v_h]\|_2\le5/4$ and
orthonormality of the matrices $E_{h,j}$; the terminal coefficient
satisfies it directly.

For this fixed collection window and its first replay minibatch,
the scalar-projection argument below gives
\stepcounter{equation}
\begin{equation*}
 \max\{\|E\|_F,\sqrt m\max_j\|E_j\|_2\}
 \le \frac{3B_Y}{2}\sqrt{d\ell}
                   (n^{-1/2}+b_{h,j_h}^{-1/2})
 \le\mathcal E.
\end{equation*}
Here $E$ is the stage-normalized replay direction minus its
population direction at initialization. Only the one specified
window and update at each stage are used. Conditional Hoeffding
bounds use scalar ranges of length $2B_Y\sqrt{d/m}$.
A $1/4$-net of the unit sphere has at most $9^d$ points and
satisfies $\|z\|\le(32/31)\max_v|v^\top z|$;
$32\sqrt2/31<3/2$ gives the displayed radius after a union
bound over hidden rows and stages. The
logarithm in Lemma~\ref{lem:dqn-horizon-iterates} dominates
$\log(32Hm9^d/\delta)$, which covers both collection and replay
deviations. To formalize the conditional argument on every prefix,
define the concentration event using the copied value continuation clipped
to $[-5/4,5/4]$. This auxiliary value continuation agrees with the actual
value continuation on the tracking event and supplies a bounded
concentration argument on every prefix.

Apply the first event in Lemma~\ref{lem:dqn-feature-init} with
its confidence parameter set to $\delta/4$. The Gram-width
inequality in \eqref{eq:dqn-reachable-budget} is exactly its
width requirement with $A=2$ and $r=D-1$; its simultaneous
failure probability is at most $\delta/16$. The fixed-window
collection and replay concentration events above each have
failure probability at most $\delta/8$, after the union over
stages. For the additional nonterminal certificate, also construct
the collection and replay events for the value continuation used at update $2$
clipped to $[-3\eta_{H,1}/8,3\eta_{H,1}/8]$, with the
prefix-measurable $\eta_{H,1}=p_{H,1}/2$, each with failure probability at
most $\delta/32$. The first-step prediction event below costs
$\delta/8$. The same logarithm covers these events: it dominates
$\log(2^{10}m9^d(1+4m\sqrt d)^d/\delta)$.
The sum of the stated failure probabilities, including the
tracking event, is at most $\delta$.

\paragraph{Cancellation of the common target offset.}
For each initial row set
\[
 M_j=\frac{U_{+,j}^{(0)}-U_{-,j}^{(0)}}2C,\qquad
 \chi_j=\frac{U_{+,j}^{(0)}+U_{-,j}^{(0)}}2c,\qquad
 \Gamma_j=2M_jw_j-(w_j^\top M_jw_j)w_j.
\]
Because both initial head entries are Rademacher, $\chi_jM_j=0$
identically. If the entries are equal, then $M_j=0$ and this row
contributes zero to the feature covariance for every hidden weight.
If they are opposite, then $\chi_j=0$ and the population hidden
gradient on this row is its quadratic gradient.

By Lemma~\ref{lem:dqn-feature-moments}, the covariance functional
at fixed target is
\[
 \mathbb E[(X^\top C_aX)F_W(X,a)]
 =\frac{\alpha_d}{\sqrt m}
                  \sum_j\frac{w_j^\top M_jw_j}{\|w_j\|}.
\]
Its initial gradient is
$G_j^{\mathrm{quad}}=\alpha_d\Gamma_j/\sqrt m$.
The exact cancellation and the initialization Gram event give
\[
 \langle G^{\mathrm{quad}},G\rangle
   =\|G^{\mathrm{quad}}\|_F^2
   \ge\frac{\alpha_d^2S_h}{2d},
 \qquad
 \|G^{\mathrm{quad}}\|_F\le2\alpha_d M.
\]
Consequently the replay cross term has magnitude at most
$\|G^{\mathrm{quad}}\|_F\mathcal E$. On the contributing rows,
the aggregate displacement is at most
$\eta_{h,j_h}(\|G^{\mathrm{quad}}\|_F+\mathcal E)$.
Each such row moves by at most
$\eta_{h,j_h}(2\alpha_dM+\mathcal E)/\sqrt m$.
The budget, $\alpha_d\le1/2$, and
\eqref{eq:dqn-signed-feature-coefficient-bound} imply
\begin{equation}\label{eq:dqn-signed-feature-width-check}
 \frac{\eta_{h,j_h}(2\alpha_dM_*+\mathcal E)}{\sqrt m}
 \le\frac12,\qquad
 \sqrt m\ge128d\alpha_dM_*\eta_{h,j_h}.
\end{equation}
Indeed, $\mathcal E\le2$, $\eta_{h,j_h}\le3/(4H)$, and
the width requirement in \eqref{eq:dqn-reachable-budget}
gives $\sqrt m\ge8(d+1)$. The second right side is at most
$15/(HD)\le15/4$, whereas $8(d+1)\ge40$; the first
inequality follows from the same bounds.

\paragraph{A finite feature increase with an explicit noise term.}
All contributing row norms remain at least $1/2$ along the update
segment. Lemma~\ref{lem:dqn-feature-hessian} and Taylor's theorem
therefore yield, writing $\eta=\eta_{h,j_h}$,
\begin{equation}\label{eq:dqn-signed-feature-finite-change}
 \mathcal G_h\ge
 \eta\|G^{\mathrm{quad}}\|_F^2
 -\eta\|G^{\mathrm{quad}}\|_F\mathcal E
 -\frac{3\alpha_dM\eta^2}{\sqrt m}
                         (\|G^{\mathrm{quad}}\|_F+\mathcal E)^2.
\end{equation}
The following bounds hold uniformly for $S_h\ge0$. First,
\[
 \|G^{\mathrm{quad}}\|_F\mathcal E
 \le\frac{12}{25}\|G^{\mathrm{quad}}\|_F^2
       +\frac{25}{48}\mathcal E^2,
 \qquad
 \|G^{\mathrm{quad}}\|_F^2
 \ge\frac{\alpha_d^2S_h}{2d}.
\]
The second inequality in \eqref{eq:dqn-signed-feature-width-check}
gives $3\alpha_dM\eta/\sqrt m\le3/(128d)<1/100$.
Using $(u+v)^2\le2u^2+2v^2$, the Taylor remainder is at most
\begin{align*}
 \frac{3\alpha_dM\eta^2}{\sqrt m}
                  (\|G^{\mathrm{quad}}\|_F+\mathcal E)^2
 &\le\frac{6\alpha_dM_*\eta^2}{\sqrt m}
                  (\|G^{\mathrm{quad}}\|_F^2+\mathcal E^2)
 \le\frac{\eta}{50}
                  (\|G^{\mathrm{quad}}\|_F^2+\mathcal E^2).
\end{align*}
Substitution in \eqref{eq:dqn-signed-feature-finite-change},
using $1-12/25-1/50=1/2$ and $25/48+1/50<1\le d/4$, gives
\[
 \mathcal G_h\ge
 \frac{\eta\alpha_d^2S_h}{4d}-\frac14\eta d\mathcal E^2.
\]
This bound also holds when $S_h=0$, when the covariance functional
itself is zero. Since
\[
 \mathcal E^2
 \le\frac{675}{64}d\ell
                              \left(\frac1n+\frac Hb\right),
\]
the effective-step bounds give \eqref{eq:dqn-reachable-gain}.
Under \eqref{eq:dqn-feature-signal-resolution},
$\mathcal E/\|G^{\mathrm{quad}}\|_F\le\sqrt{2/d}\le1/\sqrt2$.
Returning to \eqref{eq:dqn-signed-feature-finite-change} gives
\[
 \mathcal G_h\ge\eta\|G^{\mathrm{quad}}\|_F^2
                  \left(1-\frac1{\sqrt2}-\frac3{100}\right)
 \ge\frac{\eta\alpha_d^2S_h}{8d}
 \ge\frac{\alpha_d^2S_h}{32Hd}.
\]
For the terminal stage, impose the sufficient sampling conditions
\begin{equation}\label{eq:dqn-terminal-feature-budget}
 n\ge\frac{1152D^2d^5\ell}{\alpha_d^2},\qquad
 b\ge\frac{2304HD^2d^5\ell}{\alpha_d^2}
\end{equation}
Finally, \eqref{eq:dqn-terminal-feature-budget} implies
\[
 n^{-1/2}+\sqrt{2H/b}
 \le\frac{\alpha_d}{12\sqrt2\,Dd^{5/2}\sqrt\ell}.
\]
Substituting $S_H=(\beta_H^\star)^2/(2d^2)$ and
$\beta_H^\star=1/(8D)$ verifies
\eqref{eq:dqn-feature-signal-resolution} with the terminal radius
$4\mathcal E/5$, since $B_Y=1$ at this stage.
Equation~\eqref{eq:dqn-resolved-feature-gain} gives
\eqref{eq:dqn-terminal-feature-gain}.
\paragraph{A nonterminal feature certificate after one terminal step.}
Assume additionally
\begin{equation}\label{eq:dqn-endstage-feature-budget}
\begin{gathered}
 \sqrt\ell\left(\frac1{\sqrt n}+\sqrt{\frac{2H}{b}}
                         +\frac1{\sqrt m}\right)
     \le\frac{\lambda_2}{128Dd^2(d+1)^2},\\
 n\ge\frac{2^{16}D^4d^9\ell}{\alpha_d^2\lambda_2^2},\qquad
 b\ge\frac{2^{17}HD^4d^9\ell}{\alpha_d^2\lambda_2^2},
\end{gathered}
\end{equation}
Set $\eta=\eta_{H,1}$ and put
\[
 q_{H,1}(x,a)=\frac{\eta\lambda_0}{4}
             +\frac{\eta\lambda_2\beta_H^\star}{2}a g_{H,1}(x),
 \qquad v_1(x)=\max_a q_{H,1}(x,a).
\]
To control the actual finite update, apply
Lemma~\ref{lem:dqn-uniform-step} with confidence $\delta/8$,
targets $R_H\in[0,1]$, and the actual effective step size
$\eta\le3/(4H)\le1$. Its population response is exactly
$q_{H,1}$. Conditional on the realized terminal-stage minibatch
count, the lemma gives
\begin{equation}\label{eq:dqn-terminal-first-step-accuracy}
 \|Q_{H,1}-q_{H,1}\|_\infty
 \le\eta(d+1)^2\sqrt\ell
       \left(\frac1{\sqrt n}+\sqrt{\frac{2H}{b}}
                         +\frac1{\sqrt m}\right).
\end{equation}
Let $\bar\beta_{H,1}=\eta\lambda_2\beta_H^\star/2$ denote this first learned
quadratic coefficient. The first inequality in
\eqref{eq:dqn-endstage-feature-budget} and
the sharper $\kappa_d\ge4/(9d^2)$ from the proof of
Lemma~\ref{lem:dqn-axial-stage-projection} imply
\[
 \|Q_{H,1}-q_{H,1}\|_\infty
 \le\frac{\eta\lambda_2}{128Dd^2}
 \le\frac{3\kappa_d \bar\beta_{H,1}}{10}.
\]
Because $\mathbb E_\mu[g_Hv_1]=\kappa_d \bar\beta_{H,1} e_1$,
$\zeta_d\le\sqrt{2/[d(d+2)]}\le1/3$, and the maximum is
$1$-Lipschitz, the actual value continuation copied for update $2$ satisfies
\[
 \|\mathbb E_\mu[g_H\widehat v_{H-1}]\|_2
 \ge\mathbb E_\mu[g_{H,1}\widehat v_{H-1}]
 \ge\kappa_d\bar\beta_{H,1}
       -\zeta_d\|Q_{H,1}-q_{H,1}\|_\infty
 \ge\frac{9\kappa_d\bar\beta_{H,1}}{10}.
\]
Using $\eta\ge1/(4H)$, $\beta_H^\star=1/(8D)$, and the
lower bound on $\kappa_d$, we obtain
\begin{equation}\label{eq:dqn-first-nonterminal-signal}
 S_{H-1}\ge\frac{81\kappa_d^2\bar\beta_{H,1}^2}{3200D^2d^2}
       \ge\frac{\eta^2\lambda_2^2}{51200D^4d^6}
       \ge\frac{\lambda_2^2}{2^{20}H^2D^4d^6}.
\end{equation}

The first learned terminal target also has a sharper magnitude
bound. The moment formula gives
$\beta_{d+2}/\beta_d=\sqrt{d(d+2)}/(d+1)<1$.
The values $\beta_4=4/(3\pi)<4/9$ and
$\beta_5=3\sqrt5/16<4/9$ therefore imply
$\lambda_0=1+2\beta_d^2<7/5$.
Also $\lambda_2=2\beta_d^2/(d+1)\le1/10$.
Using $D\ge2$, $\kappa_d\le1$, and $\|g_{H,1}\|_\infty\le1$ gives
\[
 \|Q_{H,1}\|_\infty
 \le \frac{\eta\lambda_0}{4}+\bar\beta_{H,1}+\kappa_d\bar\beta_{H,1}
 \le\frac{57\eta}{160}
 \le\frac{3\eta}{8}.
\]
The sharper concentration events apply to the actual target at
step $2$. The sampling rows in
\eqref{eq:dqn-endstage-feature-budget} give
\[
 n^{-1/2}+\sqrt{2H/b}
 \le\frac{\alpha_d\lambda_2}
             {128D^2d^{9/2}\sqrt\ell}.
\]
Together with \eqref{eq:dqn-first-nonterminal-signal}, this bounds
the gradient error by
\[
 \mathcal E_{H-1}^{(2)}
   :=\frac{9\eta}{16}\sqrt{d\ell}
             \left(\frac1{\sqrt n}+\sqrt{\frac{2H}{b}}\right)
   \le\frac{9\alpha_d\eta\lambda_2}{2048D^2d^4}
   \le\frac{\alpha_d\sqrt{S_{H-1}}}{d}.
\]
Using this smaller error radius in the signed-action covariance
argument for \eqref{eq:dqn-resolved-feature-gain} yields
\[
 \mathcal G_{H-1}\ge\frac{\alpha_d^2S_{H-1}}{32Hd}
       \ge\frac{\alpha_d^2\lambda_2^2}{2^{25}H^3D^4d^7},
\]
which is \eqref{eq:dqn-first-nonterminal-feature-gain}.
\end{proofof}

\end{document}